\documentclass[10pt,twocolumn,letterpaper]{article}
\usepackage[accsupp]{axessibility}
\usepackage{iccv}
\usepackage{times}
\usepackage{epsfig}
\usepackage{graphicx}
\usepackage{amsmath}
\usepackage{amssymb}
\usepackage{bbding}
\usepackage{booktabs}
\usepackage{multirow}
\usepackage[table]{xcolor}
\usepackage{subcaption}
\usepackage{multicol}
\usepackage[colorlinks,linkcolor=red, breaklinks=true,bookmarks=false]{hyperref}

\usepackage[capitalize]{cleveref}
\crefname{section}{Sec.}{Secs.}
\Crefname{section}{Section}{Sections}
\Crefname{table}{Table}{Tables}
\crefname{table}{Tab.}{Tabs.}

\iccvfinalcopy 

\def\iccvPaperID{****} 

\ificcvfinal\fi

\begin{document}

\title{Implicit Neural Representation for Cooperative Low-light Image Enhancement}

\author{
Shuzhou Yang$^{1,2}$, Moxuan Ding$^{\dag}$, Yanmin Wu$^{1}$, Zihan Li$^{3}$, Jian Zhang$^{\ast 1}$\\
$\quad^{1}$Peking University Shenzhen Graduate School, China \\
$\quad^{2}$Peng Cheng Laboratory, China
$\quad^{3}$University of Washington, USA \\
{\tt\small szyang@stu.pku.edu.cn, zhangjian.sz@pku.edu.cn}
}

\maketitle
\ificcvfinal\thispagestyle{empty}\fi

\begin{abstract}
   The following three factors restrict the application of existing low-light image enhancement methods: unpredictable brightness degradation and noise, inherent gap between metric-favorable and visual-friendly versions, and the limited paired training data. To address these limitations, we propose an implicit \textit{\textbf{Ne}}ural \textit{\textbf{R}}epresentation method for \textit{\textbf{Co}}operative low-light image enhancement, dubbed \textit{\textbf{NeRCo}}. It robustly recovers perceptual-friendly results in an unsupervised manner. Concretely, NeRCo unifies the diverse degradation factors of real-world scenes with a controllable fitting function, leading to better robustness. In addition, for the output results, we introduce semantic-oriented supervision with priors from the pre-trained vision-language model. Instead of merely following reference images, it encourages results to meet subjective expectations, finding more visual-friendly solutions. Further, to ease the reliance on paired data and reduce solution space, we develop a dual-closed-loop constrained enhancement module. It is trained cooperatively with other affiliated modules in a self-supervised manner. Finally, extensive experiments demonstrate the robustness and superior effectiveness of our proposed NeRCo. Our
   code is available at \url{https://github.com/Ysz2022/NeRCo}.
\end{abstract}

\renewcommand{\thefootnote}{}
\footnotetext{Corresponding author$^{\ast}$. Independent researcher$^{\dag}$.}
\footnotetext{This work was supported in part by Shenzhen General Research Project under Grant JCYJ20220531093215035.}

\begin{figure}[t]
  \centering
  \begin{subfigure}{0.49\linewidth}
    \includegraphics[width=1.\linewidth]{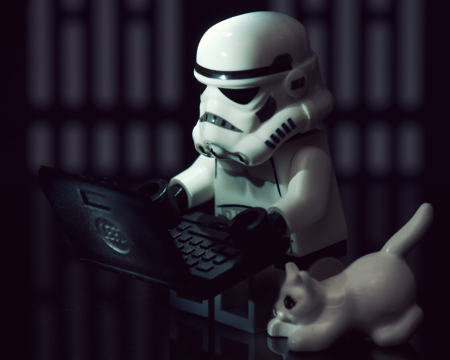}\vspace{-0.4em}
    
    \centerline{Input}
    
    \includegraphics[width=1.\linewidth]{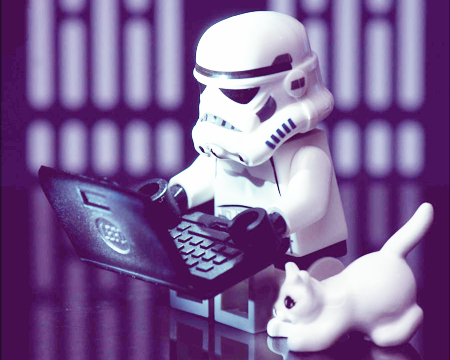}\vspace{-0.4em}
    
    \centerline{\small SCI (2022) \cite{Ma_2022_CVPR}}
    \vspace{-1.em}
    
  \end{subfigure}
  \hfill
  \begin{subfigure}{0.49\linewidth}
    \includegraphics[width=1.\linewidth]{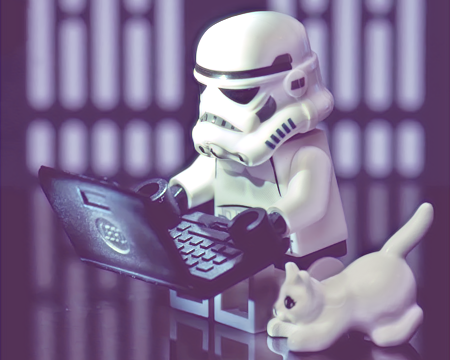}\vspace{-0.4em}
    
    \centerline{\small URetinexNet (2022) \cite{Wu_2022_CVPR}}
    
    \includegraphics[width=1.\linewidth]{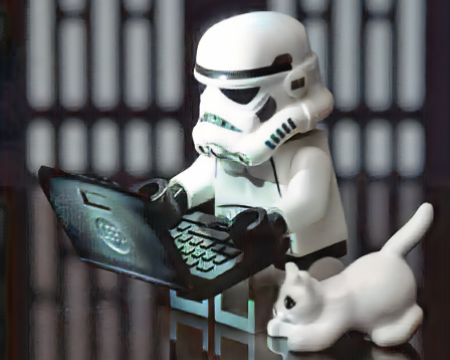}\vspace{-0.4em}
    
    \centerline{NeRCo (Ours)}
    \vspace{-1.em}
    
  \end{subfigure}

  \caption{Comparison with two state-of-the-art methods on the LIME \cite{Guo_2017_TIP} dataset. One can see that we recovered more authentic color and visual-friendly contents.}
  \label{Fig1}
  \vspace{-1.7em}
\end{figure}

\vspace{-1.em}

\section{Introduction}
Due to the degraded brightness in low-light images covering objects and reducing contrast, low-light images have severely impacted the subsequent high-level computer visual tasks (\textit{e.g.}, semantic segmentation \cite{Islam_2020_IROS}, and object detection \cite{Liu_2016_ECCV}, \textit{etc.}). Hence, it is of practical importance to remedy the brightness degradation for assisting the exploration of sophisticated dark environment. Low-light image enhancement, which aims to recover the desired content in degraded regions, has drawn wide attention in recent years \cite{Fu_2016_CVPR,Guo_2020_CVPR,Guo_2017_TIP,Jiang_2021_TIP,Liu_2023_arxiV,Pisano_1998_JDI}.

Over the past few years, prolific algorithms have been proposed to address this classic ill-posed problem, which can be roughly categorized into two groups: conventional model-based methods (\textit{e.g.}, gamma correction \cite{Moroney_2000_CIC}, Retinex-based model \cite{Ng_2011_SIAM}, and histogram equalization \cite{Pisano_1998_JDI}) and recent deep learning-based methods \cite{Jiang_2021_TIP,Liu_2021_CVPR,Wu_2022_CVPR}. The former formulates the degradation as a physical model and treats enhancement as the problem of estimating model parameters, but is limited in characterizing diverse low-light factors and requires massive hand-crafted priors. The latter elaborates various models to adjust tone and contrast, which is able to learn from massive data automatically. Essentially, they are trained to learn a mapping from input to output domain. In real-world scenarios, however, many samples are far away from the feature space of input domain, causing a trained model to lack stable effect. We propose to normalize the degradation before enhancement to bring these samples closer to the input domain. Besides, existing supervised methods highly rely on paired training data and mainly attempt to produce metric-favorable results, \textit{i.e.}, similar to the ground truth. But the limited supervised datasets and the inherent gap between metric-oriented and visual-friendly versions inevitably impact their effectiveness. We develop a cooperative training strategy to address it. As shown in \cref{Fig1}, we test on the LIME \cite{Guo_2017_TIP} dataset, which only consists of low-light images without normal-light references. One can see that even the recently proposed top-performing algorithms perform severe color cast.

Specifically, our key insights are: \textbf{i) Normalizing the input with a controllable fitting function to reduce the unpredictable degradation features in real-world scenarios.} We adopt neural representation to reproduce the degraded scene before the enhancement operation. By manipulating the positional encoding, we selectively avoid regenerating extreme degradation, which objectively realizes normalization and thereby decreases enhancement difficulty. \textbf{ii) Supervising the output with different modalities to achieve both metric-favorable and perceptual-oriented enhancement.} We employ multi-modal learning to supervise from both textual and image perspectives. Compared with image supervision, which contains varying brightness across different samples, the feature space of the designed prompt is more stable and accurate in describing brightness. During training, our results are not only encouraged to be similar to references, but also forced to match their related prompts. In this way, we bridge the gap between the metric-favorable and the perceptual-friendly versions. \textbf{iii) Developing an unsupervised training strategy to ease the reliance on the paired data.} We propose to train the enhancement module with a dual-closed-loop cooperative adversarial constraint procedure, which learns in an unsupervised manner. More related loss functions are also proposed to further reduce the solution space. Benefiting from these, we recover more authentic tone and better contrast (see \cref{Fig1}). Overall, our contributions are as follows:
\begin{itemize}
    \item[$\bullet$]
    We are the first to utilize the controllable fitting capability of neural representation in low-light image enhancement. It normalizes lightness degradation and removes natural noise without any additional operations, providing new ideas for future work.
    \item[$\bullet$]
    For the first time, we introduce multi-modal learning to low-light image enhancement. Benefiting from its efficient vision-language priors, our method learns diverse features, resulting in perceptually better results.
    \item[$\bullet$]
    We develop an unsupervised cooperative adversarial learning strategy to ease the reliance on the paired training data. In which the appearance-based discrimination ensures authenticity from both color and detail levels, improving the quality of the restored results.
    \item[$\bullet$]
    Extensive experiments are conducted on representative benchmarks, manifesting the superiority of our NeRCo against a rich set of state-of-the-art algorithms. Especially, it even outperforms some supervised methods.
\end{itemize}

\section{Related Work}
\subsection{Low-light Image Enhancement}
To improve the visibility of low-light images, model-based methods are first widely adopted. Retinex theory \cite{Zia-ur_2004_JEI} decomposes the observation into illumination and reflectance (\textit{i.e.}, clear prediction), but tends to over-expose the appearance. Various hand-crafted priors are further introduced into models as regularization terms. Fu~\textit{et al.} \cite{Fu_2016_CVPR} developed a weighted variational model to simultaneously estimate reflectance and illumination layers. Cai~\textit{et al.} \cite{Cai_2017_ICCV} proposed an edge-preserving smoothing algorithm to model brightness. Guo~\textit{et al.} \cite{Guo_2017_TIP} predicted the illumination by adopting the relative total variation \cite{Xu_2012_TOG}. However, these defined priors are labor-intensive and perform poor generalization towards real-world scenarios.

Due to these limitations, researchers took advantage of deep learning to recover in a data-driven manner \cite{Cai_2018_TIP,Guo_2020_CVPR,Liu_2021_CVPR,Wu_2022_CVPR}, which exploits priors from massive data automatically. For example, Guo~\textit{et al.} \cite{Guo_2020_CVPR} formulated light enhancement as a task of image-specific curve estimation with a lightweight deep model. Jiang~\textit{et al.} \cite{Jiang_2021_TIP} introduced adversarial training for learning from unpaired supervision. Wei~\textit{et al.} \cite{LOL} designed an end-to-end trainable RetinexNet but still troubled by heavy noise. To ameliate it, Zhang~\textit{et al.} \cite{SSIENet} proposed a decomposition-type architecture to impose constraint on reflectance. Liu~\textit{et al.} \cite{Liu_2021_CVPR} employed architecture search and built an unrolling network. Although these well-designed models have realized impressive effectiveness, they are not stable in real-world applications. To improve robustness, we pre-modulate the degradation to a uniform level with neural representation before the enhancement procedure.

\begin{figure*}[t]
  \centering
  \begin{subfigure}{1\linewidth}
    \includegraphics[width=1.\linewidth]{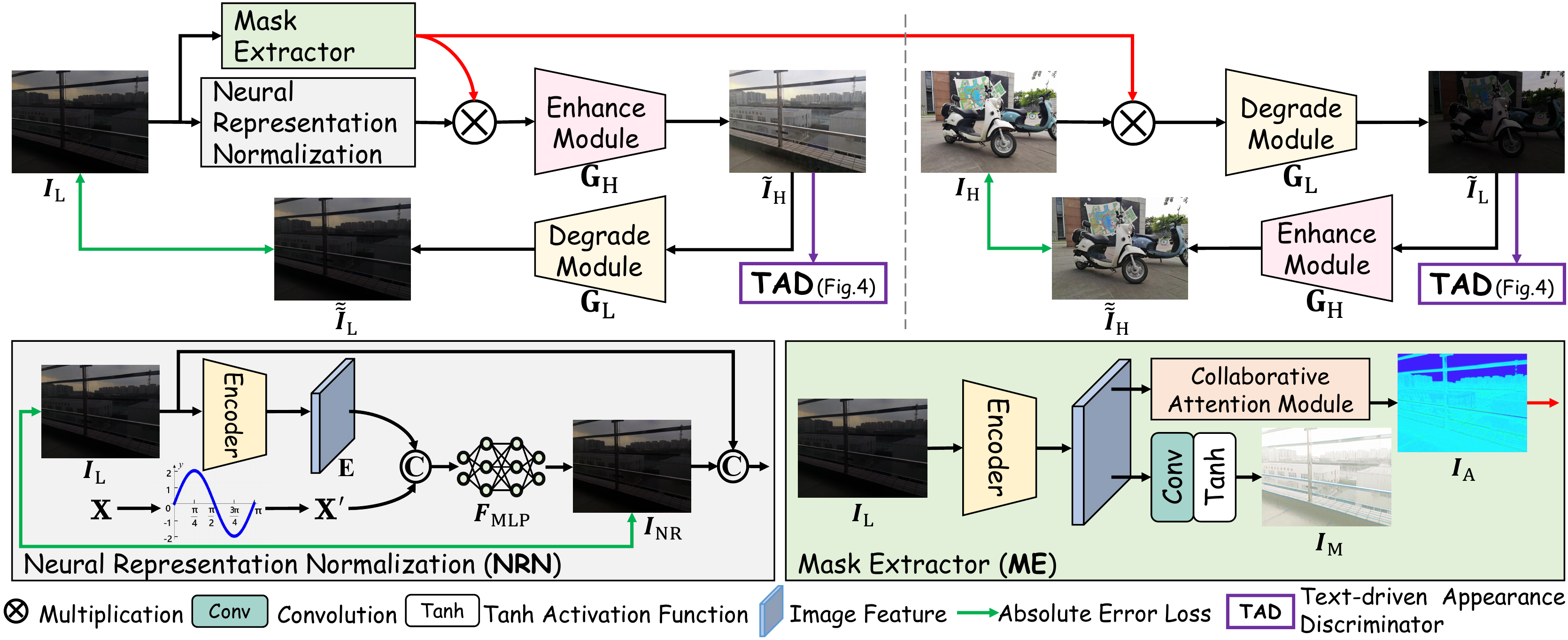}
  \end{subfigure}
  \vspace{-2.3em}
  
  \caption{Workflow of our NeRCo. It presents a cooperative adversarial enhancement process containing dual-closed-loop branches, each of which contains an enhancement operation and a degradation operation. We embed a Mask Extractor (ME) to portrait the degradation distribution and a Neural Representation Normalization (NRN) module to normalize the degradation condition of the input low-light image. All of them are trained together to constrain each other, locking on to a more accurate target domain. The \textcolor{red}{red} means the transfer of the attention map.}
  \label{totnet}
  \vspace{-1.1em}
  
\end{figure*}

\subsection{Neural Representation for Images}
Recently, neural representation has been widely adopt to depict images. Chen~\textit{et al.} \cite{Chen_2021_CVPR} firstly utilized implicit image representation for continuous image super-resolution. However, the MultiLayer Perceptron (MLP) tends to distort high-frequency components. To address this issue, Lee~\textit{et al.} \cite{Lee_2022_CVPR} developed a dominant-frequency estimator to predict local texture for natural images. Lee~\textit{et al.} \cite{Lee_2022_ECCV} further utilized implicit neural representation to warp images into continuous shape. Dupont~\textit{et al.} \cite{Dupont_2022_pmlr} tried to produce different objects with one MLP by manipulating the latent code from its hidden layers. Saragadam~\textit{et al.} \cite{Saragadam_2022_ECCV} adopted multiple MLPs to represent a single image in a multi-scale manner. Sun~\textit{et al.} \cite{Sun_2021_TCI} predicted continuous information based on the captured tomographic features. Tancik~\textit{et al.} \cite{Tancik_2021_CVPR} introduced meta-learning to initialize the parameters of MLP to accelerate training. Reed~\textit{et al.} \cite{Reed_2021_ICCV} adopted neural representation and parametric motion fields to predict the shape and location of organs. Further, some researchers adopted neural representation to compress videos \cite{Bai_2022_arxiV,Chen_2021_NIPS,Zhang_2022_ICLR}.

However, existing neural representation is mainly applied on image compressing, denoising and depicting continuous information, \textit{etc.} We are the first to apply its controllable fitting capability to low-light image enhancement.

\subsection{Multi-modal Learning}
In recent years, learning across modalities has attracted extensive attention \cite{Cheng_2023_ICASSP1,Cheng_2023_ICASSP2,Li_2023_ICCV,Li_2023_arxiV,Yang_2023_ICCV}. Various vision-language models are developed. Radford~\textit{et al.} \cite{Radford_2021_ICML} proposed to learn visual model from language supervision, called CLIP. After training on 400 million image-text pairs, it can describe any visual concept with natural language and transfer to other tasks without any specific training. Furthermore, Zhou~\textit{et al.} \cite{Zhou_2022_IJCV} developed soft prompts to replace the hand-crafted ones, which uses learnable vectors to model context words and obtains task-relevant context. To further refine prompts to the instance-level, Rao~\textit{et al.} \cite{Rao_2022_CVPR} designed context-aware prompting to combine prompts with visual features. Cho~\textit{et al.} \cite{Cho_2021_ICML} shared priors across different tasks by updating a uniform framework towards a common target of seven multi-modal tasks. Ju~\textit{et al.} \cite{Ju_2022_ECCV} adopted the pre-trained CLIP model to video understanding.

Existing methods mainly focus on high-level computer vision tasks such as image classification. For the first time, we apply priors of the pre-trained vision-language model to low-light image enhancement, developing semantic-oriented guidance and realizing better performance.


\section{Our Method}
\subsection{Framework Architecture}

As shown in \cref{totnet}, given a low-light image $\textit{\textbf{I}}_{\rm{L}}$, we first normalize it by neural representation (NRN, \cref{sec:NR}) to improve the robustness of the model to different degradation conditions. Then the Mask Extractor (ME, \cref{sec:DLGP}) module extracts the attention mask from the image to guide the enhancement of different regions. After that, the Enhance Module $\textit{G}_{H}$ (represented by ResNet) generates a high-light image $\widetilde{\textbf{\textit{I}}}_{\rm{H}}$. To ensure its quality, we design a Text-driven Appearance Discriminator (TAD, \cref{sec:TAD}) to supervise image generation, where text-driven supervision guarantees semantic reliability, and appearance supervision guarantees visual reliability. $\widetilde{\textbf{\textit{I}}}_{\rm{H}}$ is then passed through the Degrade Module $\textit{G}_{L}$ to convert back to the low-light domain $\widetilde{\widetilde{\textbf{\textit{I}}}}_{\rm{L}}$ and calculate the consistency loss (\cref{sec:DLGP}) with the original low-light image $\textit{\textbf{I}}_{\rm{L}}$. The upper right branch of \cref{totnet} inputs the high-light image $\textit{\textbf{I}}_{\rm{H}}$, implemented in a similar way.

The network is realizeed in a dual-loop way (\cref{sec:DLGP}) to achieve stable constraints based on the unpaired data. It operates bidirectional mapping: enhance-degrade ($\textit{\textbf{I}}_{\rm{L}} \to \widetilde{\textbf{\textit{I}}}_{\rm{H}} \to \widetilde{\widetilde{\textbf{\textit{I}}}}_{\rm{L}}$) and degrade-enhance ($\textit{\textbf{I}}_{\rm{H}} \to \widetilde{\textbf{\textit{I}}}_{\rm{L}} \to \widetilde{\widetilde{\textbf{\textit{I}}}}_{\rm{H}}$). This dual loop constraint fully exploits the latent general distinction between the low-light and high-light domains. Besides, the cooperative loss (\cref{sec:DLGP}) encourages all components in the framework to supervise each other collaboratively, which further reduces the solution space.

During \textbf{training}, we run the whole process in \cref{totnet}. We input two images (\textit{i.e.}, low-light $\textit{\textbf{I}}_{\rm{L}}$ and high-light $\textit{\textbf{I}}_{\rm{H}}$). $\textit{\textbf{I}}_{\rm{L}}$ is enhanced to $\widetilde{\textbf{\textit{I}}}_{\rm{H}}$, then translated back to low-light $\widetilde{\widetilde{\textbf{\textit{I}}}}_{\rm{L}}$. And vice versa for $\textit{\textbf{I}}_{\rm{H}}$. Noting that $\textit{\textbf{I}}_{\rm{H}}$ is used for training purposes only, \textit{i.e.}, training model in an unsupervised manner for better enhancement rather than degradation. Hence, we only use NRN to enhance $\textit{\textbf{I}}_{\rm{L}}$ but not degrade $\textit{\textbf{I}}_{\rm{H}}$. For \textbf{inference}, $\textit{\textbf{I}}_{\rm{L}}$ is directly enhanced to $\widetilde{\textbf{\textit{I}}}_{\rm{H}}$ without any other operations, as shown in the top left part of \cref{totnet}.

\subsection{Neural Representation for Normalization}
\label{sec:NR}
\begin{figure}[t]
  \centering
  \begin{subfigure}{0.99\linewidth}
    \includegraphics[width=1.\linewidth]{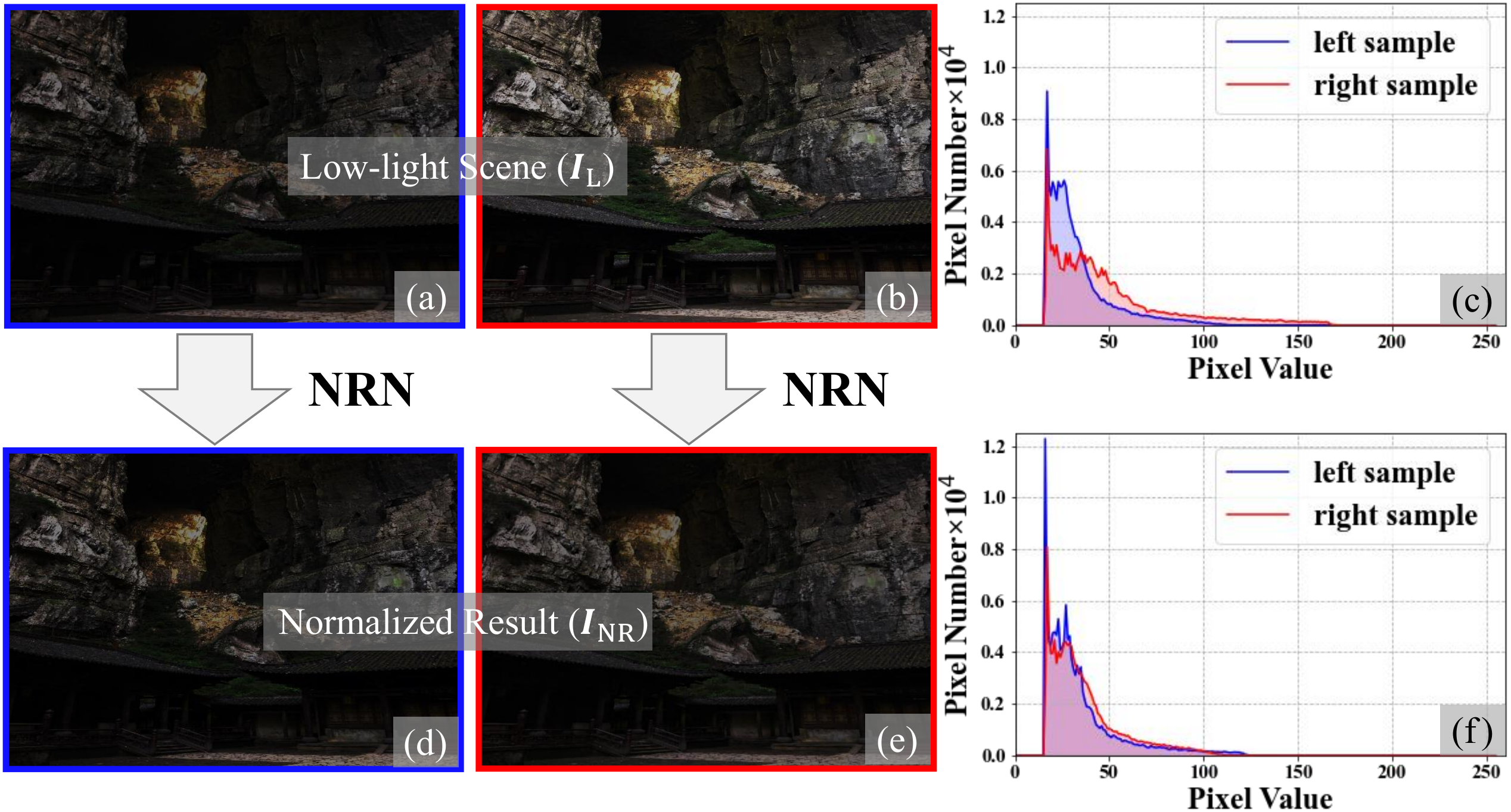}
  \end{subfigure}
  \vspace{-1.em}

  \caption{Comparisons between the captured low-light scenes (top row) and the results of NRN (bottom row). The low-light samples are from the SICE \cite{Cai_2018_TIP} dataset. It contains numerous image sets, each with common content and varying lighting conditions. The pixel value distribution of images on the Y channel is given on the right. One can see that NRN normalizes the brightness to be similar.}
  \label{NRPP}
  \vspace{-1.em}
\end{figure}

\textbf{Motivation.} Images captured in real-word typically exhibit varying degradation levels due to lighting conditions or camera parameters, as shown in Fig.~\ref{NRPP} (a)(b). We report their pixel value distribution on the Y channel in Fig.~\ref{NRPP} (c). The inconsistency between these samples is challenging for a well-trained model. We attempt to normalize degradation level (see Fig.~\ref{NRPP} (d)(e)) with neural representations (NR) to obtain a more consistent degradation distribution (see Fig.~\ref{NRPP} (f)) to reduce the difficulty of subsequent operations.

\textbf{Neural Representation.} Concretely, in NR, image $\textit{\textbf{I}}_{\rm{L}}$ is transformed into a feature map $\textbf{E} \in \mathbb{R}^{H \times W \times C}$, where \textit{H} and \textit{W} are image resolution. While the location of each pixel is recorded in a coordinate set $\textbf{X} \in \mathbb{R}^{H \times W \times 2}$, where 2 means horizontal and vertical coordinates. $\textit{\textbf{I}}_{\rm{L}}$ can thus be represented by its features and a set of coordinates. As shown in the Neural Representation Normalization (NRN) module of \cref{totnet}, we fuse $\textbf{X}$ and $\textbf{E}$, and use a decoding function $\textbf{\textit{F}}_{\rm{MLP}}$ to output image $\textit{\textbf{I}}_{\rm{NR}}$, which is parameterized as a MultiLayer Perceptron (MLP). The neural representation of the image is expressed as:
\begin{eqnarray}
\begin{aligned}
	\label{shi10}
	\textit{\textbf{I}}_{\rm{NR}}[\textit{i}, \textit{j}] = \textbf{\textit{F}}_{\rm{MLP}}(\textbf{E}[\textit{i}, \textit{j}],\textbf{X}[\textit{i}, \textit{j}]),
\end{aligned}
\end{eqnarray}
where $[\textit{i}, \textit{j}]$ is the location of a pixel and $\textit{\textbf{I}}_{\rm{NR}}[\textit{i}, \textit{j}]$ is the generated RGB value. By predicting RGB of each pixel, an image $\textit{\textbf{I}}_{\rm{NR}}$ is reproduced. We encourage $\textit{\textbf{I}}_{\rm{NR}}$ to be similar to $\textit{\textbf{I}}_{\rm{L}}$ through $\textit{l}_1$-norm. This NR-related loss expressed as:
\begin{eqnarray}
\begin{aligned}
	\mathcal{L}_{NR} = ||\textit{\textbf{I}}_{\rm{NR}}-\textit{\textbf{I}}_{\rm{L}}||_1.
\end{aligned}
\end{eqnarray}

\textbf{Why Neural Representation Works.} With the trained $\textbf{\textit{F}}_{\rm{MLP}}$, each feature map $\textbf{E}$ can form a function $\textbf{\textit{F}}_{\rm{MLP}}(\textbf{E},\cdot) : \textbf{X} \rightarrow \textit{\textbf{I}}_{\rm{NR}}$, which maps coordinates to its predicted RGB values. Without $\textbf{E}$, it is impossible for $\textbf{\textit{F}}_{\rm{MLP}}$ to depict various RGB values with the same coordinates $\textbf{X}$. Without $\textbf{X}$, we cannot normalize degradation by adjusting fitting capability, which is explained below.

\begin{figure*}[t]
  \centering
  \begin{subfigure}{1\linewidth}
    \includegraphics[width=1.\linewidth]{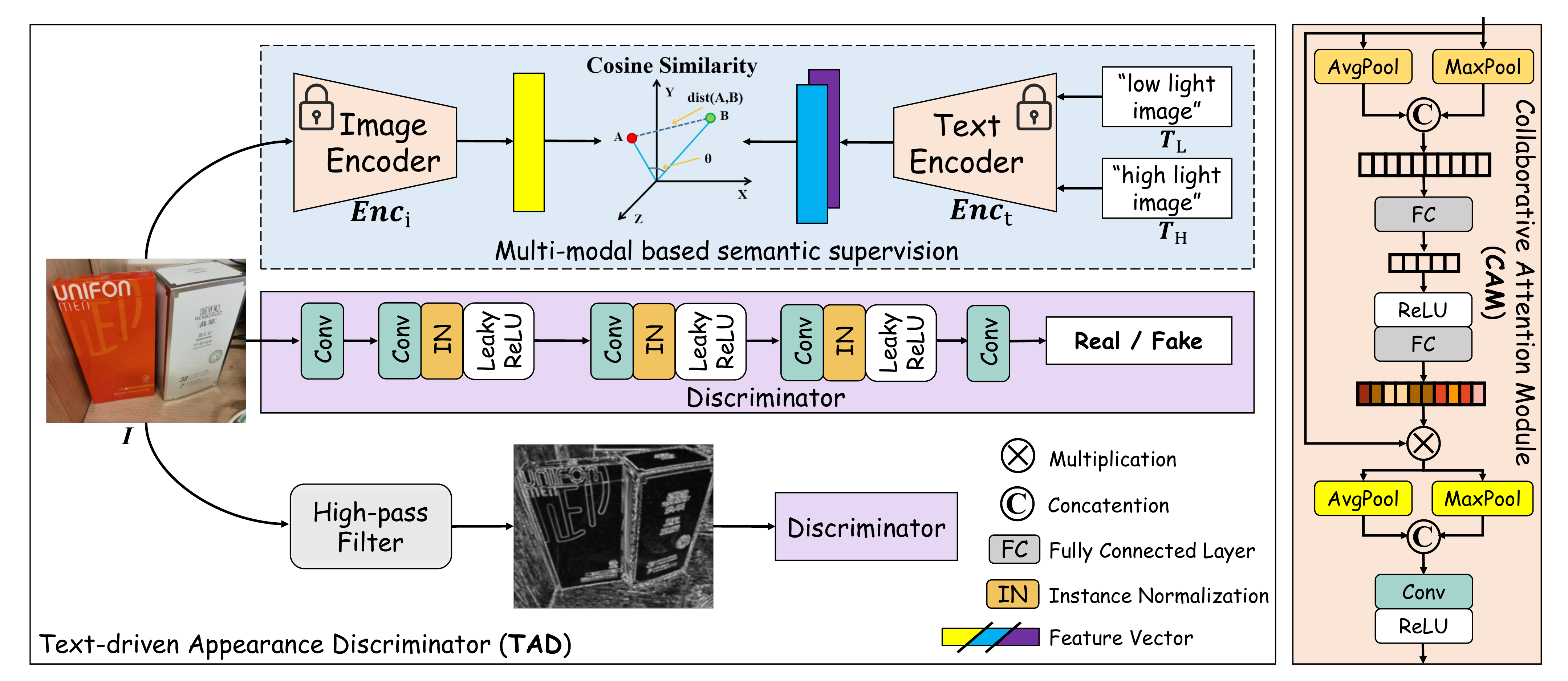}
  \end{subfigure}
  \vspace{-2.em}
  
  \caption{The details of our proposed text-driven appearance discriminator (in the left region) and our collaborative attention module (in the right region). The former supervises the input with text and image modalities, and focuses on high-frequency components. The latter adjusts attention to different channels and outputs attention map.}
  \label{discr}
  \vspace{-1.em}
  
\end{figure*}

According to \cite{Rahaman_2019_ICML}, neural networks tend to portrait lower frequency information. Despite our decoding function can approximate RGB values, some high-frequency components may be discarded during rendering. For example, for adjacent pixels around the edge, their RGB values vary a lot but coordinates vary little. It means $\textbf{\textit{F}}_{\rm{MLP}}$ should output different results based on similar inputs, which is difficult. Inspired by \cite{Mildenhall_2020_ECCV}, to fit high-frequency variation better, we map the input coordinates to a higher dimensional space before passing them to $\textbf{\textit{F}}_{\rm{MLP}}$, which is called positional encoding. As shown in the gray region of \cref{totnet}, before fusing coordinates with image feature, we use a high-frequency function $\boldsymbol{\gamma}(\cdot)$ to map the original coordinates $\textbf{\textit{x}}$ from $\mathbb{R}$ into a higher dimensional space $\mathbb{R}^{2L}$, expressed as:
\begin{eqnarray}
\begin{aligned}
	\label{shi11}
	\boldsymbol{\gamma}(\textbf{\textit{x}}) = (\cdot\cdot\cdot, \rm{sin}(2^{\textit{i}}\pi\textbf{\textit{x}}), \rm{cos}(2^{\textit{i}}\pi\textbf{\textit{x}}), \cdot\cdot\cdot),
	\\
\end{aligned}
\end{eqnarray}
where $\textit{i}$ values from 0 to $\textit{L} - 1$, $\textit{L}$ is a hyperparameter that determines the dimension value. The final coordinates are composed as: $\textbf{\textit{x}}^{\prime} = \boldsymbol{\gamma}(\textbf{\textit{x}})$. Noting that by manipulating the value of $\textit{L}$, we can change the fitting capacity of our NRN module, \textit{i.e.}, a bigger $\textit{L}$ results in a more precise fit.

However, stronger fitting capability is not always better. Since NRN aims to normalize various degradation, its output is not expected to be exactly the same as the input, especially the degradation components. We want to choose an $\textit{L}$ that does not overfit to remain all information while faithfully preserves desired content. Chen \textit{et al.} \cite{Chen_2021_NIPS} have demonstrated that NR is robust to perturbations and can denoise without any special design. We think this is because MLPs lack spatial correlation priors, some extreme information is thus hard to be reproduced faithfully. Hence, this underfitting property objectively limits the unpredictable degradation. We further found in experiments that MLPs tend to learn an average brightness range of training data rather than fitting the unique brightness of different images. As shown in \cref{NRPP}, the output lightness of a trained MLP is similar. By visualizing their pixel value distribution in the Y channel, we further prove the normalization property of our NRN. More results are given in the \textcolor{red}{supplementary material (SM)}. Thereby, we set $\textit{L}$ = 8 for the trade-off between degradation normalization and content fidelity. By reducing these fickle degradation signals, we decrease the difficulty of subsequent enhancement procedure.


\subsection{Text-driven Appearance Discriminator}
\label{sec:TAD}
\textbf{Motivation.} Existing methods mainly adopt image-level supervision, \textit{i.e.}, forcing the output to be close to the target images. However, the brightness across different references varies greatly, confusing model training; some references are visually poor (\textit{e.g.}, unnatural lightness), leading to visual-unfriendly results. To reduce training difficulty and bridge the gap between the metric-favorable and visual-friendly versions, we design a Text-driven Appearance Discriminator (TAD) to supervise image generation from the semantic level and appearance level, respectively.

\textbf{Text-driven Discrimination}. We denote the low-light domain as $\boldsymbol{\mathcal{L}}$ and the high-light domain as $\boldsymbol{\mathcal{H}}$. As shown in \cref{discr}, we introduce multi-modal learning to supervise images with both image and text modalities. Concretely, inspired by Radford~\textit{et al.} \cite{Radford_2021_ICML}, we employ the recent well-known CLIP model to get efficient priors. It consists of two pre-trained encoders, \textit{i.e.}, $\textbf{\textit{Enc}}_{\rm{t}}$ for text and $\textbf{\textit{Enc}}_{\rm{i}}$ for image. We first manually design two prompts, \textit{i.e.}, $\textit{low-light image}$ and $\textit{high-light image}$, to describe $\boldsymbol{\mathcal{L}}$ and $\boldsymbol{\mathcal{H}}$ respectively, denoted as $\textit{\textbf{T}}_{\rm{L}}$ and $\textit{\textbf{T}}_{\rm{H}}$ in \cref{discr}. Experiments on more other texts are given in the \textcolor{red}{SM}. $\textbf{\textit{Enc}}_{\rm{t}}$ extracts two feature vectors of size $1 \times 512$ from two prompts. Similarly, $\textbf{\textit{Enc}}_{\rm{i}}$ extracts a vector of the same size from our intermediate result $\textit{\textbf{I}}$. We compute the cosine similarity between the image vector and the text vector to measure their discrepancy, formulated as:
\begin{eqnarray}
\begin{aligned}
    \label{cosh}
    \mathcal{D}_{cos}(\textit{\textbf{I}}, \textit{\textbf{T}}) &= \frac{\langle \textbf{\textit{Enc}}_{\rm{i}}(\textit{\textbf{I}}), \textbf{\textit{Enc}}_{\rm{t}}(\textit{\textbf{T}})\rangle}{||\textbf{\textit{Enc}}_{\rm{i}}(\textit{\textbf{I}})|| \, ||\textbf{\textit{Enc}}_{\rm{t}}(\textit{\textbf{T}})||},
\end{aligned}
\end{eqnarray}
where $\textit{\textbf{I}}$ denotes the predicted image and $\textit{\textbf{T}}$ is the prompt. For the enhanced results (\textit{e.g.}, $\widetilde{\textbf{\textit{I}}}_{\rm{H}}$ and $\widetilde{\widetilde{\textbf{\textit{I}}}}_{\rm{H}}$ in \cref{totnet}), we encourage their vectors to be similar to those of $\textit{\textbf{T}}_{\rm{H}}$ and away from $\textit{\textbf{T}}_{\rm{L}}$, and vice versa for low-light predictions. In this way, we encourage semantically consistent outputs. This cosine objective function is formulated as:
\begin{equation}
\begin{aligned}
	\label{cosh}
	\mathcal{L}_{cos}(\textit{\textbf{I}}_{\rm{H}},\textit{\textbf{T}}_{\rm{L}},\textit{\textbf{T}}_{\rm{H}}) = \mathcal{D}_{cos}(\textit{\textbf{I}}_{\rm{H}},\textit{\textbf{T}}_{\rm{L}}) - \mathcal{D}_{cos}(\textit{\textbf{I}}_{\rm{H}},\textit{\textbf{T}}_{\rm{H}}),
\end{aligned}
\end{equation}
\begin{equation}
\begin{aligned}
	\mathcal{L}_{cos}(\textit{\textbf{I}}_{\rm{L}},\textit{\textbf{T}}_{\rm{L}},\textit{\textbf{T}}_{\rm{H}}) = \mathcal{D}_{cos}(\textit{\textbf{I}}_{\rm{L}},\textit{\textbf{T}}_{\rm{H}}) - \mathcal{D}_{cos}(\textit{\textbf{I}}_{\rm{L}},\textit{\textbf{T}}_{\rm{L}}).
\end{aligned}
\end{equation}

\textbf{Appearance-based Discrimination.} Admittedly, text descriptions cannot provide low-level guidance like images. To generate faithful content, image supervision is necessary. As shown in the purple region of \cref{discr}, we stack a discriminator to distinguish the predicted results from real images, encouraging image-level authenticity (\textit{e.g.}, color, texture, and structure). Considering detail distortion in image processing, we embed a high-frequency path consisting of a high-pass filter and a discriminator of the same structure. The filter extracts high-frequency components and the discriminator supervises them at edge-level. Based on this double-path color-edge discriminative structure, we realize the trade-off between color and detail.

During training, TAD plays an adversarial role in learning bidirectional mapping relationship between $\boldsymbol{\mathcal{L}}$ and $\boldsymbol{\mathcal{H}}$. We develop an adversarial loss on each generation loop to realize it. As shown in \cref{totnet}, for the enhancement operation $\textit{G}_{H}(\cdot)$: $\textit{\textbf{I}}_{\rm{L}}$ $\rightarrow$ $\widetilde{\textbf{\textit{I}}}_{\rm{H}}$, we apply a TAD module and dub its appearance discrimination as $\textit{D}_{H}$. Developing an adversarial objective function as:
\begin{small}
\begin{equation}
\begin{aligned}
	\label{shi3}
	\mathcal{L}_{adv}(\textit{\textbf{I}}_{\rm{L}},\textit{\textbf{I}}_{\rm{H}},\textit{G}_{H},\textit{D}_{H}) &= \mathcal{L}_{cos}(\textit{G}_{H}(\textit{\textbf{I}}_{\rm{L}}),\textit{\textbf{T}}_{\rm{L}},\textit{\textbf{T}}_{\rm{H}}) \\
	&+ \mathbb{E}_{\textit{\textbf{I}}_{\rm{H}} \sim \boldsymbol{\mathcal{H}}}[\log\textit{D}_{H}(\textit{\textbf{I}}_{\rm{H}})] \\
	&+ \mathbb{E}_{\textit{\textbf{I}}_{\rm{L}} \sim \boldsymbol{\mathcal{L}}}[\log(1-\textit{D}_{H}(\textit{G}_{H}(\textit{\textbf{I}}_{\rm{L}})))],
\end{aligned}
\end{equation}
\end{small}where $\textit{D}_{H}$ aims to determine whether an image is captured or generated, that is, distinguish the enhanced results $\textit{G}_{H}(\textit{\textbf{I}}_{\rm{L}})$ from the real high-light domain $\boldsymbol{\mathcal{H}}$. While $\textit{G}_{H}(\cdot)$ aims at deceiving $\textit{D}_{H}$, \textit{i.e.}, generating results close to $\boldsymbol{\mathcal{H}}$. Simultaneously, the aforementioned cosine constraint is also adopted to supervise $\textit{G}_{H}(\cdot)$. The reverse mapping $\textit{G}_{L}(\cdot)$: $\textit{\textbf{I}}_{\rm{H}}$ $\rightarrow$ $\widetilde{\textbf{\textit{I}}}_{\rm{L}}$ adopts a similar objective function, which is supervised by another TAD.

In both functions, generators try to minimize the objective, whilst TAD modules maximize it. During this adversarial learning, our model realizes better semantic consistency, which is proved by experiments presented in \textcolor{red}{SM}.

\subsection{Dual Loop Generation Procedure}
\label{sec:DLGP}
Previous methods mainly map the low-light image $\textit{\textbf{I}}_{\rm{L}}$ to its high-light version $\widetilde{\textbf{\textit{I}}}_{\rm{H}}$ directly. To provide stable constraint without the paired data, we stack a forward enhancement module with a backward degradation one for bidirectional mapping, which operates in an unsupervised manner.

\textbf{Dual Loop.} Specifically, the forward enhancement procedure aims to realize the mapping $\textit{G}_{H}(\cdot)$: $\textit{\textbf{I}}_{\rm{L}}$ $\rightarrow$ $\widetilde{\textbf{\textit{I}}}_{\rm{H}}$. While the other does the opposite, \textit{i.e.}, depicting a low-light scene from a clean image $\textit{\textbf{I}}_{\rm{H}}$ with the mapping $\textit{G}_{L}(\cdot)$: $\textit{\textbf{I}}_{\rm{H}}$ $\rightarrow$ $\widetilde{\textbf{\textit{I}}}_{\rm{L}}$. Our generation loop is composed of alternating these two operations. As shown in the left end of \cref{totnet}, our input is the observed low-light image. It is first extracted an attention-guidance $\textit{\textbf{I}}_{\rm{A}}$ and normalized by a neural representation module. Then the subsequent procedure first translates it to the high-light domain, and maps the enhanced image $\widetilde{\textbf{\textit{I}}}_{\rm{H}}$ back to the low-light version $\widetilde{\widetilde{\textbf{\textit{I}}}}_{\rm{L}}$. This enhancement-degradation generation branch is formulated as:
\begin{eqnarray}
	\label{shi1}
	\widetilde{\widetilde{\textbf{\textit{I}}}}_{\rm{L}} = \textit{G}_{L}( \widetilde{\textbf{\textit{I}}}_{\rm{H}}) = \textit{G}_{L}(\textit{G}_{H}(\textit{ME}(\textbf{\textit{I}}_{\rm{L}}) \otimes \textit{NRN}(\textbf{\textit{I}}_{\rm{L}}))),
\end{eqnarray}
where $\textit{G}_{H}(\cdot)$ and $\textit{G}_{L}(\cdot)$ denote enhancement and degeneration operations, respectively, and $\textit{NRN}$ is the neural representation normalization module discussed in \cref{sec:NR}. $\textit{ME}$ means mask extractor, as shown in the green region of \cref{totnet}, in which we develop a Collaborative Attention Module (CAM) to extract the attention map $\textit{\textbf{I}}_{\rm{A}}$. The details of CAM are displayed at the right end of \cref{discr}.

The degeneration-enhancement generation branch, as shown in the right end of  \cref{totnet}, is formulated similarly:
\begin{eqnarray}
	\label{shi2}
	\widetilde{\widetilde{\textbf{\textit{I}}}}_{\rm{H}} = \textit{G}_{H}(\widetilde{\textbf{\textit{I}}}_{\rm{L}}) = \textit{G}_{H}(\textit{G}_{L}(\textit{ME}(\textbf{\textit{I}}_{\rm{L}}) \otimes \textbf{\textit{I}}_{\rm{H}})).
\end{eqnarray}

To constrain this bidirectional mapping, during training, we develop cycle consistency to directly impose supervision at the pixel-level. For example, for the left branch in \cref{totnet}, we ensure that: $\textbf{\textit{I}}_{\rm{L}} \approx \widetilde{\widetilde{\textbf{\textit{I}}}}_{\rm{L}} = \textit{G}_{L}(\textit{G}_{H}(\textbf{\textit{I}}_{\rm{L}}))$. Accordingly, the other cycle follows: $\textbf{\textit{I}}_{\rm{H}} \approx \widetilde{\widetilde{\textbf{\textit{I}}}}_{\rm{H}} = \textit{G}_{H}(\textit{G}_{L}(\textbf{\textit{I}}_{\rm{H}}))$. We adopt the $\textit{l}_1$-norm to measure discrepancy and develop the consistency constraint as:
\begin{eqnarray}
\begin{aligned}
	\label{shi5}
	\mathcal{L}_{con} &= \mathbb{E}_{\textit{\textbf{I}}_{\rm{L}} \sim \boldsymbol{\mathcal{L}}}[||\textit{G}_{L}(\textit{G}_{H}(\textit{\textbf{I}}_{\rm{L}}))-\textit{\textbf{I}}_{\rm{L}}||_1]\\ &+ \mathbb{E}_{\textit{\textbf{I}}_{\rm{H}} \sim \boldsymbol{\mathcal{H}}}[||\textit{G}_{H}(\textit{G}_{L}(\textit{\textbf{I}}_{\rm{H}}))-\textit{\textbf{I}}_{\rm{H}}||_1].
\end{aligned}
\end{eqnarray}


\begin{figure}[t]
  \centering
  \begin{subfigure}{1\linewidth}
    \includegraphics[width=1.\linewidth]{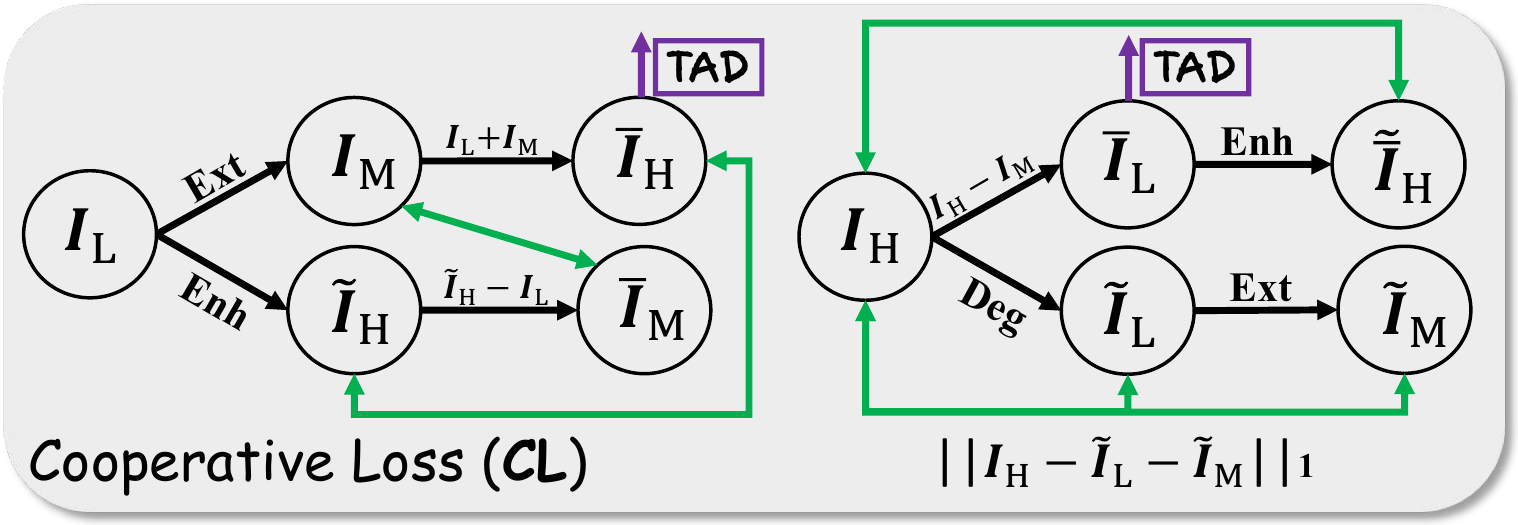}
  \end{subfigure}
  \vspace{-2.em}
  
  \caption{Details of the proposed cooperative loss function.}
  
  \label{loss}
  \vspace{-1.em}
  
\end{figure}

\begin{table*}[t]
	\centering
	\resizebox{\linewidth}{!}{
	\begin{tabular}{c|c|c c|c c c|c c c c c|c}
		\toprule[1.5pt]
		\multirow{2}{*}{Datasets} & \multirow{2}{*}{Metrics} & \multicolumn{2}{c|}{\cellcolor{gray!40}\emph{Model-based Methods}} & \multicolumn{3}{c|}{\cellcolor{gray!40}\emph{Supervised Learning Methods}} & \multicolumn{6}{c}{\cellcolor{gray!40}\emph{Unsupervised Learning Methods}} \\ \cline{3-13}
		& & \cellcolor{gray!40}LECARM & \cellcolor{gray!40}SDD & \cellcolor{gray!40}RetinexNet & \cellcolor{gray!40}KinD & \cellcolor{gray!40}URetinexNet & \cellcolor{gray!40}ZeroDCE & \cellcolor{gray!40}SSIENet & \cellcolor{gray!40}RUAS & \cellcolor{gray!40}EnGAN & \cellcolor{gray!40}SCI & \cellcolor{gray!40}Ours \\ \bottomrule
        \toprule
		\multirow{4}{*}{LOL \cite{LOL}} & PSNR $\uparrow$ & 14.41 & 13.34 & 16.77 & 17.65 & \textbf{\color{blue}19.54} & 14.80 & 19.50 & 16.40 & 17.48 & 14.78 & \textbf{\color{red}19.84} \\
		& SSIM $\uparrow$ & 0.5448 & 0.6342 & 0.4249 & 0.7614 & \textbf{\color{blue}0.7621} & 0.5607 & 0.7003 & 0.5034 & 0.6515 & 0.5254 & \textbf{\color{red}0.7713} \\ \cline{2-13}
		& NIQE $\downarrow$ & 12.34 & 13.77 & 12.51 & 14.81 & 11.39 & 12.62 & 15.89 & \textbf{\color{red}11.19} & 12.53 & 11.72 & \textbf{\color{blue}11.26} \\
		& LOE $\downarrow$ & 187.9 & 263.8 & 486.2 & 350.8 & 158.0 & 216.6 & 224.1 & 125.6 & 366.2 & \textbf{\color{red}102.3} & \textbf{\color{blue}117.7}  \\ \hline
		\multirow{4}{*}{LSRW \cite{LSRW}} & PSNR $\uparrow$ & 15.34 & 14.71 & 15.48 & 16.41 & \textbf{\color{blue}18.10} & 15.80 & 16.14 & 14.11 & 17.06 & 15.24 & \textbf{\color{red}19.00} \\
		& SSIM $\uparrow$ & 0.4212 & 0.4849 & 0.3468 & 0.4760 & \textbf{\color{blue}0.5149} & 0.4450 & 0.4627 & 0.4112 & 0.4601 & 0.4192 & \textbf{\color{red}0.5360} \\ \cline{2-13}
		& NIQE $\downarrow$ & 18.31 & 11.68 & 10.31 & 11.13 & 10.76 & 11.83 & 12.70 & 11.08 & 11.94 & \textbf{\color{blue}10.22} & \textbf{\color{red}9.23} \\
		& LOE $\downarrow$ & \textbf{\color{red}146.3} & 218.5 & 535.6 & 255.4 & 202.4 & 216.0 & 196.0 & 198.9 & 385.1 & 234.6 & \textbf{\color{blue}189.5} \\ \hline
		\multirow{2}{*}{LIME \cite{Guo_2017_TIP}} & NIQE $\downarrow$ & 12.80 & 15.21 & \textbf{\color{blue}11.88} & 14.72 & 14.48 & 12.85 & 16.16 & 12.44 & 14.59 & 12.38 & \textbf{\color{red}11.01} \\
		& LOE $\downarrow$ & 261.7 & 217.5 & 589.6 & 249.6 & \textbf{\color{red}166.7} & 192.1 & 216.6 & 288.7 & 421.1 & 212.6 & \textbf{\color{blue}187.2} \\ \bottomrule[1.5pt]
	\end{tabular}
	}
	\vspace{-1.em}
	
	\caption{\label{Quantitative} Comparison on three benchmarks. The best and the second best results are highlighted in \textbf{\color{red}red} and \textbf{\color{blue}blue} respectively.}
	\vspace{-1.em}
	
\end{table*}

\begin{figure*}[t]
  \centering
  \begin{subfigure}{0.162\linewidth}
    \includegraphics[width=1.\linewidth]{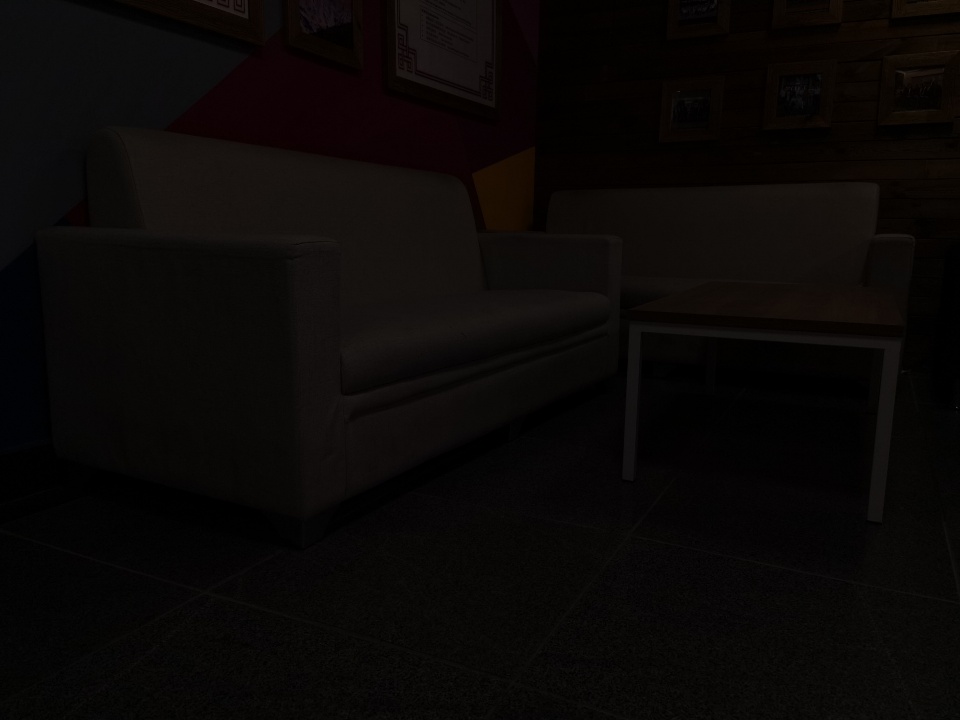}\vspace{-0.4em}
    \centerline{Input}\medskip
    
    \includegraphics[width=1.\linewidth]{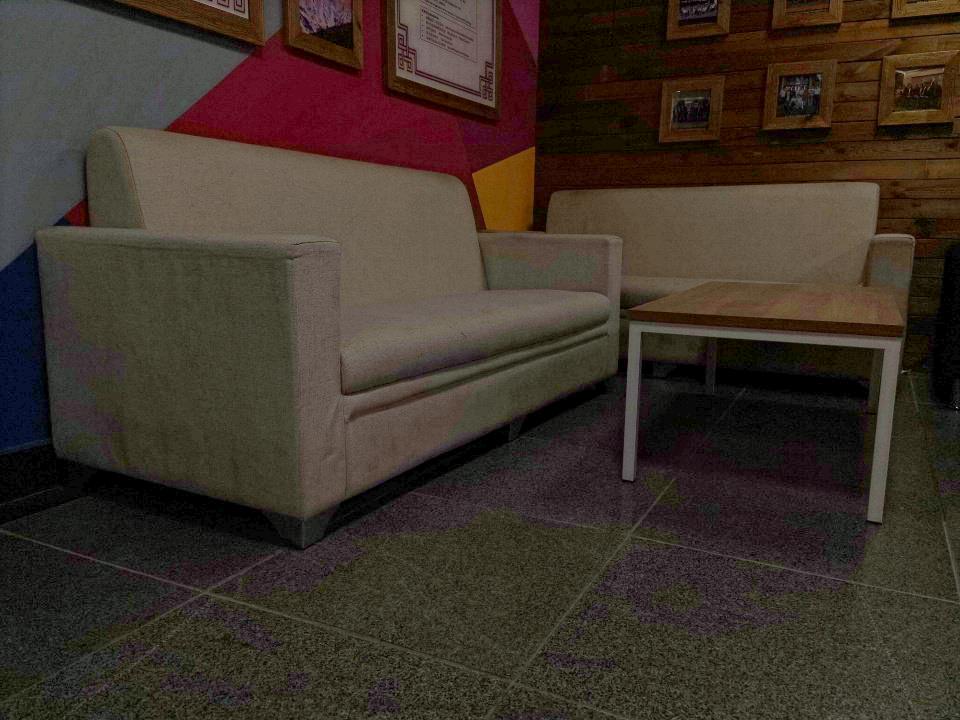}\vspace{-0.2em}
    \centerline{ZeroDCE}\medskip
  \end{subfigure}
  \hfill
  \begin{subfigure}{0.162\linewidth}
    \includegraphics[width=1.\linewidth]{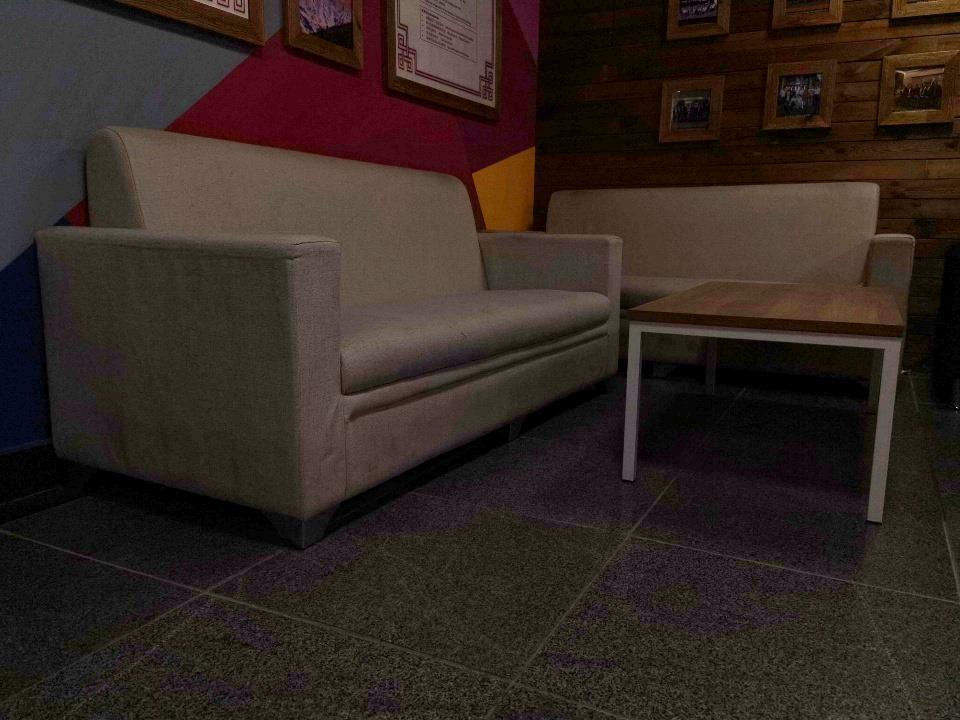}\vspace{-0.2em}
    \centerline{LECARM}\medskip
    
    \includegraphics[width=1.\linewidth]{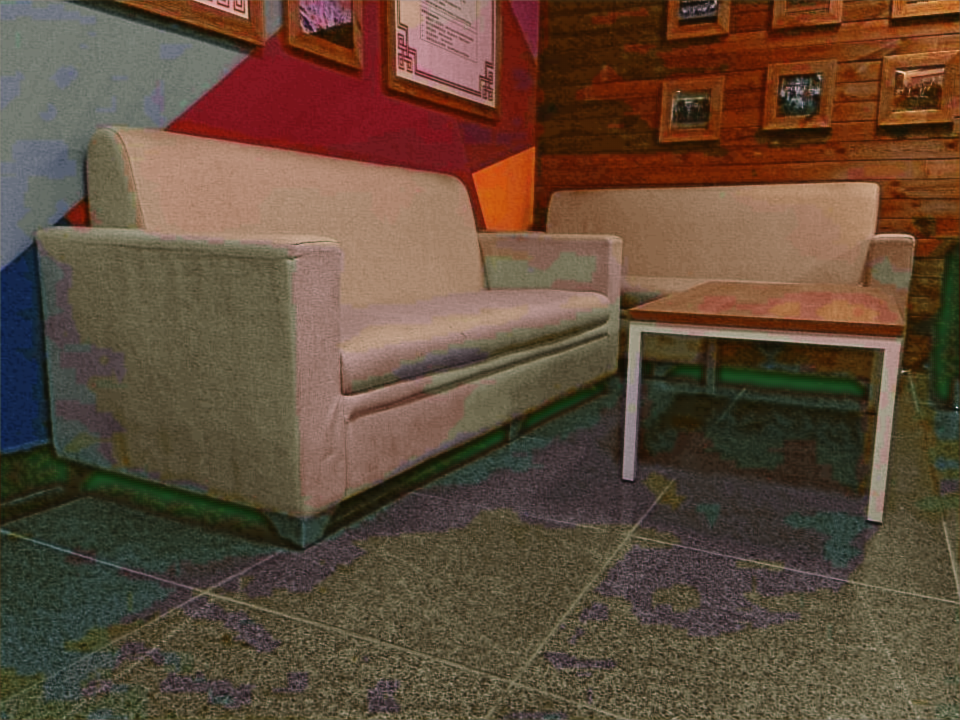}\vspace{-0.2em}
    \centerline{SSIENet}\medskip
  \end{subfigure}
  \hfill
  \begin{subfigure}{0.162\linewidth}
    \includegraphics[width=1.\linewidth]{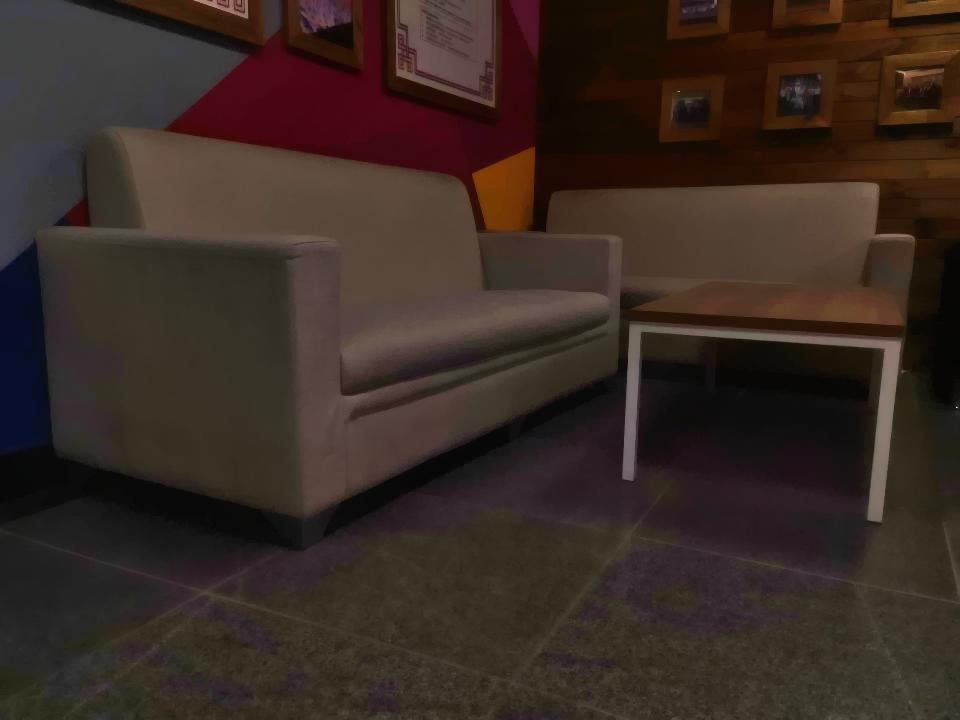}\vspace{-0.2em}
    \centerline{SDD}\medskip
    
    \includegraphics[width=1.\linewidth]{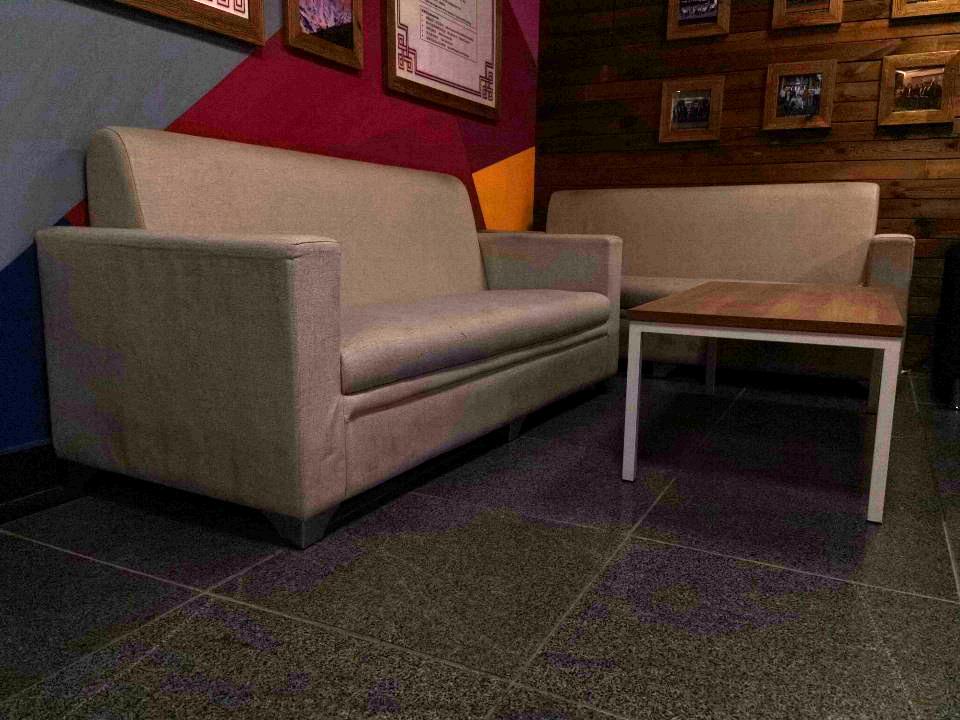}\vspace{-0.2em}
    \centerline{RUAS}\medskip
  \end{subfigure}
  \hfill
  \begin{subfigure}{0.162\linewidth}
    \includegraphics[width=1.\linewidth]{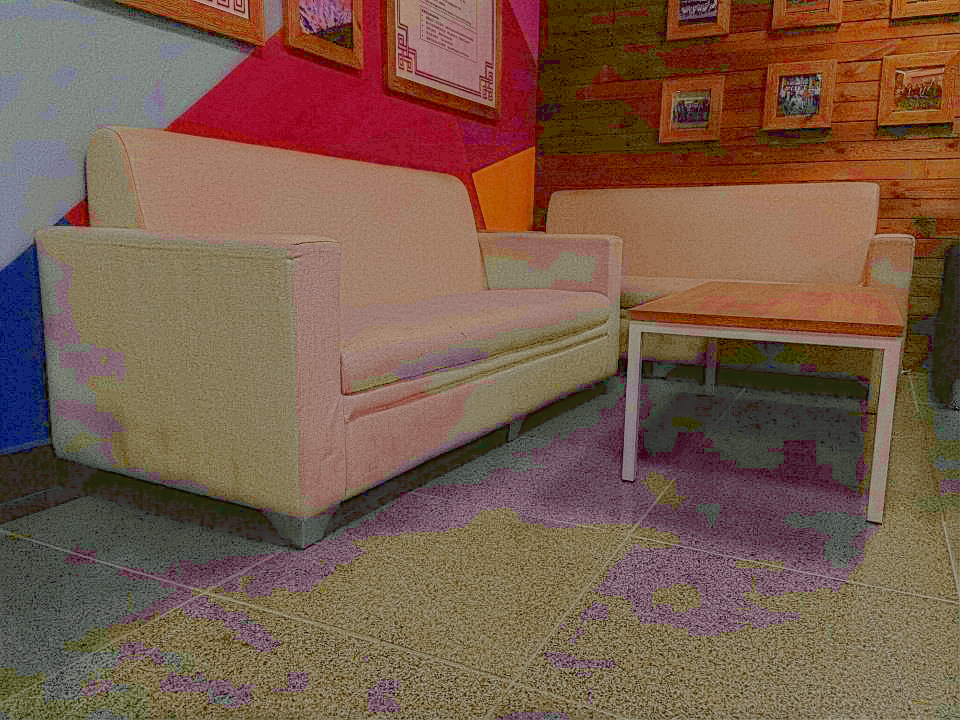}\vspace{-0.2em}
    \centerline{RetinexNet}\medskip
    
    \includegraphics[width=1.\linewidth]{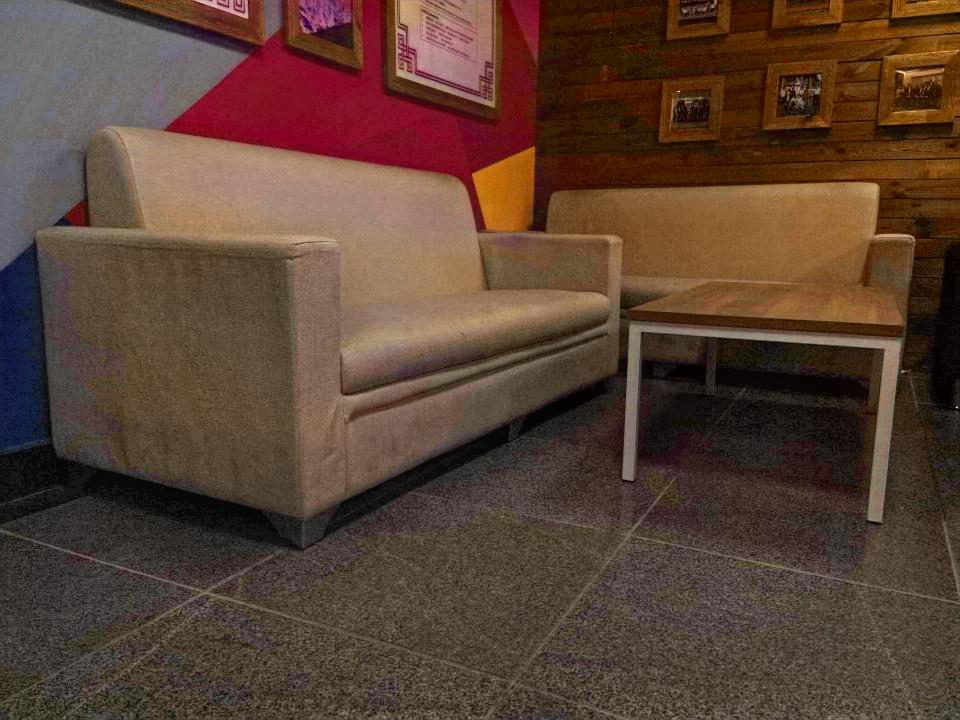}\vspace{-0.2em}
    \centerline{EnGAN}\medskip
  \end{subfigure}
  \hfill
  \begin{subfigure}{0.162\linewidth}
    \includegraphics[width=1.\linewidth]{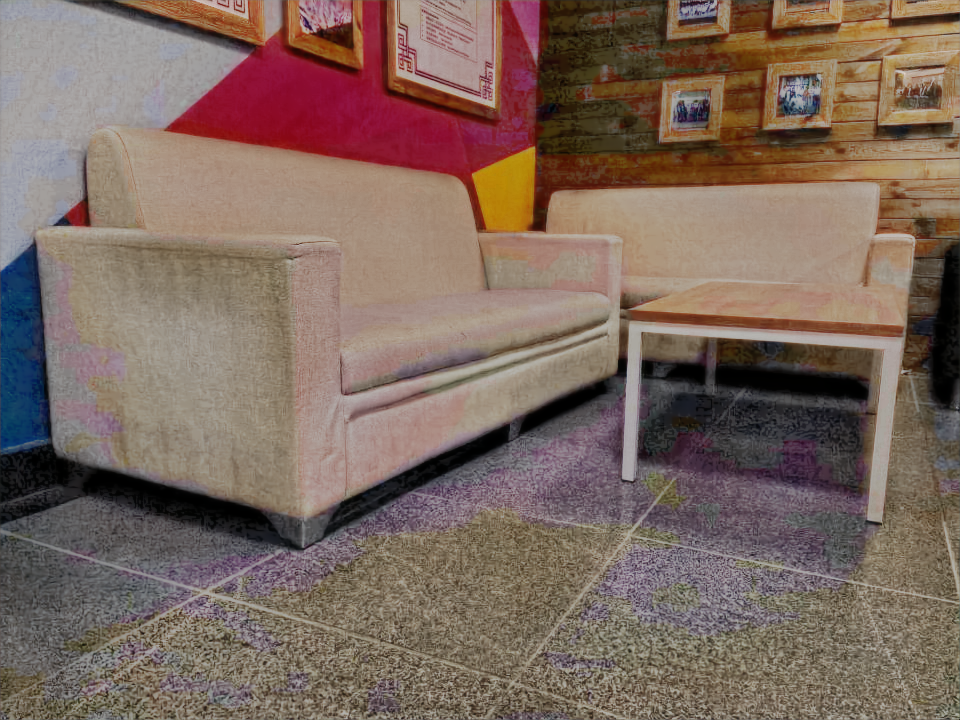}\vspace{-0.2em}
    \centerline{KinD}\medskip
    
    \includegraphics[width=1.\linewidth]{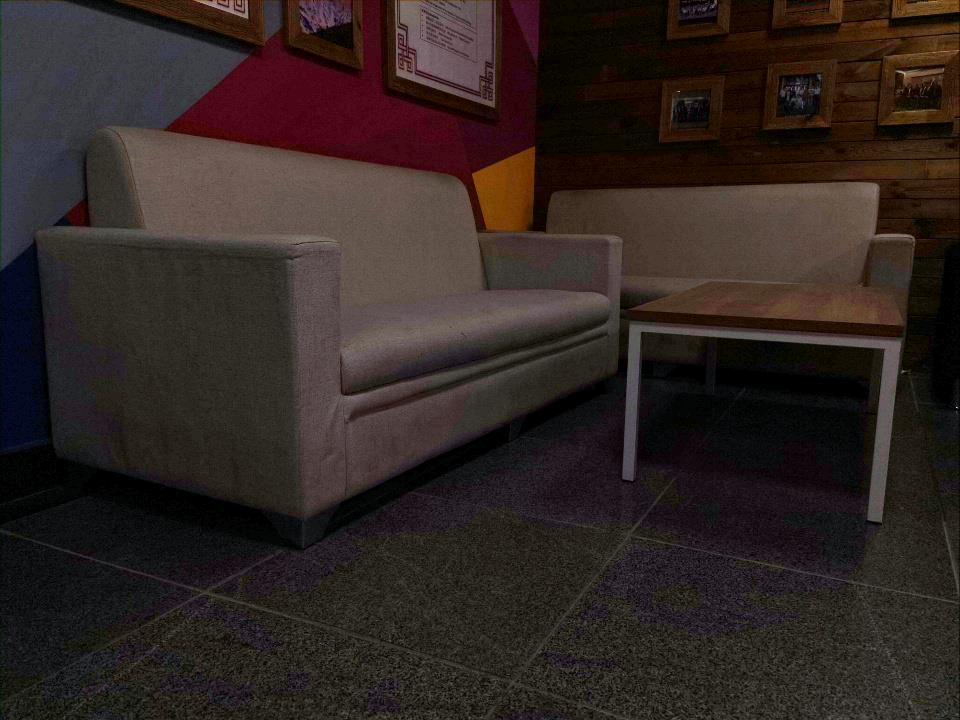}\vspace{-0.2em}
    \centerline{SCI}\medskip
  \end{subfigure}
  \begin{subfigure}{0.162\linewidth}
    \includegraphics[width=1.\linewidth]{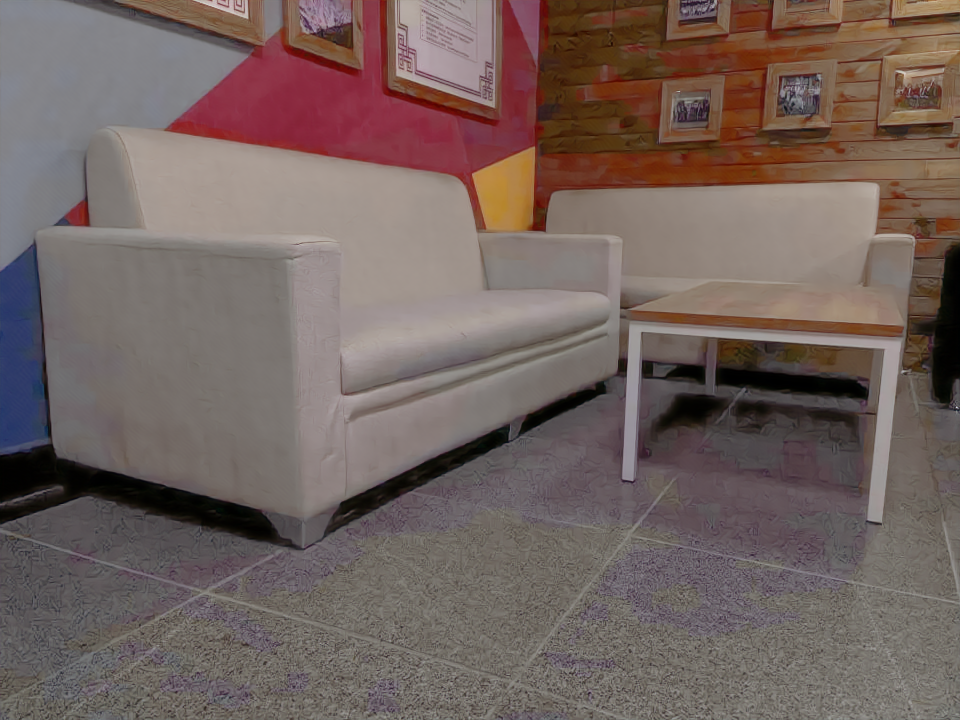}\vspace{-0.2em}
    \centerline{URetinexNet}\medskip
    
    \includegraphics[width=1.\linewidth]{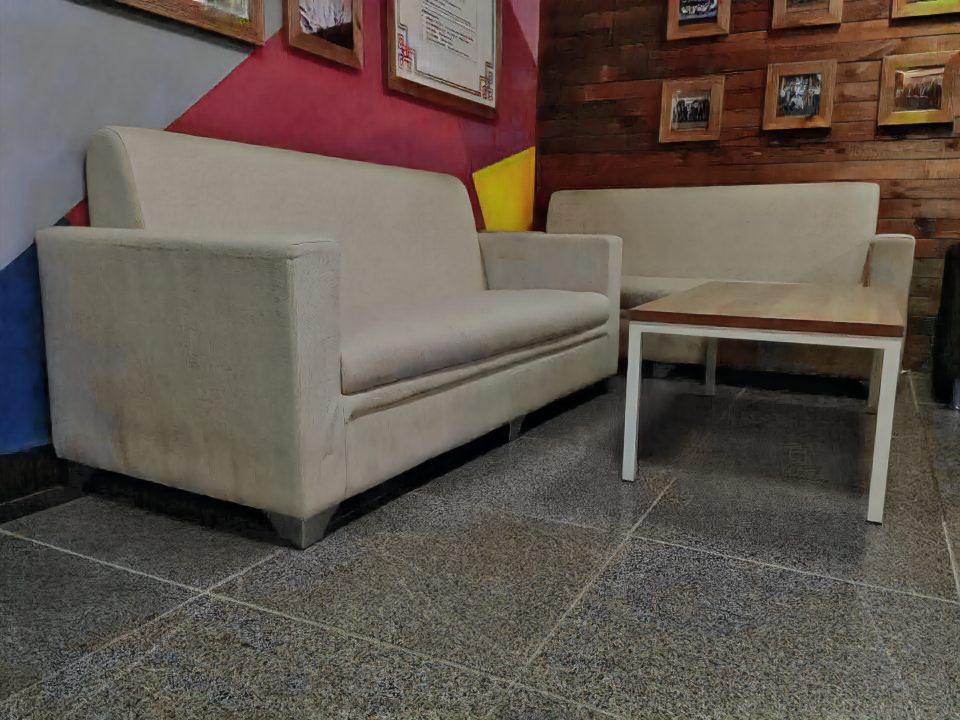}\vspace{-0.2em}
    \centerline{Ours}\medskip
  \end{subfigure}
  \vspace{-1.em}

  \caption{Subjective comparison on the LSRW dataset among state-of-the-art low-light image enhancement algorithms.}
  \label{comlsrw}
  \vspace{-1.em}
\end{figure*}

\textbf{Cooperative Loss.} To reduce solution space and enhance attention guidance, we elaborate Cooperative Loss (CL). Inspired by Lee~\textit{et al.} \cite{Lee_2021_TMI}, this function trains different modules cooperatively by imposing mutual constraints. As shown in \cref{loss}, we indirectly supervise attention guidance $\textit{\textbf{I}}_{\rm{A}}$ and provide stronger constraints for all modules.

Concretely, as shown in Mask Extractor (ME) of \cref{totnet}, since attention map $\textit{\textbf{I}}_{\rm{A}}$ is generated based on the extracted image features, the content of features heavily impacts the quality of $\textit{\textbf{I}}_{\rm{A}}$. To get better $\textit{\textbf{I}}_{\rm{A}}$, we generate a \textbf{lightness mask} $\textit{\textbf{I}}_{\rm{M}}$ from the same features and co-supervise it with other information, including our enhanced image.

As shown in the left end of \cref{loss}, for the low-light input $\textit{\textbf{I}}_{\rm{L}}$, on the one hand, we extract $\textit{\textbf{I}}_{\rm{M}}$ with ME, and obtain a pseudo-high-light image $\overline{\textbf{\textit{I}}}_{\rm{H}}$ by adding $\textit{\textbf{I}}_{\rm{M}}$ with $\textit{\textbf{I}}_{\rm{L}}$. On the other hand, we subtract $\textit{\textbf{I}}_{\rm{L}}$ at the pixel level from the predicted high-light result $\widetilde{\textbf{\textit{I}}}_{\rm{H}}$, obtaining a pseudo-lightness mask $\overline{\textit{\textbf{I}}}_{\rm{M}}$. By using consistency loss, we encourage the estimated $\overline{\textit{\textbf{I}}}_{\rm{M}}$ to be consistent with the extracted $\textit{\textbf{I}}_{\rm{M}}$, the calculated $\overline{\textbf{\textit{I}}}_{\rm{H}}$ to be similar to the depicted $\widetilde{\textbf{\textit{I}}}_{\rm{H}}$, expressed as:
\begin{eqnarray}
\begin{aligned}
	\label{shi6}
	\mathcal{L}_{\rm{rec}1} = ||\overline{\textit{\textbf{I}}}_{\rm{M}}-\textit{\textbf{I}}_{\rm{M}}||_1 + ||\overline{\textbf{\textit{I}}}_{\rm{H}}-\widetilde{\textbf{\textit{I}}}_{\rm{H}}||_1.
\end{aligned}
\end{eqnarray}

For another branch of given the high-light input $\textit{\textbf{I}}_{\rm{H}}$, as shown in the right part of \cref{loss}, the loss is similar:
\begin{eqnarray}
\begin{aligned}
	\label{shi7}
	\mathcal{L}_{\rm{rec}2} = ||\widetilde{\overline{\textbf{\textit{I}}}}_{\rm{H}}-\textbf{\textit{I}}_{\rm{H}}||_1 + ||\textbf{\textit{I}}_{\rm{H}}-\widetilde{\textbf{\textit{I}}}_{\rm{L}}-\widetilde{\textit{\textbf{I}}}_{\rm{M}}||_1.
\end{aligned}
\end{eqnarray}

For the calculated images based on $\textit{\textbf{I}}_{\rm{M}}$, \textit{i.e.}, $\overline{\textbf{\textit{I}}}_{\rm{H}}$ and $\overline{\textbf{\textit{I}}}_{\rm{L}}$, we further use the double-path discriminator from our TAD to inspect their authenticity, formulated as:
\begin{eqnarray}
\begin{aligned}
	\label{shi8}
	\mathcal{L}_{\rm{insp}}(\overline{\textit{\textbf{I}}}_{\rm{H}},\overline{\textbf{\textit{I}}}_{\rm{L}},\textit{D}_{H},\textit{D}_{L}) &= \mathbb{E}_{\overline{\textit{\textbf{I}}}_{\rm{H}} \sim \overline{\boldsymbol{\mathcal{H}}}}[\log(1-\textit{D}_{H}(\overline{\textbf{\textit{I}}}_{\rm{H}}))]\\ &+ \mathbb{E}_{\overline{\textbf{\textit{I}}}_{\rm{L}} \sim \overline{\boldsymbol{\mathcal{L}}}}[\log(1-\textit{D}_{L}(\overline{\textbf{\textit{I}}}_{\rm{L}}))],
\end{aligned}
\end{eqnarray}
where $\overline{\boldsymbol{\mathcal{H}}}$ and $\overline{\boldsymbol{\mathcal{L}}}$ mean the pseudo high-light and low-light image domain respectively. The final cooperative loss is:
\begin{eqnarray}
\begin{aligned}
	\label{shi9}
	\mathcal{L}_{CL} = \mathcal{L}_{\rm{rec}1} + \mathcal{L}_{\rm{rec}2} + \mathcal{L}_{\rm{insp}}.
\end{aligned}
\end{eqnarray}

Combined with the NR-related loss in \cref{sec:NR}, the adversarial constraint in \cref{sec:TAD}, and the consistency loss in \cref{sec:DLGP}, our final objective function is expressed as:
\begin{eqnarray}
\begin{aligned}
	\label{shi12}
	\mathcal{L} = \mathcal{L}_{NR} + \mathcal{L}_{adv} + \mathcal{L}_{con} + \mathcal{L}_{CL}.
\end{aligned}
\end{eqnarray}

\section{Experiments}
\label{sec:Exp}
In this section, we first present the implementation details of our approach. Then we compare it with the state-of-the-art methods through multiple benchmarks. To identify the contribution of each component, we further conduct ablation analyses. All experiments are implemented with PyTorch and conducted on a single NVIDIA Tesla V100 GPU.


\subsection{Implementation Details}
\textbf{Parameter Settings.} For training, we adopt Adam optimizer \cite{Adam} with the hyper parameters $\beta_1$ = 0.5, $\beta_2$ = 0.999, and $\epsilon$ = 10$^{-8}$. Our model is trained for 300 epochs with an initial learning rate of 2$\times$10$^{-4}$ and decaying linearly to 0 in the last 200 epochs. Batch size is set to be 1 and patch size is resized to 256$\times$256 for training in the concern of efficiency. Heuristically, we adopt the MLP with 3 hidden layers to normalize the degradation level.

\textbf{Benchmarks and Metrics.} To validate the effectiveness of our method, we train and test the model on LSRW dataset \cite{LSRW}, which contains 1000 low-light-normal-light image pairs for training and 50 pairs for evaluation. Each pair consists of a degraded image and a well-exposed reference, which are captured from real world with different exposure times. For a more convincing comparison, we further extend evaluation to other benchmarks such as LOL \cite{LOL} and LIME \cite{Guo_2017_TIP}. As LOLv1 only contains 15 images for evaluation, we randomly sample 35 images from the test set of LOLv2 (not used for training) and evaluate on these 50 images. To demonstrate the generalization to real-world degradation scenarios, we test on LIME with the model trained on LSRW. Noting that to prove the superiority of our unsupervised learning manner, during training, we only adopt the low-light part of the paired training data, and replace the references with 300 images from BSD300 dataset \cite{Martin_2001_ICCV}. We use two full-reference metrics, \textit{i.e.}, Peak Signal-to-Noise Ratio (PSNR) and Structural Similarity Index (SSIM) \cite{Zhou_2004_TIP}, and two no-reference metrics, namely NIQE \cite{Mittal_2013_SPL} and LOE \cite{Wang_2013_TIP}, to evaluate the effectiveness of different algorithms objectively. In general, a higher PSNR or SSIM means more authentic restored results, while a lower NIQE or LOE represents higher quality details, lightness, and tone.


\begin{figure}[t]
  \centering
  \begin{subfigure}{0.32\linewidth}
    \includegraphics[width=1.\linewidth]{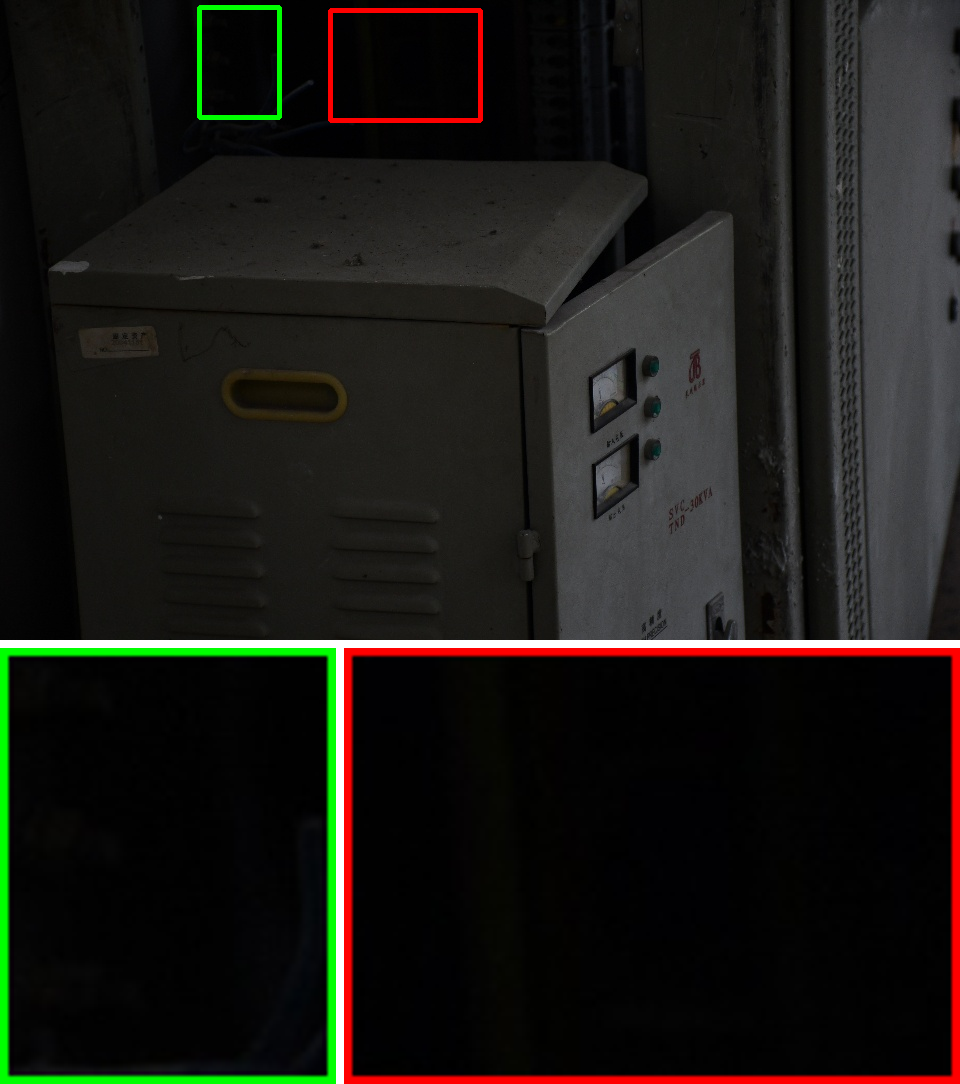}\vspace{-0.5em}
    \centerline{Input}\vspace{-0.6em}\medskip
    
    \includegraphics[width=1.\linewidth]{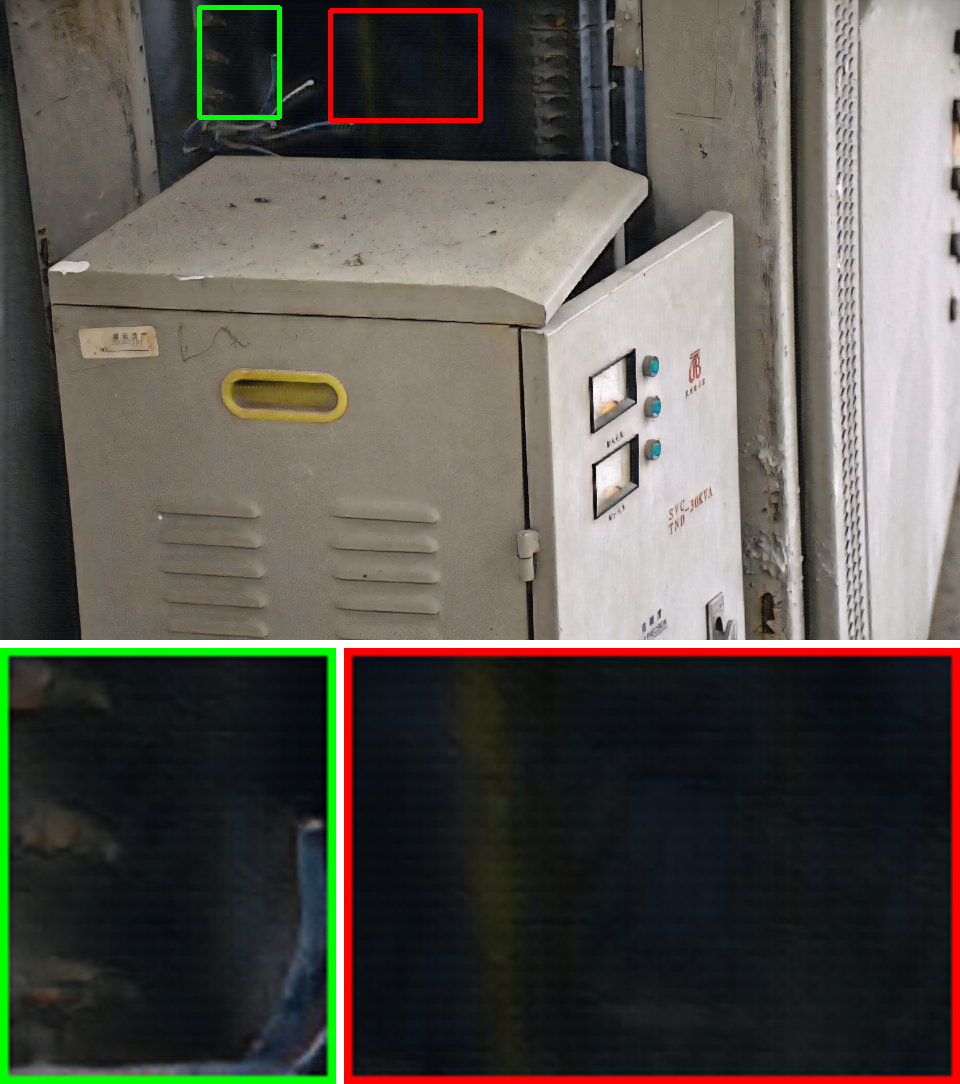}\vspace{-0.4em}
    \centerline{\#3}\vspace{-0.2em}\medskip
  \end{subfigure}
  \hfill
  \begin{subfigure}{0.32\linewidth}
    \includegraphics[width=1.\linewidth]{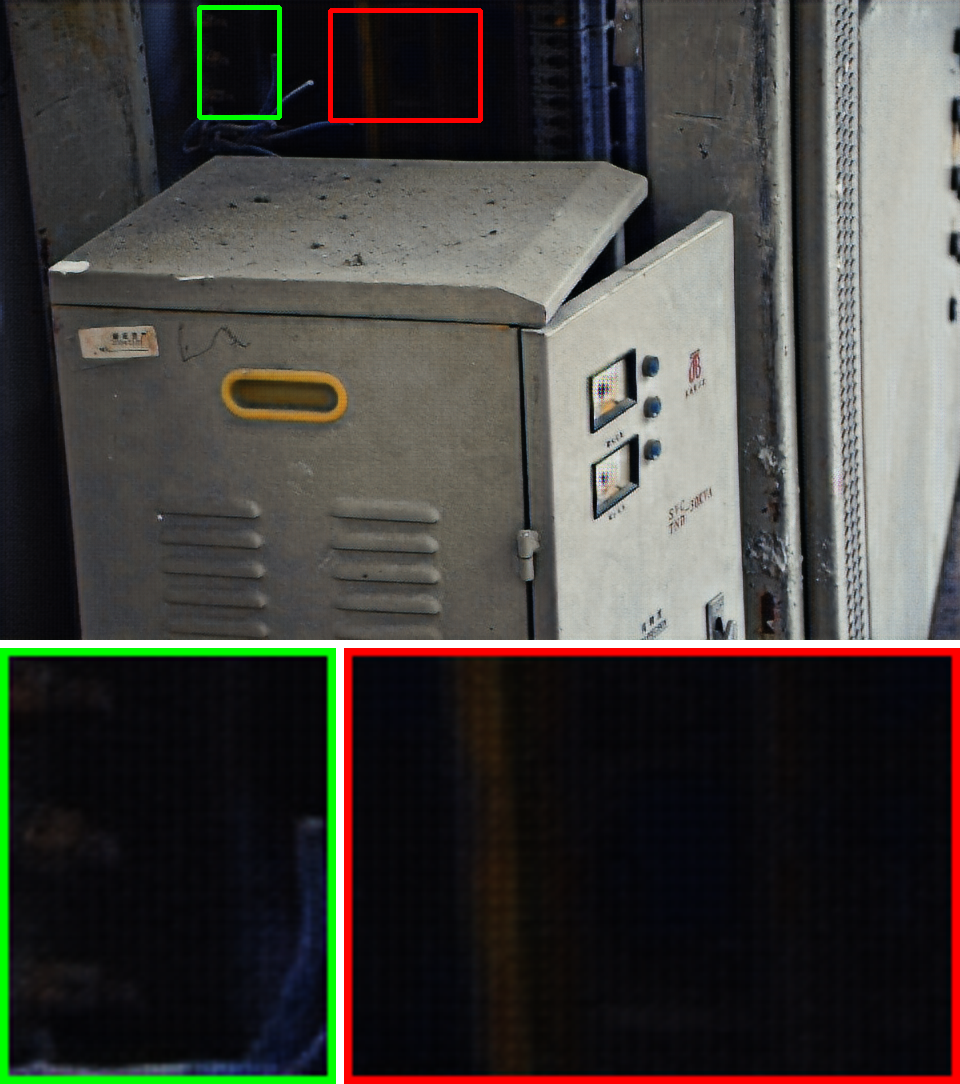}\vspace{-0.3em}
    \centerline{\#1}\vspace{-0.6em}\medskip
    
    \includegraphics[width=1.\linewidth]{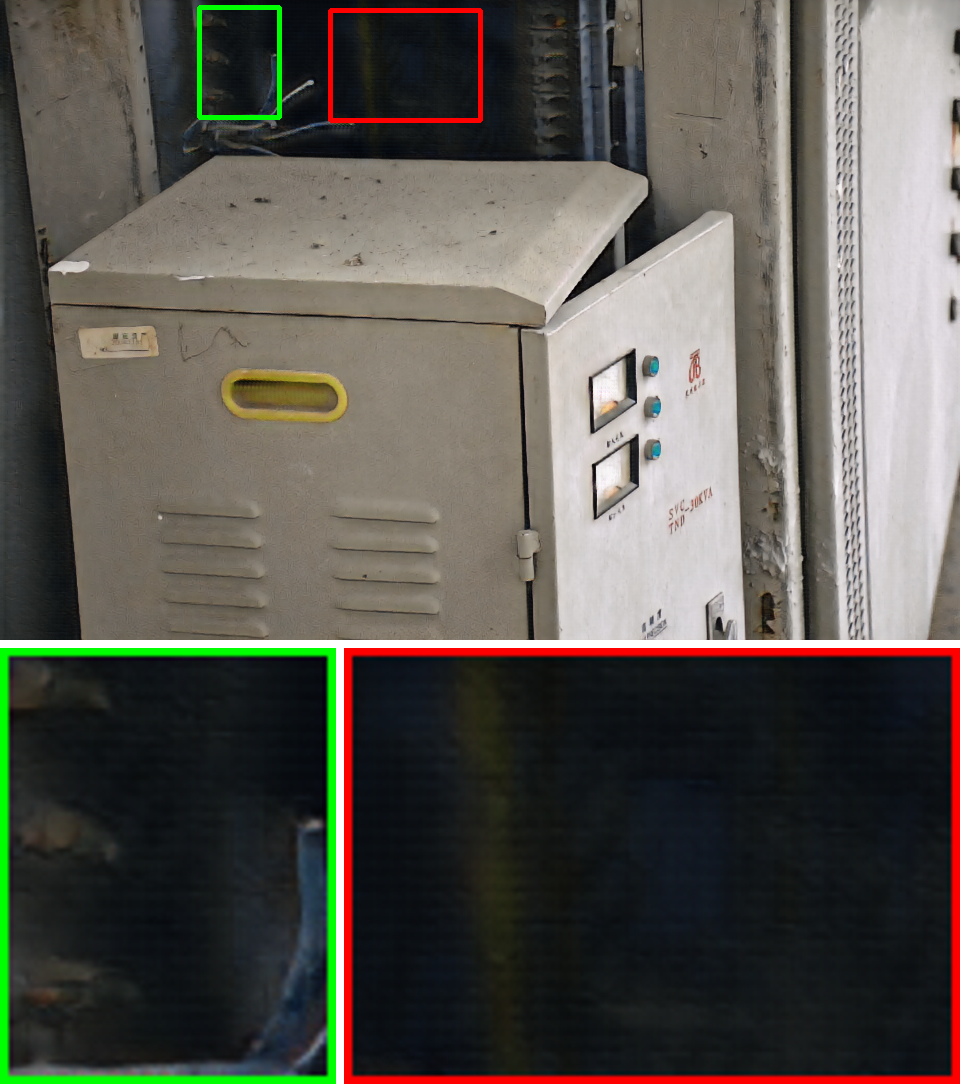}\vspace{-0.4em}
    \centerline{\#4}\vspace{-0.2em}\medskip
  \end{subfigure}
  \hfill
  \begin{subfigure}{0.32\linewidth}
    \includegraphics[width=1.\linewidth]{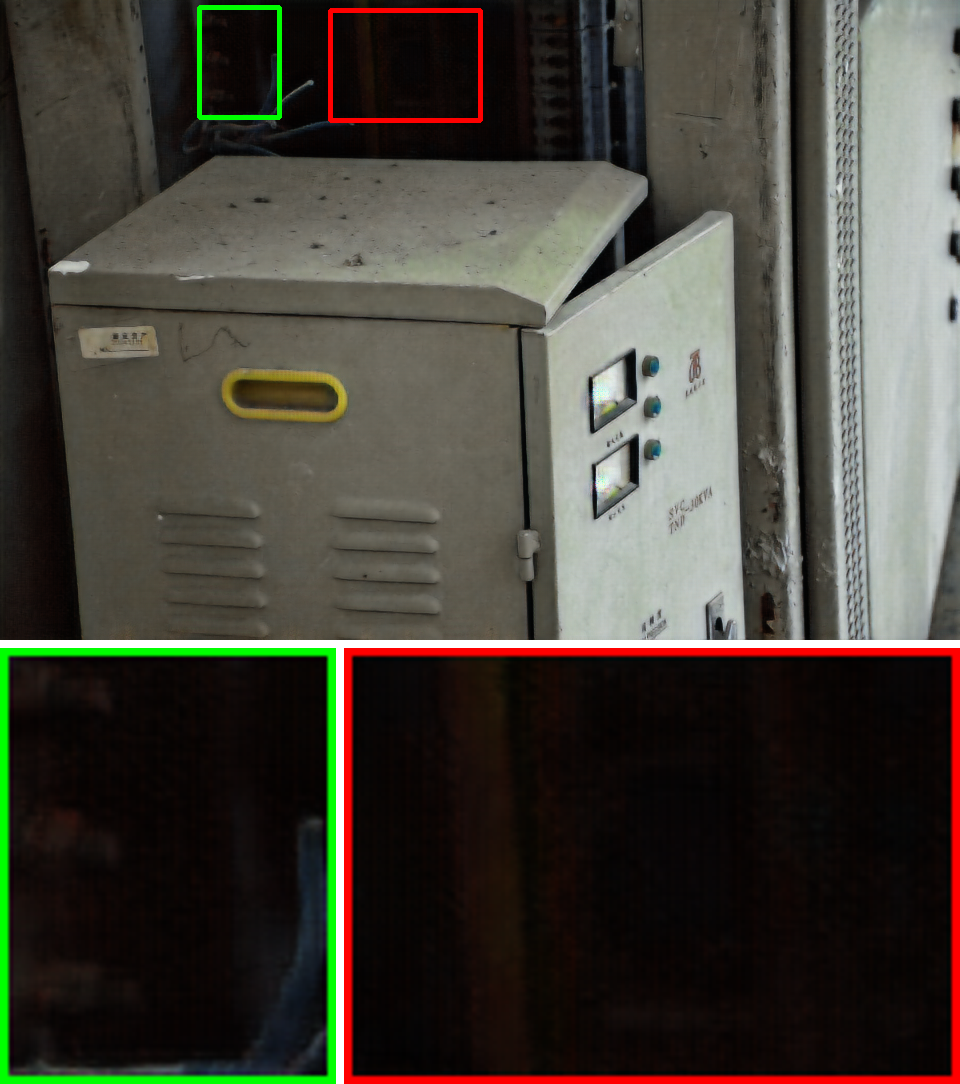}\vspace{-0.3em}
    \centerline{\#2}\vspace{-0.6em}\medskip
    
    \includegraphics[width=1.\linewidth]{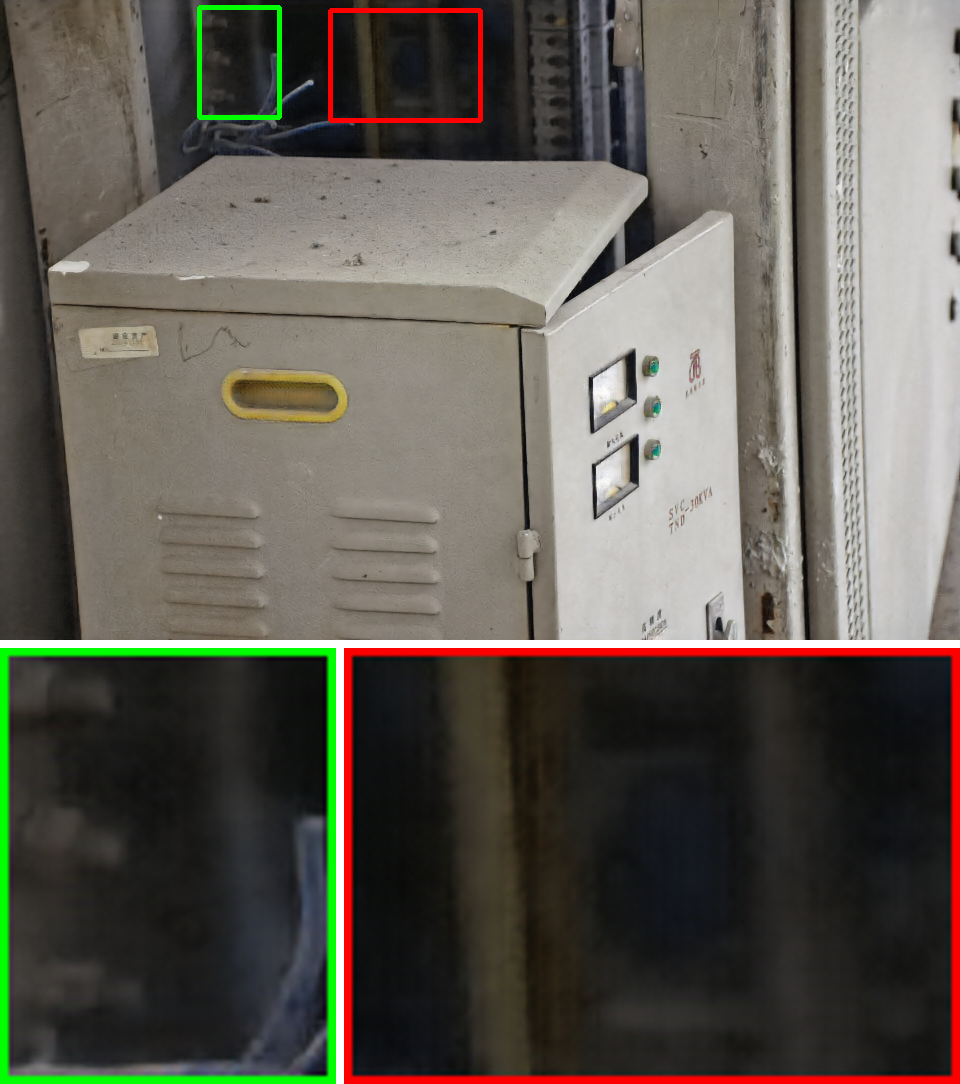}\vspace{-0.4em}
    \centerline{NeRCo}\vspace{-0.2em}\medskip
  \end{subfigure}
  \vspace{-1.3em}

  \caption{Visual results of ablation study. The full set performs best, especially in the regions boxed in green and red.}
  \label{ablat}
  \vspace{-1.2em}
  
\end{figure}

\subsection{Comparison with the State-of-the-Art}
For a more comprehensive analysis, we compare our method with two recently-proposed model-based methods, namely LECARM \cite{LECARM} and SDD \cite{SDD}, three advanced supervised learning methods (\textit{i.e.}, RetinexNet \cite{LOL}, KinD \cite{Zhang_2019_MM}, and URetinex-Net \cite{Wu_2022_CVPR}), and five unsupervised learning methods, including ZeroDCE \cite{Guo_2020_CVPR}, SSIENet \cite{SSIENet}, RUAS \cite{Liu_2021_CVPR}, EnGAN \cite{Jiang_2021_TIP}, and SCI \cite{Ma_2022_CVPR}.

\textbf{Quantitative analysis.} We obtain quantitative results of other methods by adopting official pre-trained models and running their respective public codes. As shown in \cref{Quantitative}, our method achieves nearly SOTA performance in both full-reference and no-reference metrics across all benchmarks. It validates the superior effect of the proposed framework. Noting that our method even performs better than the supervised ones. Compared with some recent competitive unsupervised approaches such as EnGAN \cite{Jiang_2021_TIP}, and SCI \cite{Ma_2022_CVPR}. We provide stronger constraints than EnGAN, including visual-oriented guidance from prompts. Besides, due to the normalization property, our method outperforms SCI especially on some challenging scenes. The visual results of all methods (including NeRCo and SCI) in a scenario with severe brightness degradation are given below.

\textbf{Qualitative analysis.} For a more intuitive comparison, we report the visual results of all approaches in \cref{comlsrw}. Our input is a severely degraded image. One can see that the recent traditional methods can not recover enough brightness. While the advanced deep learning-based methods over-smooth background or introduce unknown veils, resulting in miserable artifacts and unnatural tone. In particular, SCI fails to effectively enhance such a challenging scenario. By comparison, our model realizes the best visual quality with prominent contrast and vivid colors. Details are also remained well. More results are presented in \textcolor{red}{SM}.

	
	

\subsection{Ablation Study}
\label{sec:AB}
As shown in \cref{blatq}, we consider four ablation settings by adding the proposed components to the dual loop successively. All ablation studies are conducted on the LSRW dataset. \textbf{i)} ``\#1'' is a naive dual loop without any other operations, which only adopts a vanilla color discriminator. This framework has achieved comparable scores, indicating its effectiveness. \textbf{ii)} ``\#2'' adds the mask extractor (ME) and the related cooperative loss (CL) to ``\#1''. ME provides attention guidance and CL reduces solution space. Note that ``\#2'' further adopts edge-path discriminator but without text supervision. \textbf{iii)} ``\#3'' adds TAD to ``\#2'', which employs visual-oriented guidance from textual modality. Further ablation study on TAD will be given in \textcolor{red}{SM}. \textbf{iv)} ``\#4'' adds NRN to  ``\#2'' but without TAD, which is designed to compare the performance gain of TAD and NRN. \textbf{v)} Finally, we adopt a complete NeRCo. Due to the improved robustness to different degradation conditions and visual-friendly guidance, this setting achieves apparent performance gain.

We report qualitative results of all settings in \cref{ablat}. One can see that in the first sample, ``\#1'' increases contrast and roughly recovers objects in dark region, but the enhanced tone is still inauthentic. ``\#2'' relieves color cast phenomenon to a certain degree but still remains undesired veils. Both ``\#3'' and ``\#4'' generate cleaner results with realistic lightness. Further, the result of NeRCo faithfully preserves the most details and performs the best perceptual effectiveness, especially in the regions boxed in green and red. We attribute this to the neural representation normalization and text-driven appearance discriminator, the former unifies degradation level and reduces the difficulty of enhancement task, while the latter guides visual-friendly optimization.



\begin{table}[t]
	\centering
	\resizebox{\linewidth}{!}{
	\begin{tabular}{c|c c c|c|c|c|c}
		\toprule[1.pt]
		\cellcolor{gray!40}index & \cellcolor{gray!40}NRN & \cellcolor{gray!40}TAD & \cellcolor{gray!40}CL\&ME & \cellcolor{gray!40}PSNR $\uparrow$ & \cellcolor{gray!40}SSIM $\uparrow$ & \cellcolor{gray!40}NIQE $\downarrow$ & \cellcolor{gray!40}LOE $\downarrow$ \\ \bottomrule
		\toprule
		\#1 & \XSolidBrush & \XSolidBrush & \XSolidBrush & 16.77 & 0.4565 & 12.34 & 272.4  \\ \cline{5-8}
		\#2 & \XSolidBrush & \XSolidBrush & \CheckmarkBold & 17.65 & 0.5023 & 10.60 & 247.9 \\ \cline{5-8}
		\#3 & \XSolidBrush & \CheckmarkBold & \CheckmarkBold & 18.32 & 0.5201 & 10.83 & 230.9 \\ \cline{5-8}
  		\#4 & \CheckmarkBold & \XSolidBrush & \CheckmarkBold & \textbf{\color{blue}18.62} & \textbf{\color{blue}0.5239} & \textbf{\color{blue}9.63} & \textbf{\color{blue}218.8} \\ \cline{5-8}
		NeRCo & \CheckmarkBold & \CheckmarkBold & \CheckmarkBold & \textbf{\color{red}19.00} & \textbf{\color{red}0.5360} & \textbf{\color{red}9.23} & \textbf{\color{red}189.5} \\ \bottomrule[1.pt]
	\end{tabular}
	}
	\vspace{-1.em}
	
	\caption{\label{blatq} Quantitative evaluation on the enhanced results obtained from different settings. The best and the second best results are highlighted in \textbf{\color{red}red} and \textbf{\color{blue}blue} respectively.}
	\vspace{-1.em}
	
\end{table}

\section{Conclusion}
We proposed an implicit Neural Representation method for Cooperative low-light image enhancement, dubbed NeRCo, to recover visual-friendly results in an unsupervised manner. Firstly, for the input degraded image, we employed neural representation to normalize degradation levels (\textit{e.g.}, dark lightness and natural noise). Besides, for the output enhanced image, we equipped the discriminator with a high-frequency path and utilized priors from the pre-trained vision-language model to impart perceptual-oriented guidance. Finally, to ease the reliance on paired data and enhance low-light scenarios in a self-supervised manner, a dual-closed-loop cooperative constraint was developed to train the enhancement module. It encourages all components to constrain each other, further reducing solution space. Experiments proved the superiority of our method compared with other top-performing ones. The proposed components provide valuable inspiration for other low-level tasks, such as image dehazing \cite{Ye_2022_ECCV}, compressive sensing \cite{Li_2022_arxiV}, and hyperspectral imaging \cite{Zhang_2023_arxiV,Zhang_2022_CVPR}.

{\small
\bibliographystyle{ieee_fullname}
\bibliography{egbib}
}

\clearpage
\appendix
\onecolumn
\onecolumn
\begin{center}
\textbf{\Large{Implicit Neural Representation for Cooperative Low-light Image Enhancement \\ -- Supplemental Document --}}
\vspace{1cm}
\centering
\captionsetup{type=figure}
\def\svgwidth{1.0\linewidth}
\end{center}

\begin{abstract}

This is the supplementary material for the paper: ``Implicit Neural Representation for Cooperative Low-light Image Enhancement''. Firstly, we provide more results of our NRN module in \textbf{Section A} for a comprehensive illustration. Besides, in \textbf{Section B}, we train the framework with different prompts and compare their performance with other ablation settings, which verifies the effect of text-driven supervision. In \textbf{Section C}, we further conduct ablation experiments on TAD to determine the role of each path. In \textbf{Section D}, to demonstrate the semantic advantage of our method, we classify the enhanced results of different methods with a pre-trained vision-language model and report their accuracy. Our results are considered best for textual description of high-light image. Finally, more qualitative analyses on three well-known benchmarks are displayed in \textbf{Section E}, including LSRW dataset, LOL dataset and LIME dataset. It is obvious that the proposed NeRCo achieves the best performance, further verifying our superiority.
\end{abstract}

\section{Normalized Results}
Deep learning based models learn to map a sample from the input domain to the target domain. In real-world application, however, degradation conditions are various. For some inputs far from the learned input domain, it is hard for a trained model to perform stable superior performance. Hence, we developed Neural Representation Normalization (NRN) module to normalize different conditions, which has been illustrated in \cref{sec:NR}. In order to provide more convincing proof, here we add here more experimental results. 

\begin{figure*}[t]
  \centering
  \begin{subfigure}{0.98\linewidth}
    \includegraphics[width=1.\linewidth]{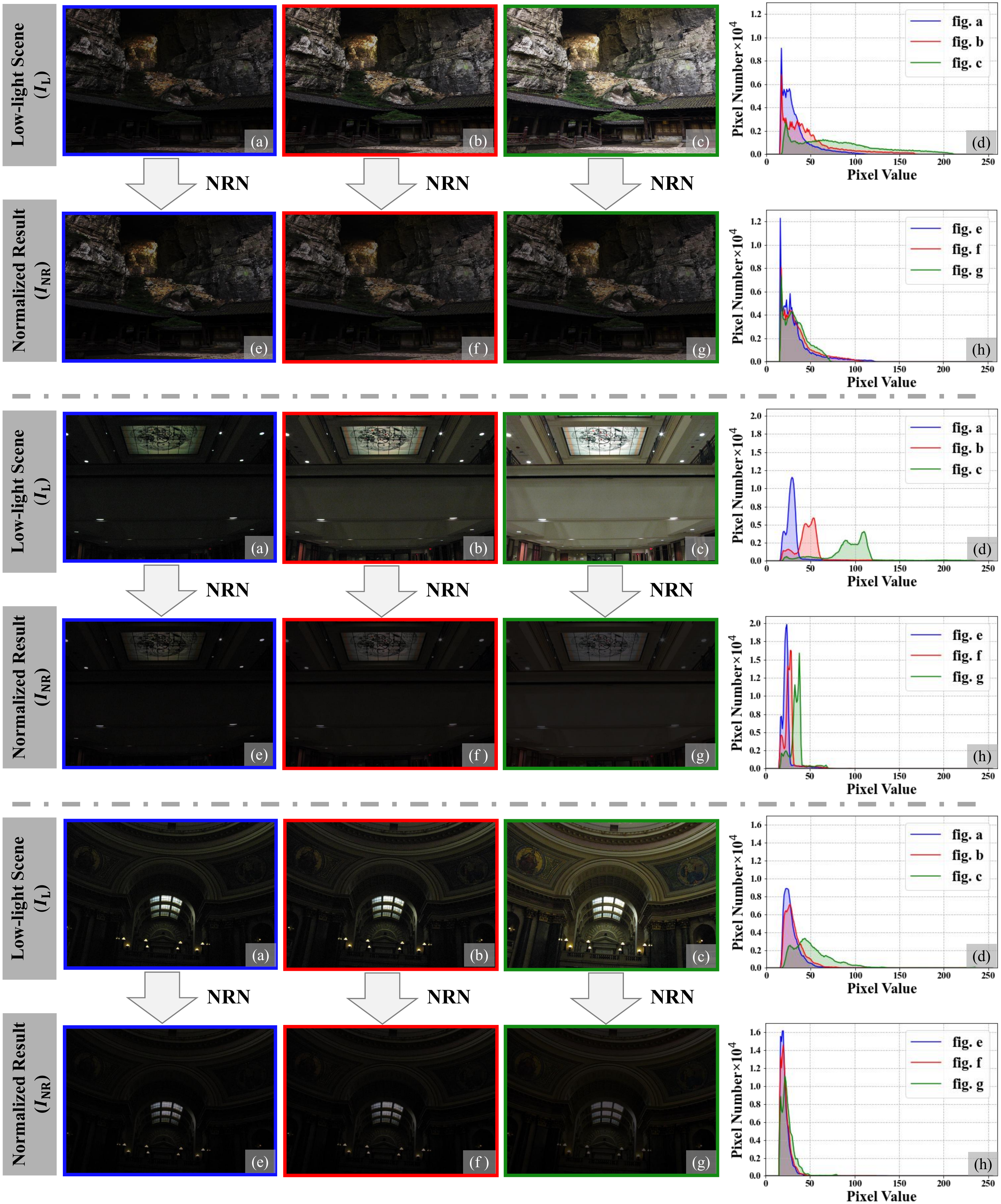}
    
  \end{subfigure}
  \vspace{-0.7em}

  \caption{Comparisons between the captured low-light scenes ($\textit{\textbf{I}}_{\rm{L}}$) and the results of NRN ($\textit{\textbf{I}}_{\rm{NR}}$). The low-light samples are from the SICE \cite{Cai_2018_TIP} dataset. It contains numerous image sets, each with common content and varying degradation conditions. The pixel value distribution of images on the Y channel is given on the right. One can see that NRN normalizes the brightness to be similar.}
  \label{NRPP1}
  \vspace{-1.em}
  
\end{figure*}

As shown in \cref{NRPP1}, we adopted three sets of images from the SICE \cite{Cai_2018_TIP} dataset, each set contains three images with different brightness and the same content. We box them with blue, red and green lines respectively. All of them are processed by our NRN module, the corresponding results are boxed with the same color. One can see that the brightness of original inputs varies a lot, while the output brightness of NRN is similar. For more intuitive, on the right, we provide a visualization of their pixel distribution on the Y channel. It is obvious that NRN constricts the the range of brightness changes and normalizes degradation levels.

\section{Experiments with Alternative Textual Prompts}
To investigate the contribution of our proposed text-driven supervision, we compare the performance of models trained on different prompts. Specifically, we consider three pairs of alternative prompts to guide model training: i) $\textit{dark}$ and $\textit{bright}$. ii) $\textit{dim}$ and $\textit{light}$. iii) $\textit{night}$ and $\textit{day}$. These alteration experiments are conducted on the ``\#3'' ablation setting mentioned in \cref{sec:AB}, which removes the neural representation function from the NeRCo.

As shown in \cref{blatq1}, we report the results of different settings on LSRW dataset \cite{LSRW}. We design different prompts to study the impact of different texts on model performance. One can see that the ``\#3'' settings with diverse prompts realize decent scores on all four metrics. Although their values are different, they are within a stable range, \textit{i.e.}, better than other ablation settings and worse than NeRCo. On the one hand, we can see that text-driven supervision does have a gain in performance. On the other hand, it also indirectly proves the contribution of our proposed Neural Representation Normalization (NRN) module. Since $\textit{low-light image}$ and $\textit{high-light image}$ are two texts widely adopted to describe images in this task, we used this pair in other experiments for more intuitive validation.

\section{TAD Ablation}
TAD contains three paths: color discrimination, edge supervision, and text-driven discrimination. To further define the role of each component, we conduct ablation study on TAD, which removes different paths from the ablation setting ``\#3'' in the submitted paper. Note that as at least one supervision is required, we adopt color discrimination as the base discriminator. Results are given in \cref{blatTAD}, one can see that both edge path and text supervision improve the effect.

\clearpage
\section{Semantic Evaluation}
In order to demonstrate the superiority of our NeRCo at the semantic level, we employ the pre-trained CLIP model \cite{Radford_2021_ICML} to calculate semantic score of different methods. Concretely, we first design a prompt $\textit{high-light image}$. The image vector and the text vector are then generated by CLIP model. We calculate their cosine discrepancy, and use a softmax function to obtain the semantic score, which values from 0 to 1. A higher score represents the better semantic consistency between the enhanced image and the text $\textit{high-light image}$. 

We report the average prediction accuracy of the results from different methods in \cref{Semantic}. One can see that the pre-trained CLIP considers the input low-light image to be the least likely high-light image, while the high-light references are classified accurately, except for LIME \cite{Guo_2017_TIP} which only contains degraded scenarios. Although some methods output semantically impressive results, our NeRCo achieves the best scores, even better than the ground truth. It proves that the quality of the reference images from the dataset is semantically good, as they achieve the second-best scores. Due to the text-driven discrimination during training, our method produces more perceptual-friendly results than references.
\begin{table*}[t]
	\centering
	\scriptsize
	\resizebox{\linewidth}{!}{
	\begin{tabular}{c|c|c|c|c|c|c|c}
		\toprule[1.pt]
		\multirow{2}{*}{Models} & \multirow{2}{*}{\#1} & \multirow{2}{*}{\#2} & \#3 & \#3 & \#3 & \#3 & \multirow{2}{*}{NeRCo} \\
        &&& ($\textit{low-light image}$ / $\textit{high-light image}$) & ($\textit{dark}$ / $\textit{bright}$) & ($\textit{dim}$ / $\textit{light}$) & ($\textit{night}$ / $\textit{day}$) & \\
		\bottomrule
		\toprule
		PSNR $\uparrow$ & 16.77 & 17.65 & 18.32 & \textbf{\color{blue}18.58} & 18.23 & 18.56 & \textbf{\color{red}19.00} \\ \hline
		SSIM $\uparrow$ & 0.4565 & 0.5023 & 0.5201 & 0.5118 & 0.5208 & \textbf{\color{blue}0.5277} & \textbf{\color{red}0.5360} \\ \hline
		NIQE $\downarrow$ & 12.34 & 10.60 & 10.83 & 9.45 & 9.59 & \textbf{\color{blue}9.33} & \textbf{\color{red}9.23} \\ \hline
		LOE $\downarrow$ & 272.4 & 247.9 & 230.9 & 202.9 & \textbf{\color{blue}191.0} & 206.6 & \textbf{\color{red}189.5} \\ \bottomrule[1.pt]
	\end{tabular}
	}
	\vspace{-1.2em}
	
	\caption{\label{blatq1} Ablation study with different prompt options. The best and the second best results are highlighted in \textbf{\color{red}red} and \textbf{\color{blue}blue} respectively. One can see that settings trained with prompts outperform other versions, and prompts can be replaced with synonyms. It proves that text-driven supervision has a gain in model performance.}
	\vspace{-1.em}
	
\end{table*}

\begin{table*}[t]
	\centering
	\resizebox{\linewidth}{!}{
	\begin{tabular}{c|c c c c c c c c c c c c c}
		\toprule[1.5pt]
		\multirow{2}{*}{Datasets} & \multirow{2}{*}{Input} & LECARM & SDD & RetinexNet & KinD & URetinexNet & ZeroDCE & SSIENet & RUAS & EnGAN & SCI & \multirow{2}{*}{NeRCo} & \multirow{2}{*}{Reference} \\ 
		& & \cite{LECARM} & \cite{SDD} & \cite{LOL} & \cite{Zhang_2019_MM} & \cite{Wu_2022_CVPR} & \cite{Guo_2020_CVPR} & \cite{SSIENet} & \cite{Liu_2021_CVPR} & \cite{Jiang_2021_TIP} & \cite{Ma_2022_CVPR} & &
		\\ \bottomrule
        \toprule
		LOL \cite{LOL} & 0.1590 & 0.3685 & 0.3397 & 0.4831 & 0.5130 & 0.5554 & 0.3463 & 0.4639 & 0.4381 & 0.4450 & 0.3402 & \textbf{\color{red}0.6366} & \textbf{\color{blue}0.5910} \\ \hline
		LSRW \cite{LSRW} & 0.3164 & 0.6028 & 0.5907 & 0.6653 & 0.6052 & 0.6539 & 0.6584 & 0.6746 & 0.6418 & 0.5969 & 0.6176 & \textbf{\color{red}0.7581} & \textbf{\color{blue}0.6955} \\ \hline
		LIME \cite{Guo_2017_TIP} & 0.3281 & 0.5168 & 0.4669 & 0.5657 & 0.5589 & \textbf{\color{blue}0.6265} & 0.4719 & 0.6147 & 0.5427 & 0.4912 & 0.5781 & \textbf{\color{red}0.7499} & - \\ \bottomrule[1.5pt]
	\end{tabular}
	}
	\vspace{-1.2em}
	
	\caption{\label{Semantic} The average semantic scores of different settings on three benchmarks. The best and the second best results are highlighted in \textbf{\color{red}red} and \textbf{\color{blue}blue} respectively. One can see that the pre-trained vision-language model classifies our results more accurately than those of other methods, which demonstrates the better semantic consistency of our method.}
    \vspace{-1.em}

\end{table*}

\begin{table*}[t]
	\centering
	\scriptsize
	\resizebox{120mm}{!}{
    \begin{tabular}{c|c|c|c|c|c|c}
    \toprule
    Color Path & Edge Path & Text Supervision & PSNR $\uparrow$ & SSIM $\uparrow$ & NIQE $\downarrow$ & LOE $\downarrow$ \\ \bottomrule
    \toprule
    \CheckmarkBold & \XSolidBrush & \XSolidBrush & 17.37 & 0.4942 & 11.72 & 252.1 \\
    \CheckmarkBold & \CheckmarkBold & \XSolidBrush & \textbf{\color{blue}17.65} & \textbf{\color{blue}0.5023} & \textbf{\color{red}10.60} & \textbf{\color{blue}247.9} \\
    \CheckmarkBold & \CheckmarkBold & \CheckmarkBold & \textbf{\color{red}18.32} & \textbf{\color{red}0.5201} & \textbf{\color{blue}10.83} & \textbf{\color{red}230.9} \\
    \bottomrule
    \end{tabular}
    }
	\vspace{-1.2em}
	
	\caption{\label{blatTAD} Ablation study on TAD, based on the ablation setting ``\#3''. We conduct experiments on LSRW dataset. The best and the second best results are highlighted in \textbf{\color{red}red} and \textbf{\color{blue}blue} respectively.}
	\vspace{-1.em}
	
\end{table*}

\section{Qualitative Analysis}
We have provided adequate quantitative results (\cref{Quantitative}) in our paper. However, due to the limit of space, only parts of visual comparisons are given (\cref{comlsrw}). Here, we supplement more qualitative analysis compared with other SOTA methods, including LECARM \cite{LECARM}, SDD \cite{SDD}, RetinexNet \cite{LOL}, KinD \cite{Zhang_2019_MM}, URetinex-Net \cite{Wu_2022_CVPR}, ZeroDCE \cite{Guo_2020_CVPR}, SSIENet \cite{SSIENet}, RUAS \cite{Liu_2021_CVPR}, EnGAN \cite{Jiang_2021_TIP}, and SCI \cite{Ma_2022_CVPR}.

\begin{figure*}[t]
  \centering
  \begin{subfigure}{0.162\linewidth}
    \includegraphics[width=1.\linewidth]{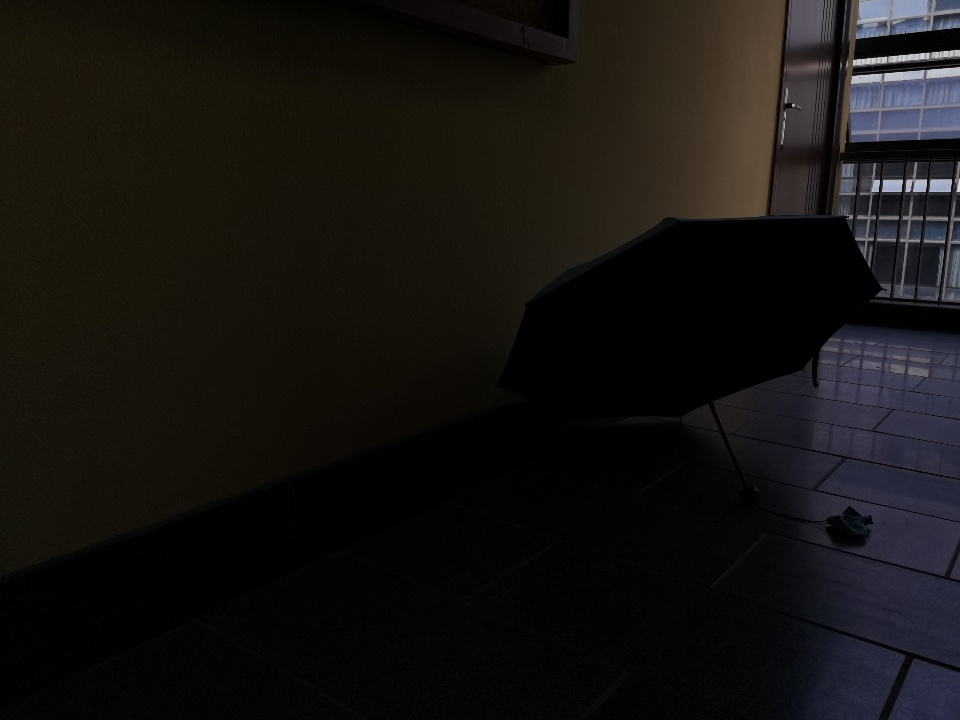}\vspace{-0.4em}
    \centerline{Input}\vspace{-0.5em}\medskip
    
    \includegraphics[width=1.\linewidth]{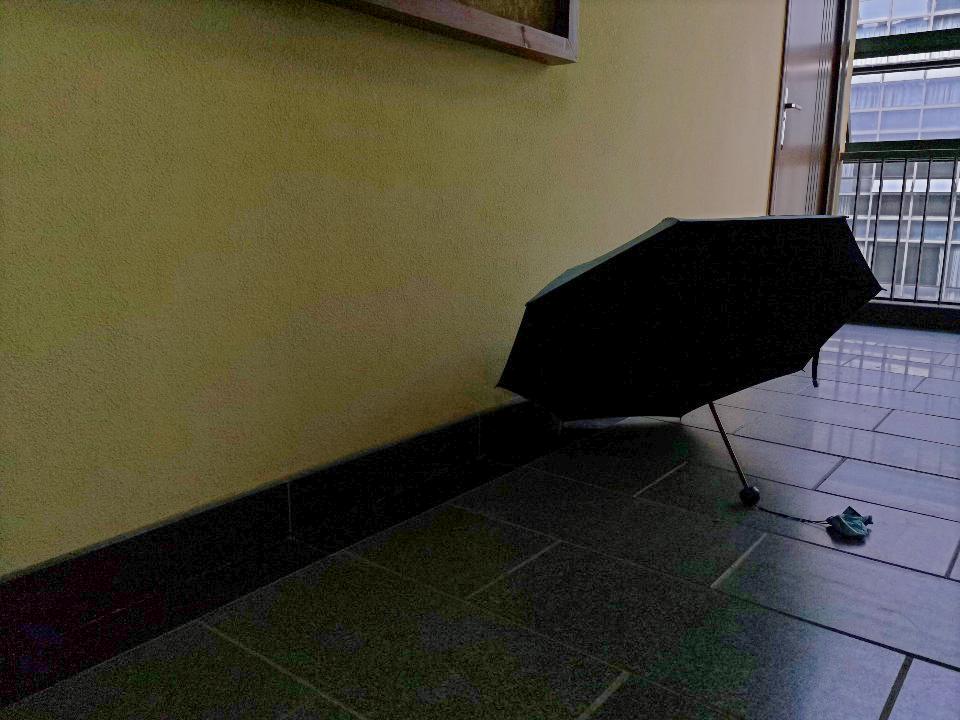}\vspace{-0.2em}
    \centerline{ZeroDCE}\medskip
    
    \includegraphics[width=1.\linewidth]{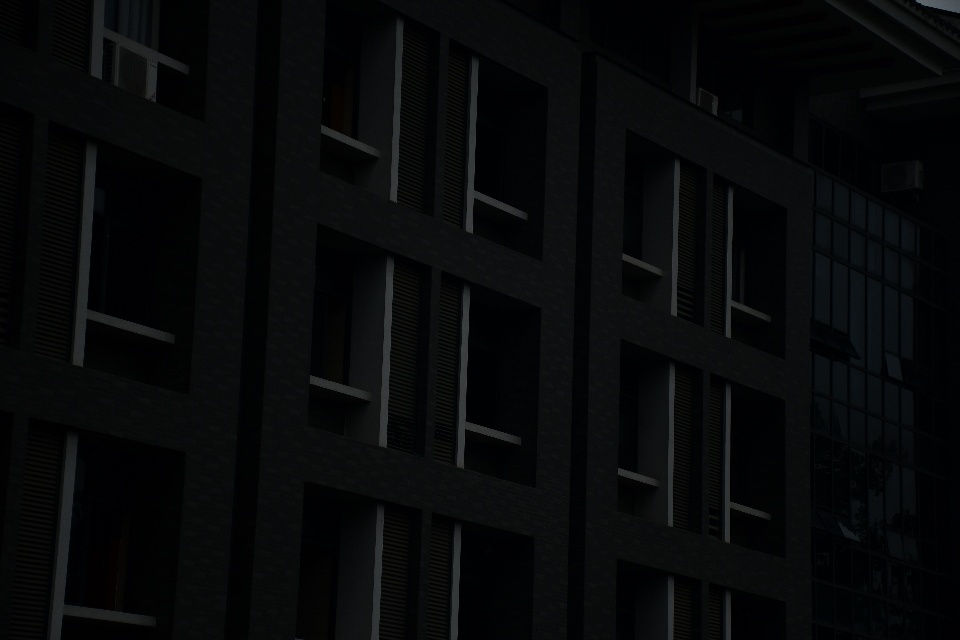}\vspace{-0.4em}
    \centerline{Input}\vspace{-0.5em}\medskip
    
    \includegraphics[width=1.\linewidth]{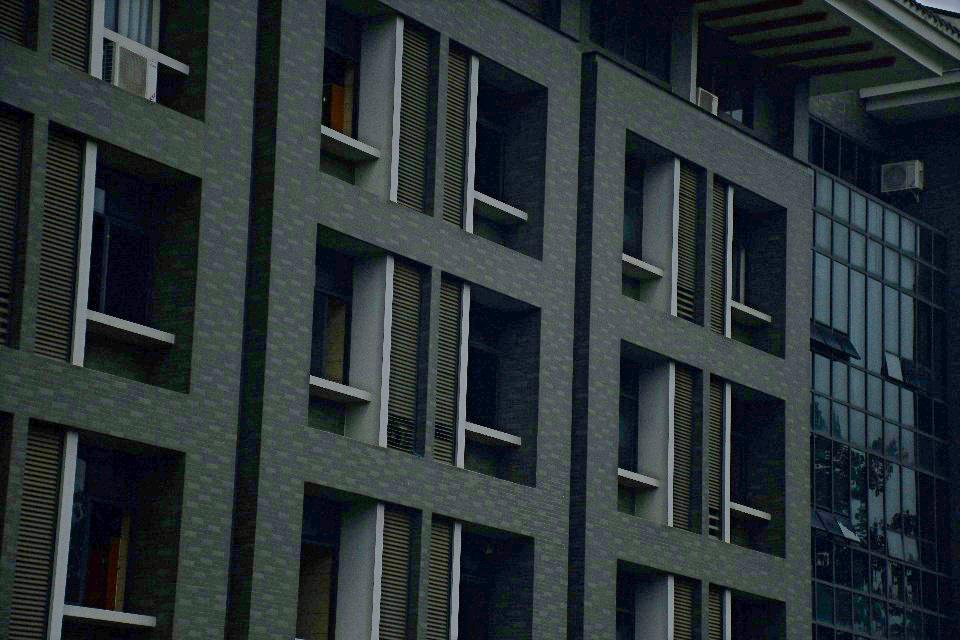}\vspace{-0.2em}
    \centerline{ZeroDCE}\medskip
  \end{subfigure}
  \hfill
  \begin{subfigure}{0.162\linewidth}
    \includegraphics[width=1.\linewidth]{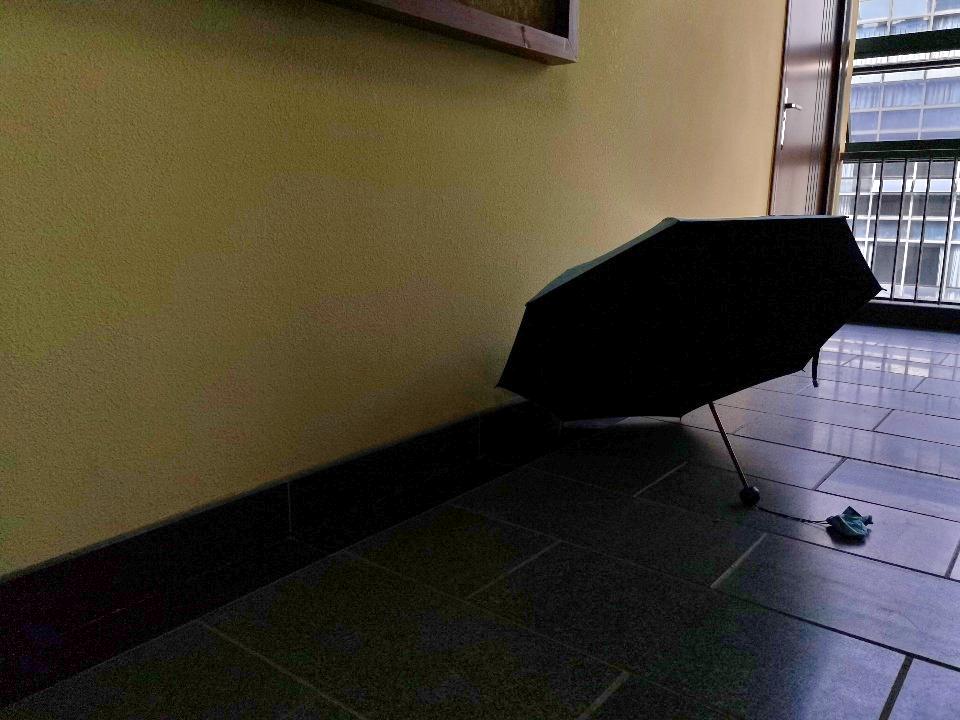}\vspace{-0.2em}
    \centerline{LECARM}\vspace{-0.5em}\medskip
    
    \includegraphics[width=1.\linewidth]{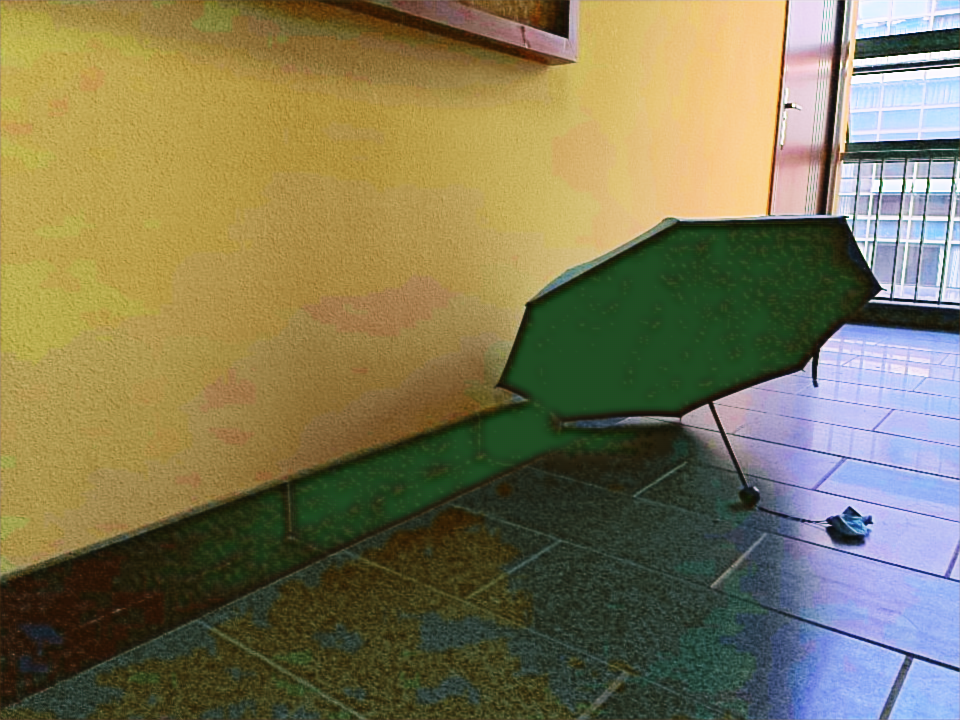}\vspace{-0.2em}
    \centerline{SSIENet}\medskip
    
    \includegraphics[width=1.\linewidth]{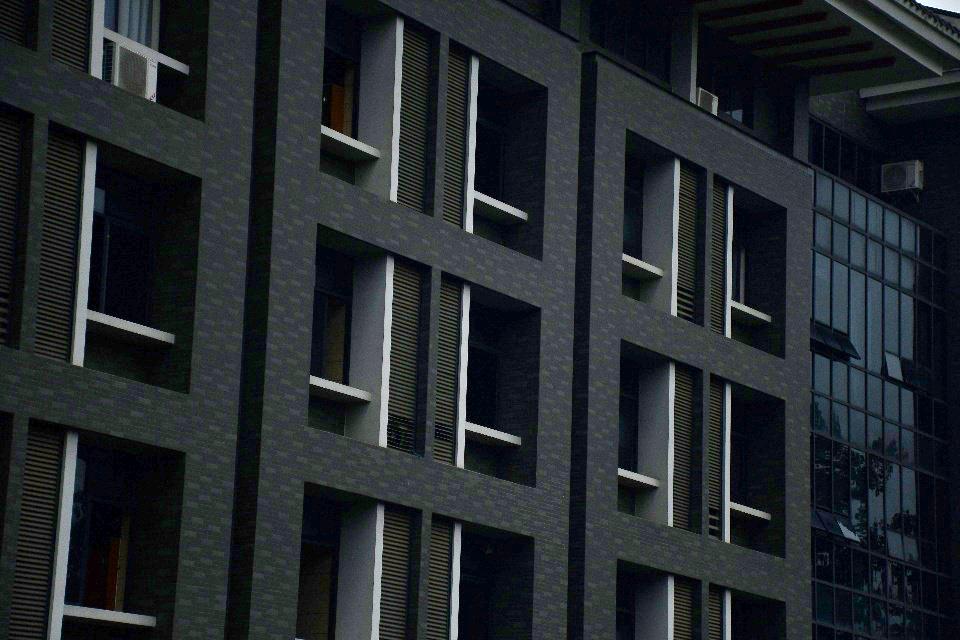}\vspace{-0.2em}
    \centerline{LECARM}\vspace{-0.5em}\medskip
    
    \includegraphics[width=1.\linewidth]{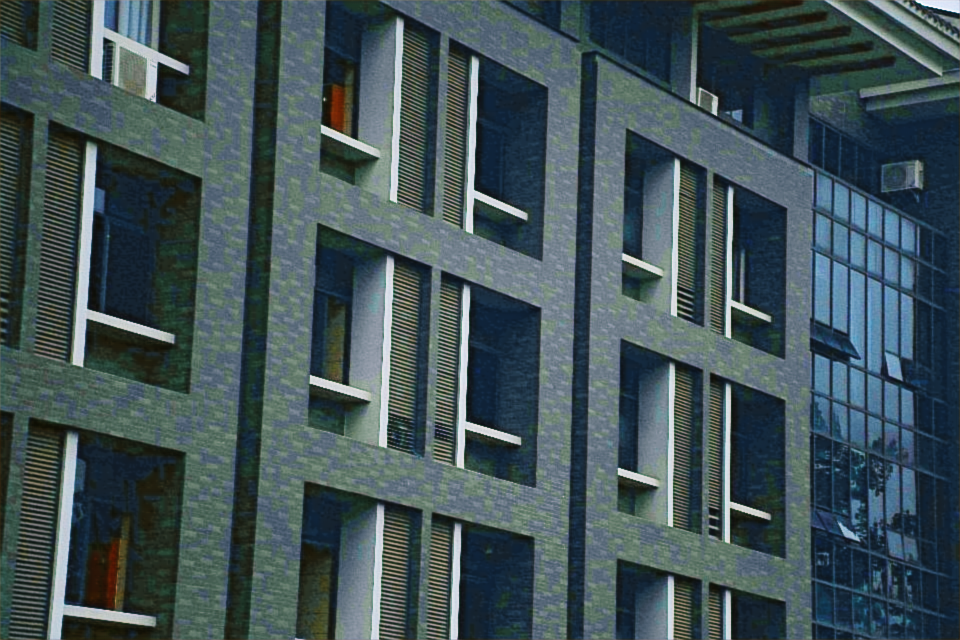}\vspace{-0.2em}
    \centerline{SSIENet}\medskip
  \end{subfigure}
  \hfill
  \begin{subfigure}{0.162\linewidth}
    \includegraphics[width=1.\linewidth]{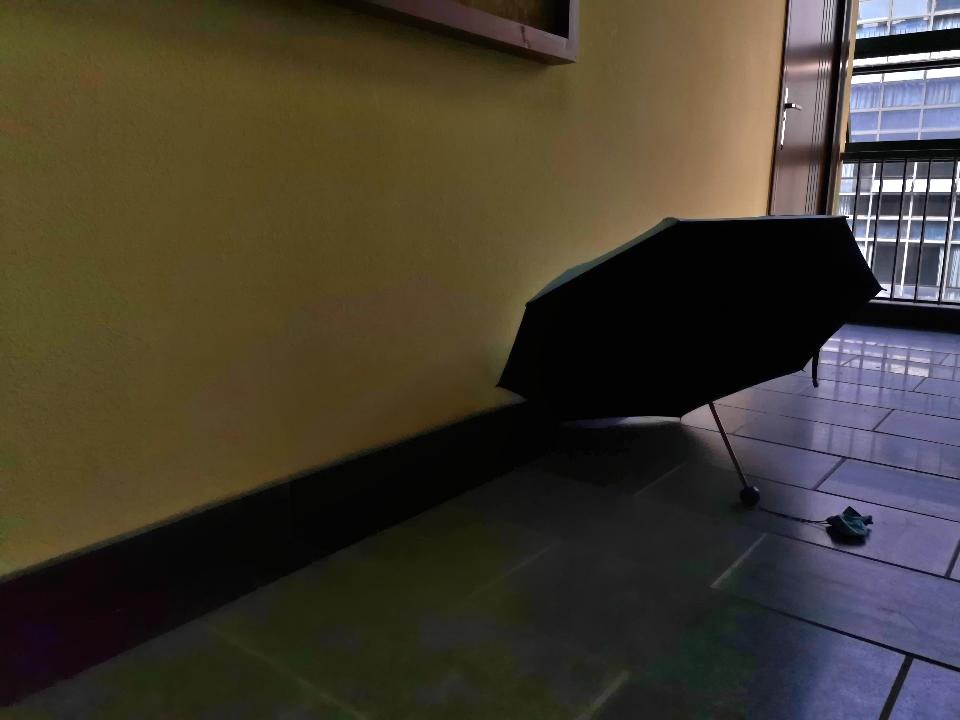}\vspace{-0.2em}
    \centerline{SDD}\vspace{-0.5em}\medskip
    
    \includegraphics[width=1.\linewidth]{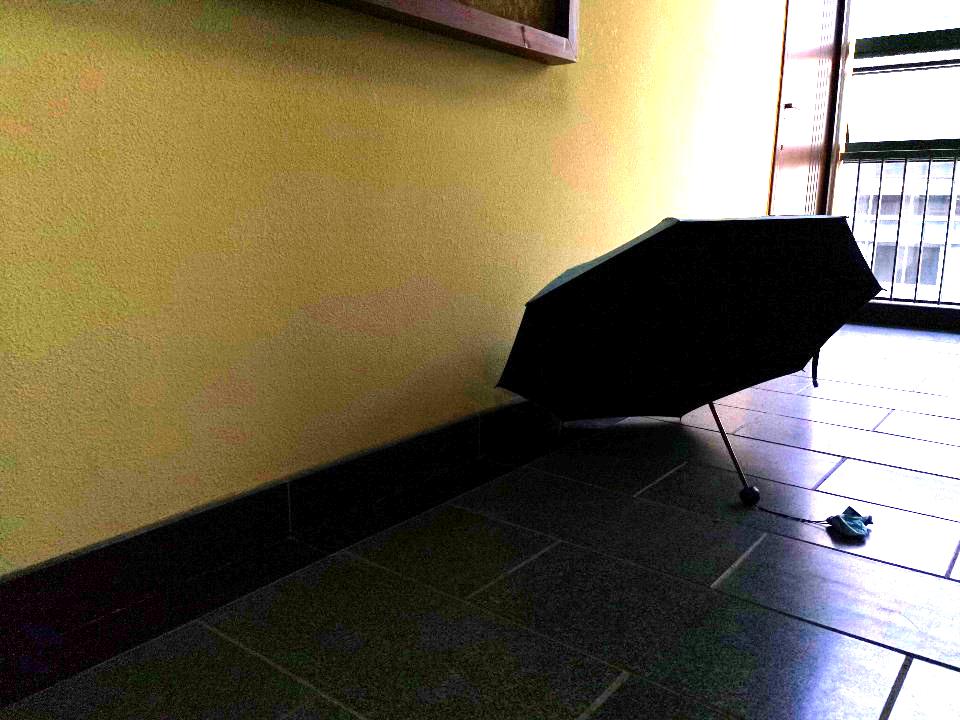}\vspace{-0.2em}
    \centerline{RUAS}\medskip
    
    \includegraphics[width=1.\linewidth]{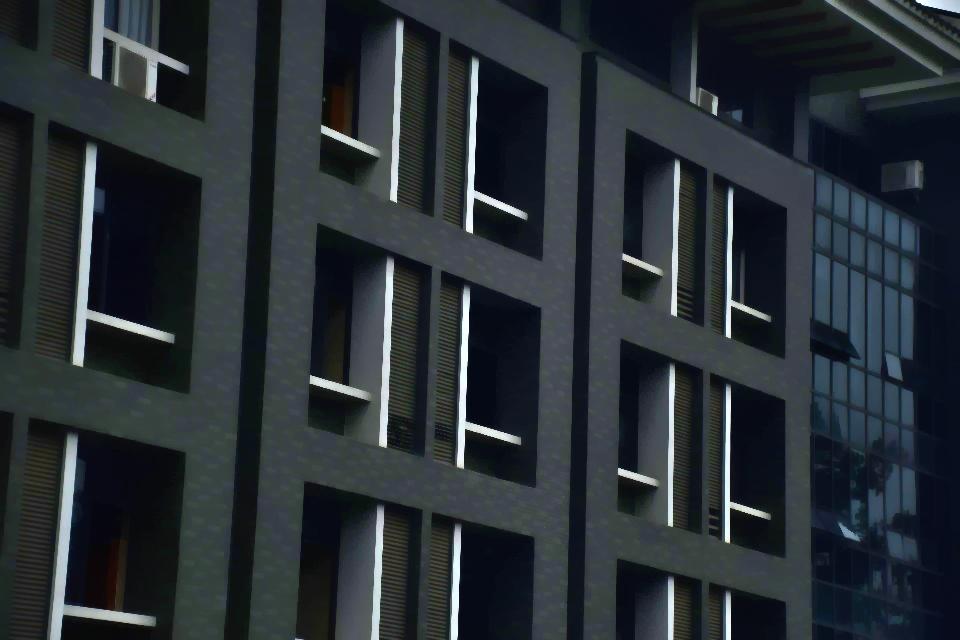}\vspace{-0.2em}
    \centerline{SDD}\vspace{-0.5em}\medskip
    
    \includegraphics[width=1.\linewidth]{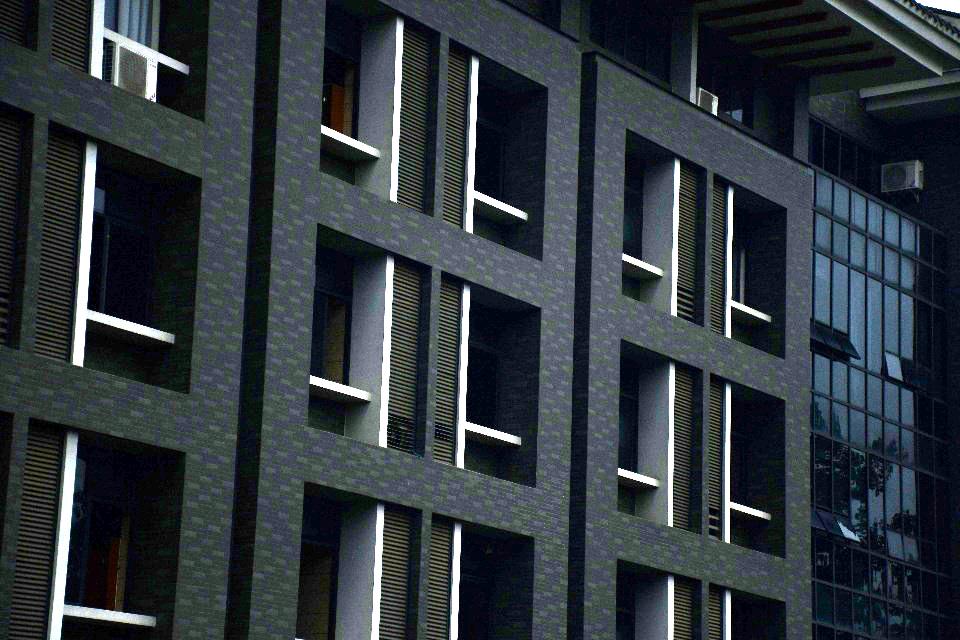}\vspace{-0.2em}
    \centerline{RUAS}\medskip
  \end{subfigure}
  \hfill
  \begin{subfigure}{0.162\linewidth}
    \includegraphics[width=1.\linewidth]{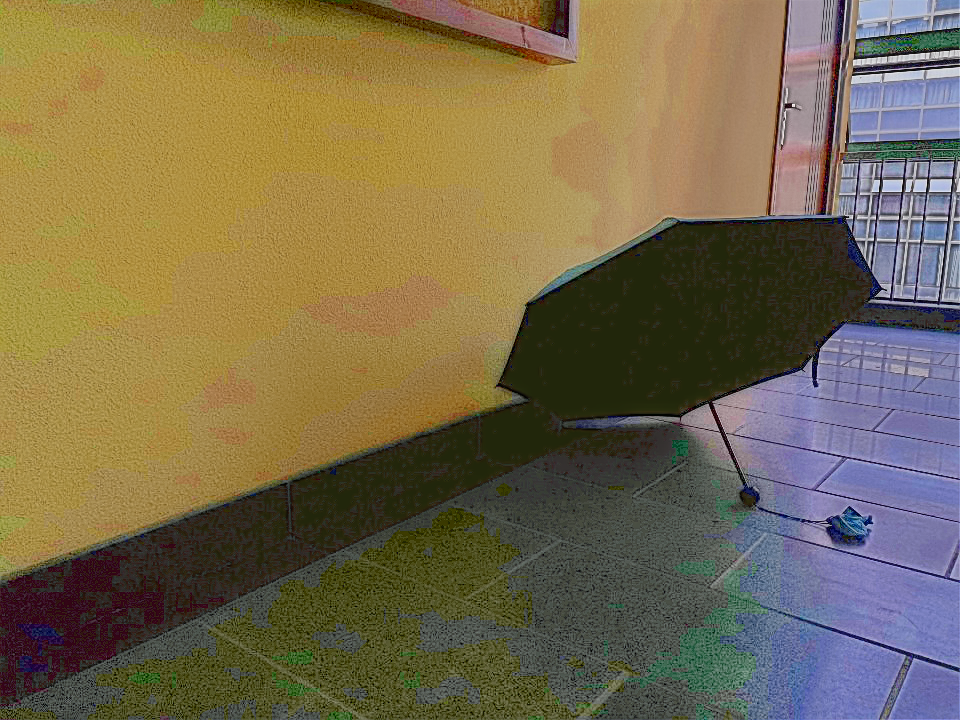}\vspace{-0.2em}
    \centerline{RetinexNet}\vspace{-0.5em}\medskip
    
    \includegraphics[width=1.\linewidth]{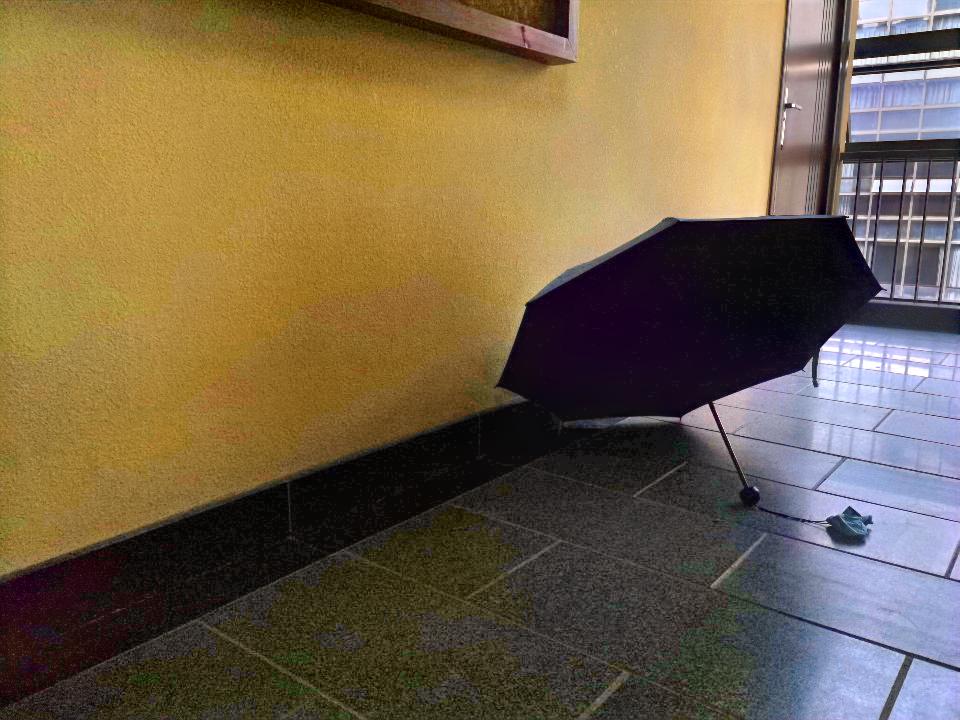}\vspace{-0.2em}
    \centerline{EnGAN}\medskip
    
    \includegraphics[width=1.\linewidth]{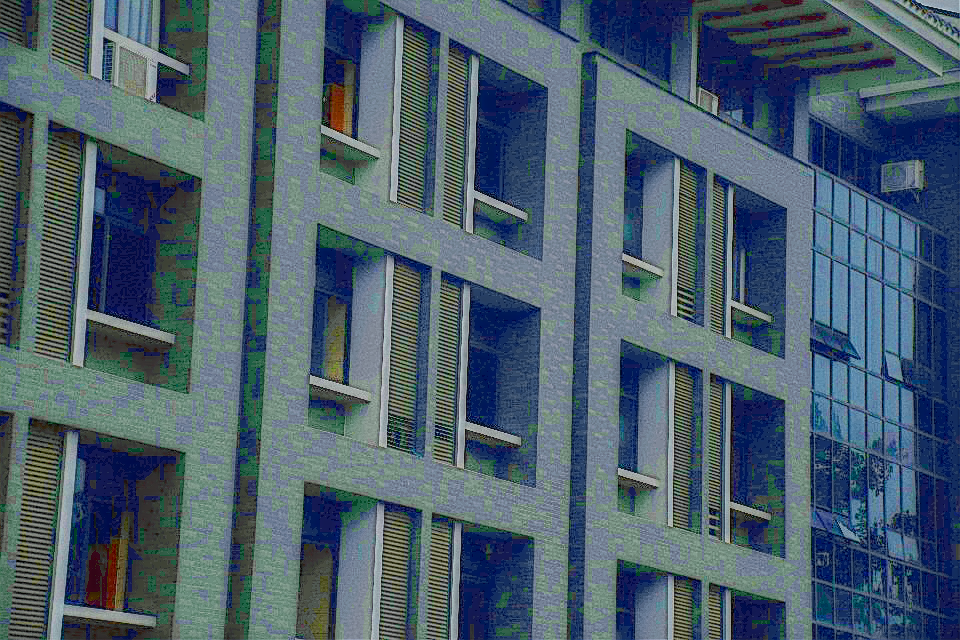}\vspace{-0.2em}
    \centerline{RetinexNet}\vspace{-0.5em}\medskip
    
    \includegraphics[width=1.\linewidth]{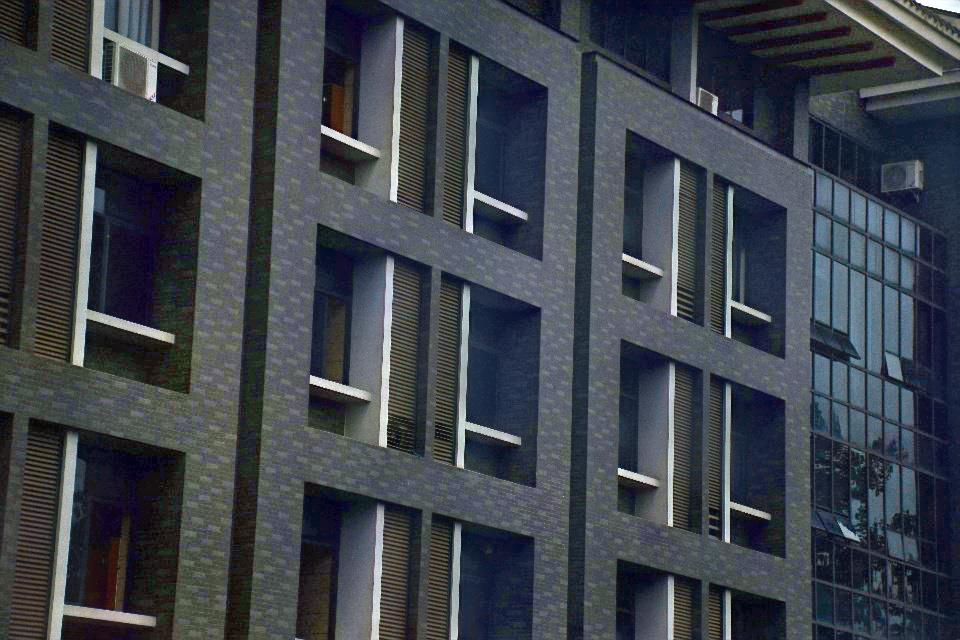}\vspace{-0.2em}
    \centerline{EnGAN}\medskip
  \end{subfigure}
  \hfill
  \begin{subfigure}{0.162\linewidth}
    \includegraphics[width=1.\linewidth]{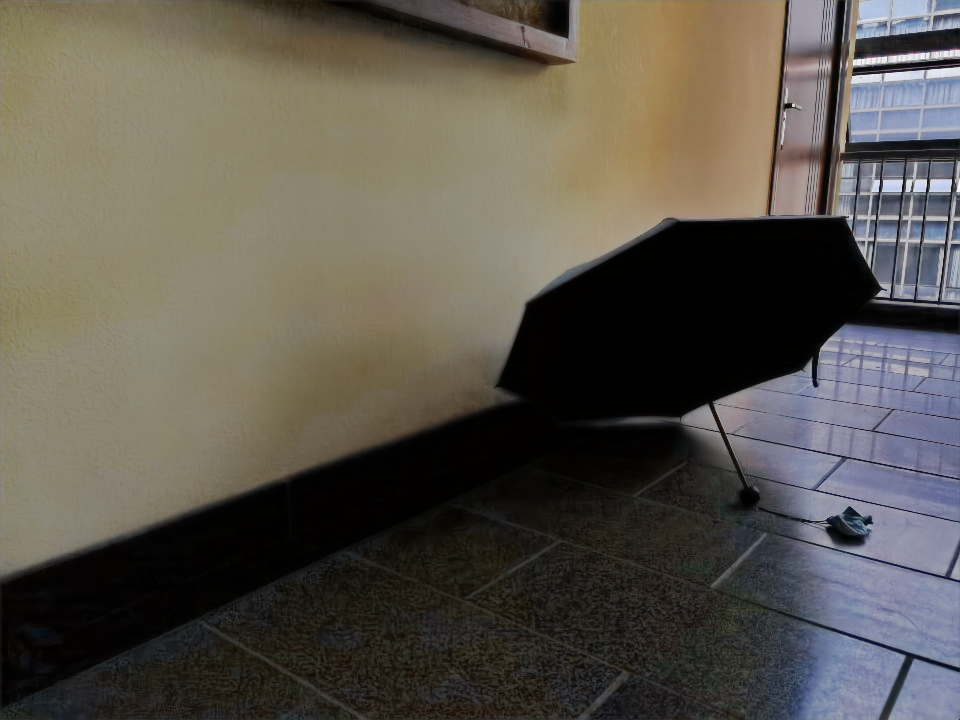}\vspace{-0.2em}
    \centerline{KinD}\vspace{-0.5em}\medskip
    
    \includegraphics[width=1.\linewidth]{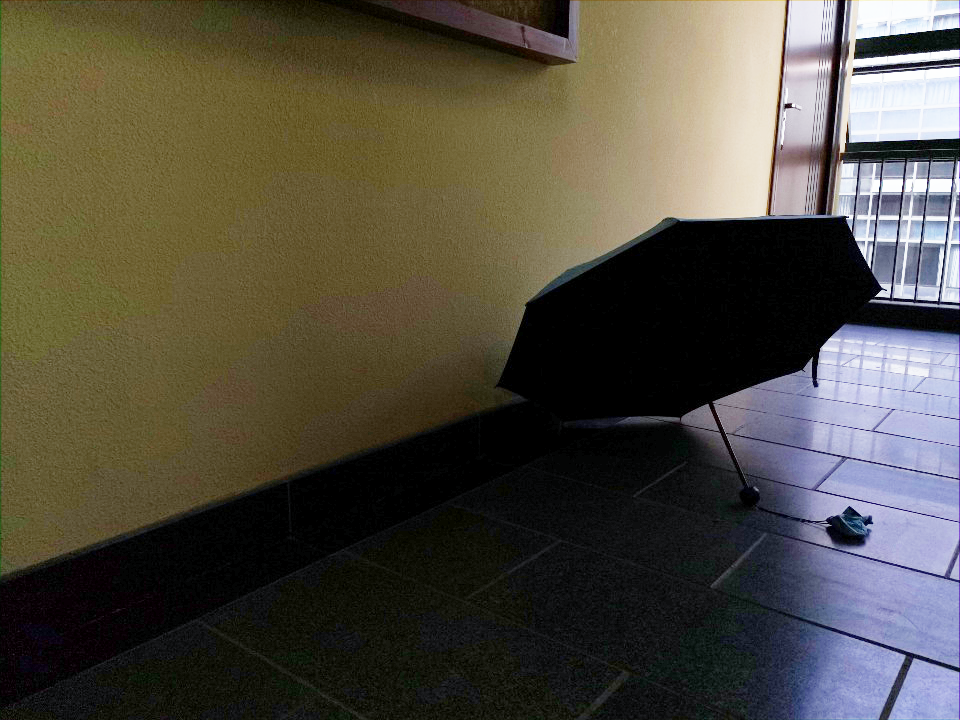}\vspace{-0.2em}
    \centerline{SCI}\medskip
    
    \includegraphics[width=1.\linewidth]{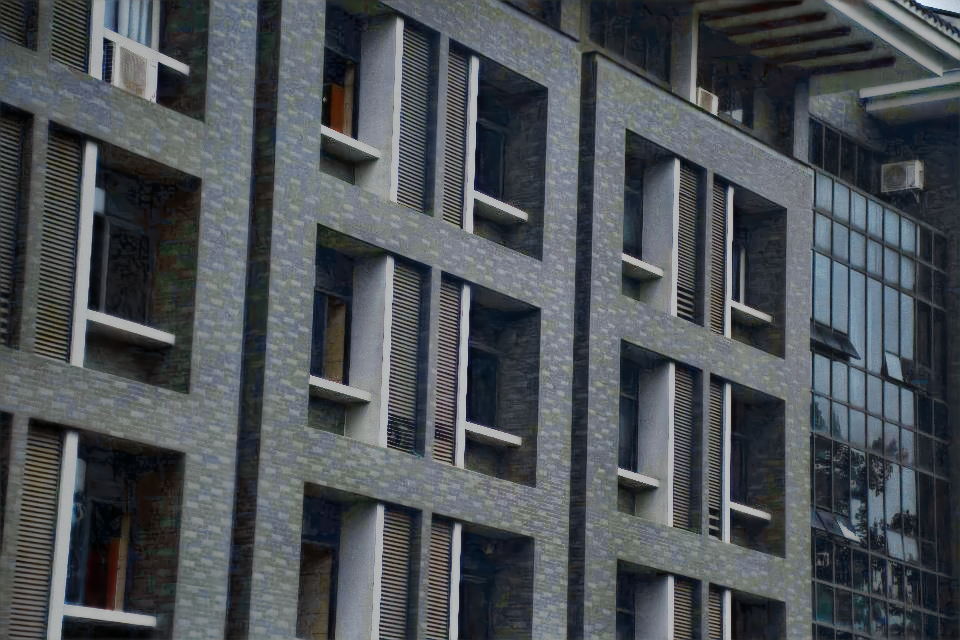}\vspace{-0.2em}
    \centerline{KinD}\vspace{-0.5em}\medskip
    
    \includegraphics[width=1.\linewidth]{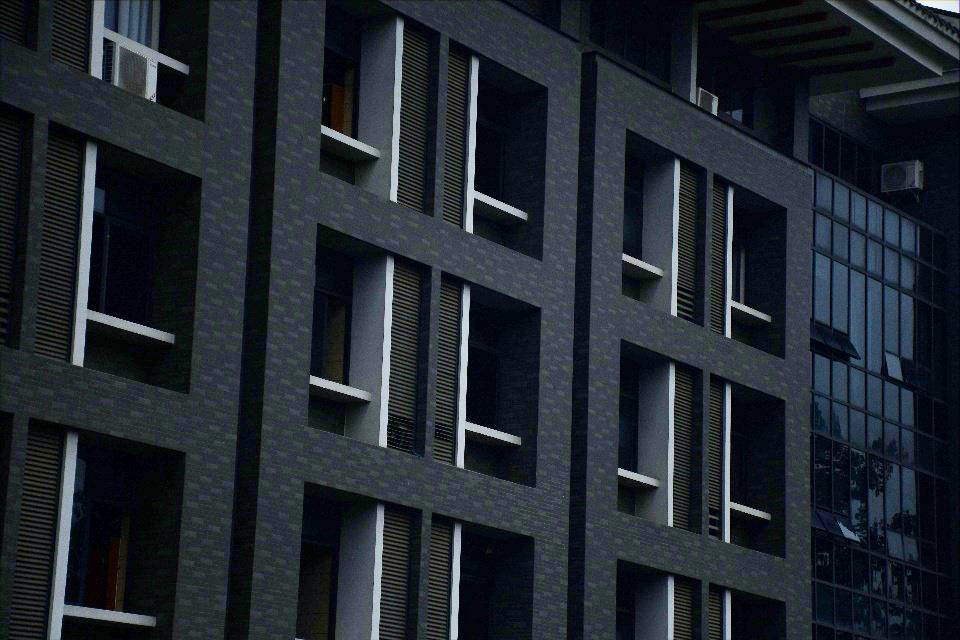}\vspace{-0.2em}
    \centerline{SCI}\medskip
  \end{subfigure}
  \begin{subfigure}{0.162\linewidth}
    \includegraphics[width=1.\linewidth]{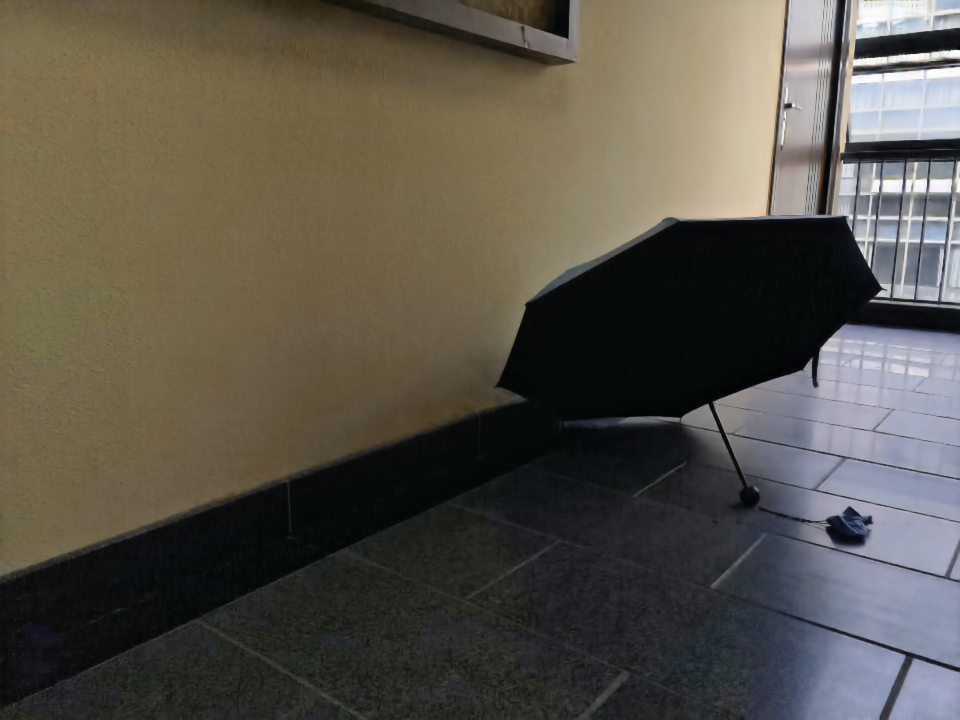}\vspace{-0.2em}
    \centerline{URetinexNet}\vspace{-0.5em}\medskip
    
    \includegraphics[width=1.\linewidth]{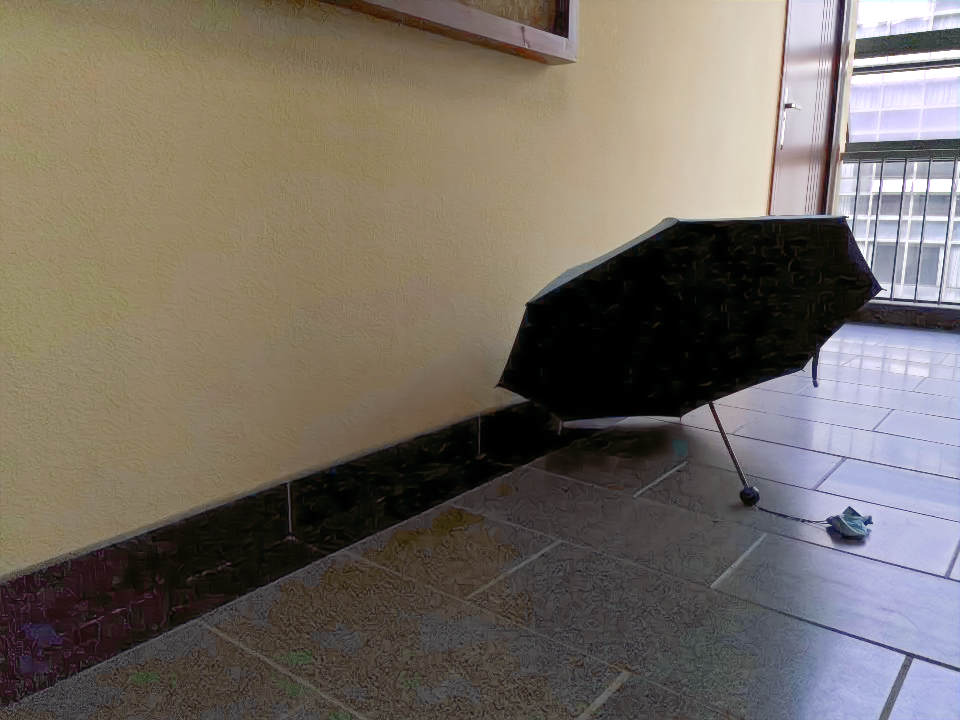}\vspace{-0.2em}
    \centerline{Ours}\medskip
    
    \includegraphics[width=1.\linewidth]{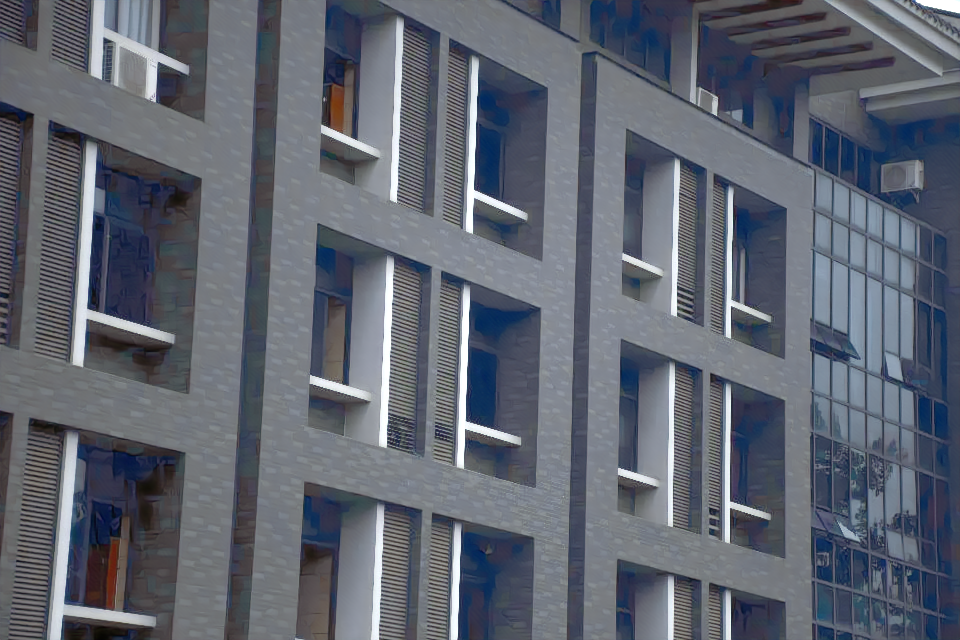}\vspace{-0.2em}
    \centerline{URetinexNet}\vspace{-0.5em}\medskip
    
    \includegraphics[width=1.\linewidth]{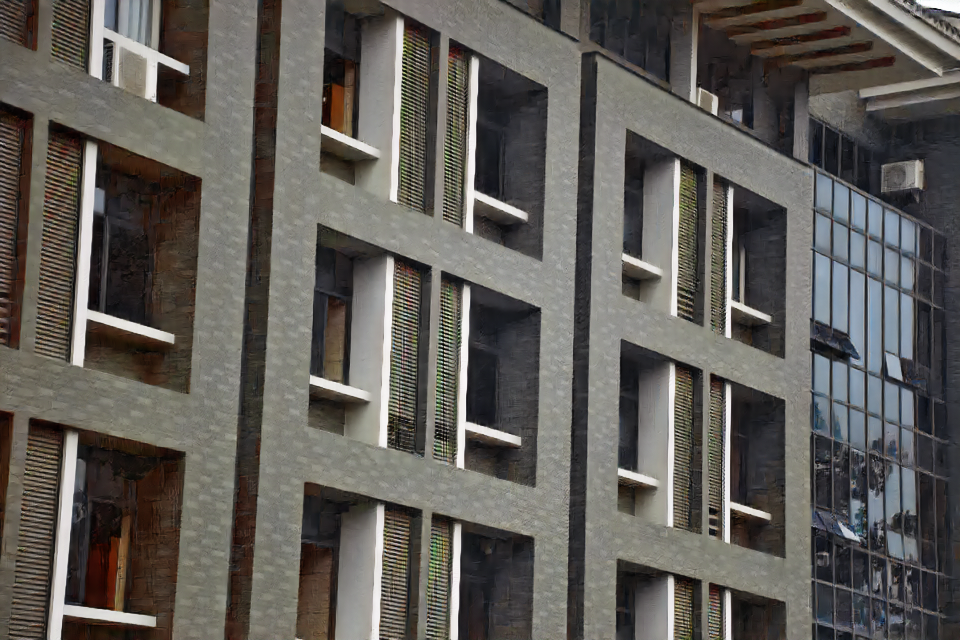}\vspace{-0.2em}
    \centerline{Ours}\medskip
  \end{subfigure}
  \vspace{-1.em}

  \caption{Subjective comparison on the LSRW dataset among state-of-the-art low-light image enhancement algorithms. Obviously, the proposed method has achieved the best performance, further verifying its effectiveness.}
  \label{comlsrw1}
  \vspace{-1.em}
\end{figure*}

\begin{figure*}[t]
  \centering
  \begin{subfigure}{0.162\linewidth}
    \includegraphics[width=1.\linewidth]{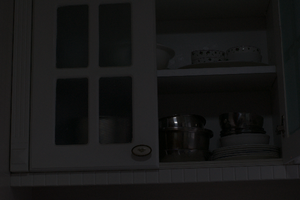}\vspace{-0.4em}
    \centerline{Input}\vspace{-0.5em}\medskip
    
    \includegraphics[width=1.\linewidth]{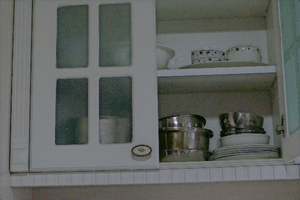}\vspace{-0.2em}
    \centerline{ZeroDCE}\medskip
  \end{subfigure}
  \hfill
  \begin{subfigure}{0.162\linewidth}
    \includegraphics[width=1.\linewidth]{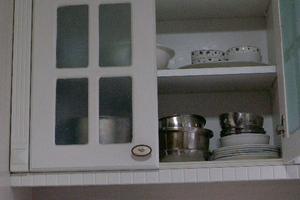}\vspace{-0.2em}
    \centerline{LECARM}\vspace{-0.5em}\medskip
    
    \includegraphics[width=1.\linewidth]{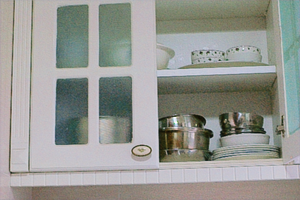}\vspace{-0.2em}
    \centerline{SSIENet}\medskip
  \end{subfigure}
  \hfill
  \begin{subfigure}{0.162\linewidth}
    \includegraphics[width=1.\linewidth]{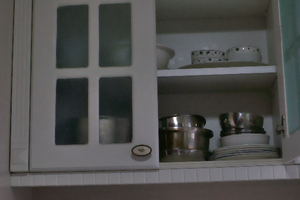}\vspace{-0.2em}
    \centerline{SDD}\vspace{-0.5em}\medskip
    
    \includegraphics[width=1.\linewidth]{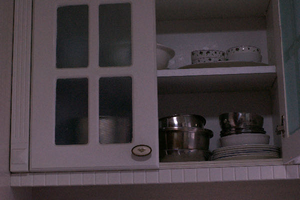}\vspace{-0.2em}
    \centerline{RUAS}\medskip
  \end{subfigure}
  \hfill
  \begin{subfigure}{0.162\linewidth}
    \includegraphics[width=1.\linewidth]{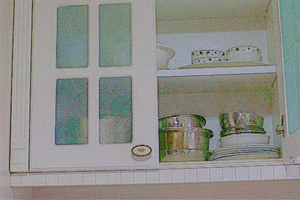}\vspace{-0.2em}
    \centerline{RetinexNet}\vspace{-0.5em}\medskip
    
    \includegraphics[width=1.\linewidth]{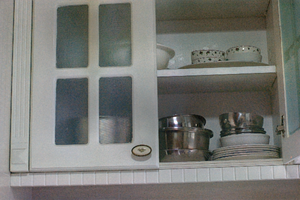}\vspace{-0.2em}
    \centerline{EnGAN}\medskip
  \end{subfigure}
  \hfill
  \begin{subfigure}{0.162\linewidth}
    \includegraphics[width=1.\linewidth]{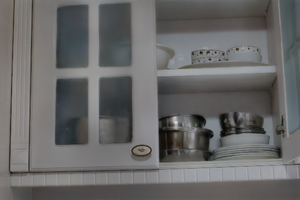}\vspace{-0.2em}
    \centerline{KinD}\vspace{-0.5em}\medskip
    
    \includegraphics[width=1.\linewidth]{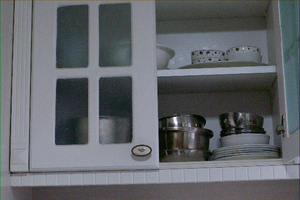}\vspace{-0.2em}
    \centerline{SCI}\medskip
  \end{subfigure}
  \begin{subfigure}{0.162\linewidth}
    \includegraphics[width=1.\linewidth]{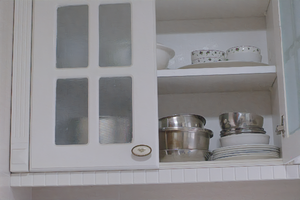}\vspace{-0.2em}
    \centerline{URetinexNet}\vspace{-0.5em}\medskip
    
    \includegraphics[width=1.\linewidth]{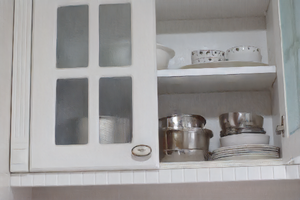}\vspace{-0.2em}
    \centerline{Ours}\medskip
  \end{subfigure}
  \vspace{-1.em}

  \caption{Subjective comparison on the LOL dataset among state-of-the-art low-light image enhancement algorithms. It is obvious that our method recovers the most authentic results, demonstrating its superiority.}
  \label{comlol}
  \vspace{-1.em}
  
\end{figure*}

\begin{figure*}[t]
  \centering
  \begin{subfigure}{0.162\linewidth}
    \includegraphics[width=1.\linewidth]{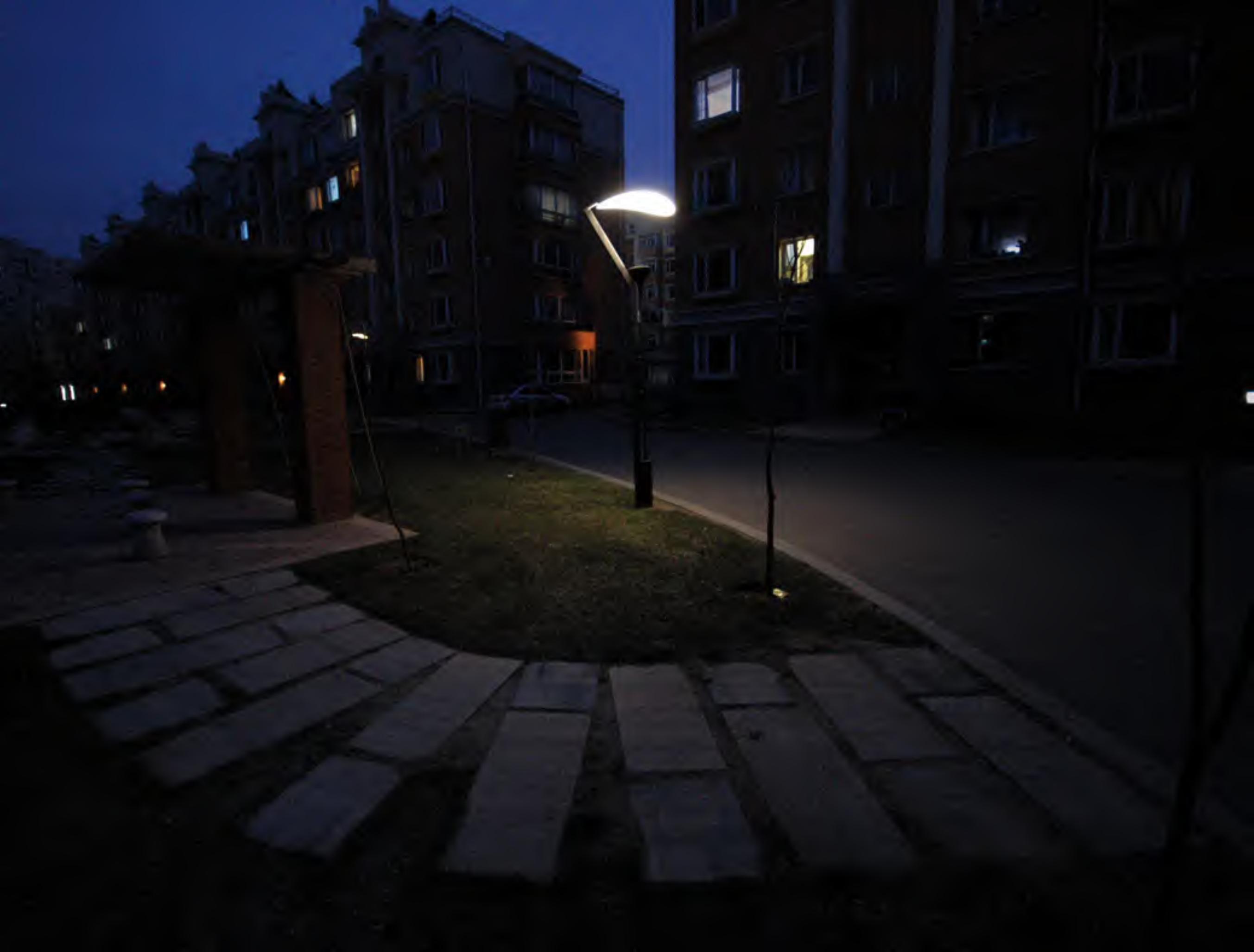}\vspace{-0.4em}
    \centerline{Input}\vspace{-0.5em}\medskip
    
    \includegraphics[width=1.\linewidth]{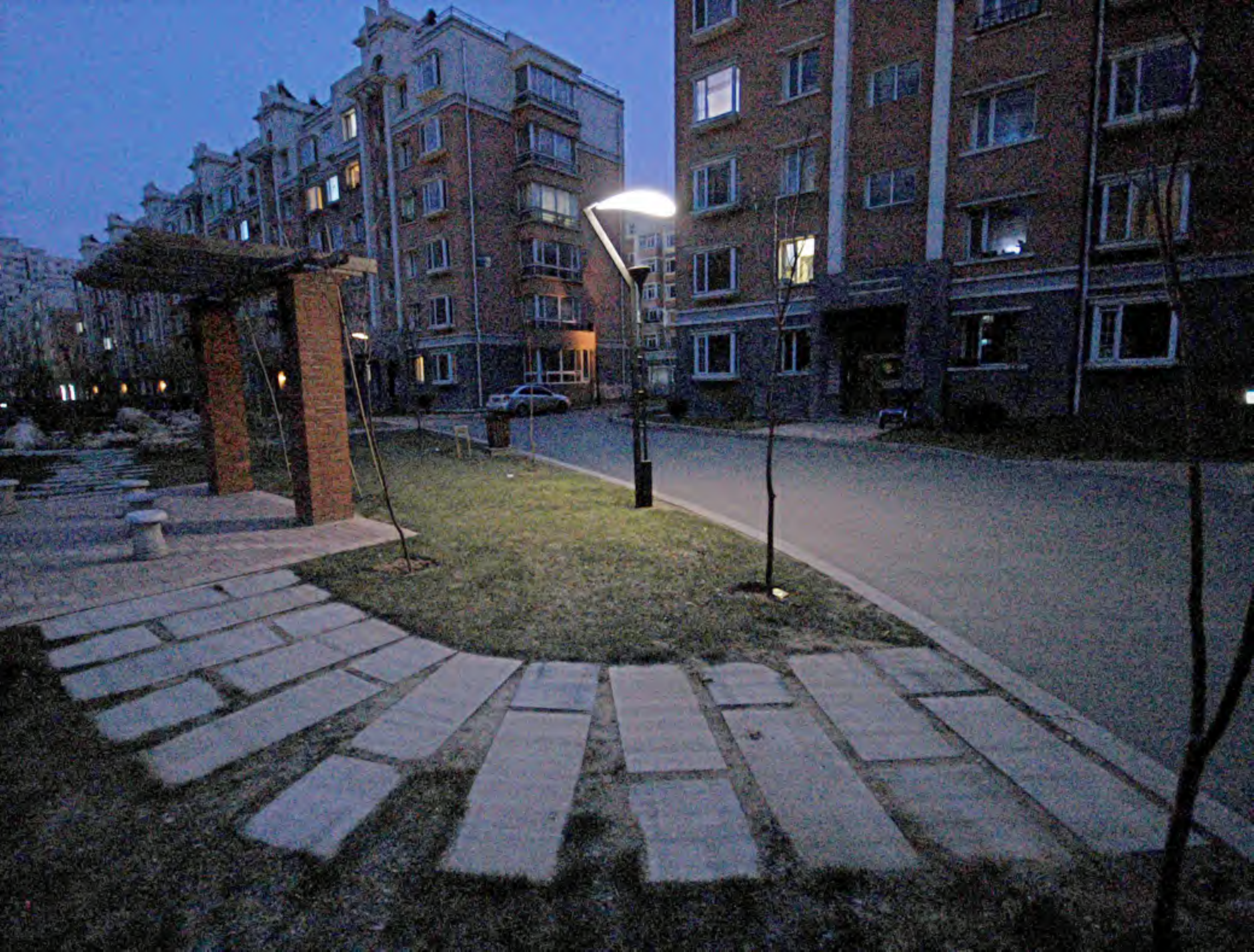}\vspace{-0.2em}
    \centerline{ZeroDCE}\medskip
  \end{subfigure}
  \hfill
  \begin{subfigure}{0.162\linewidth}
    \includegraphics[width=1.\linewidth]{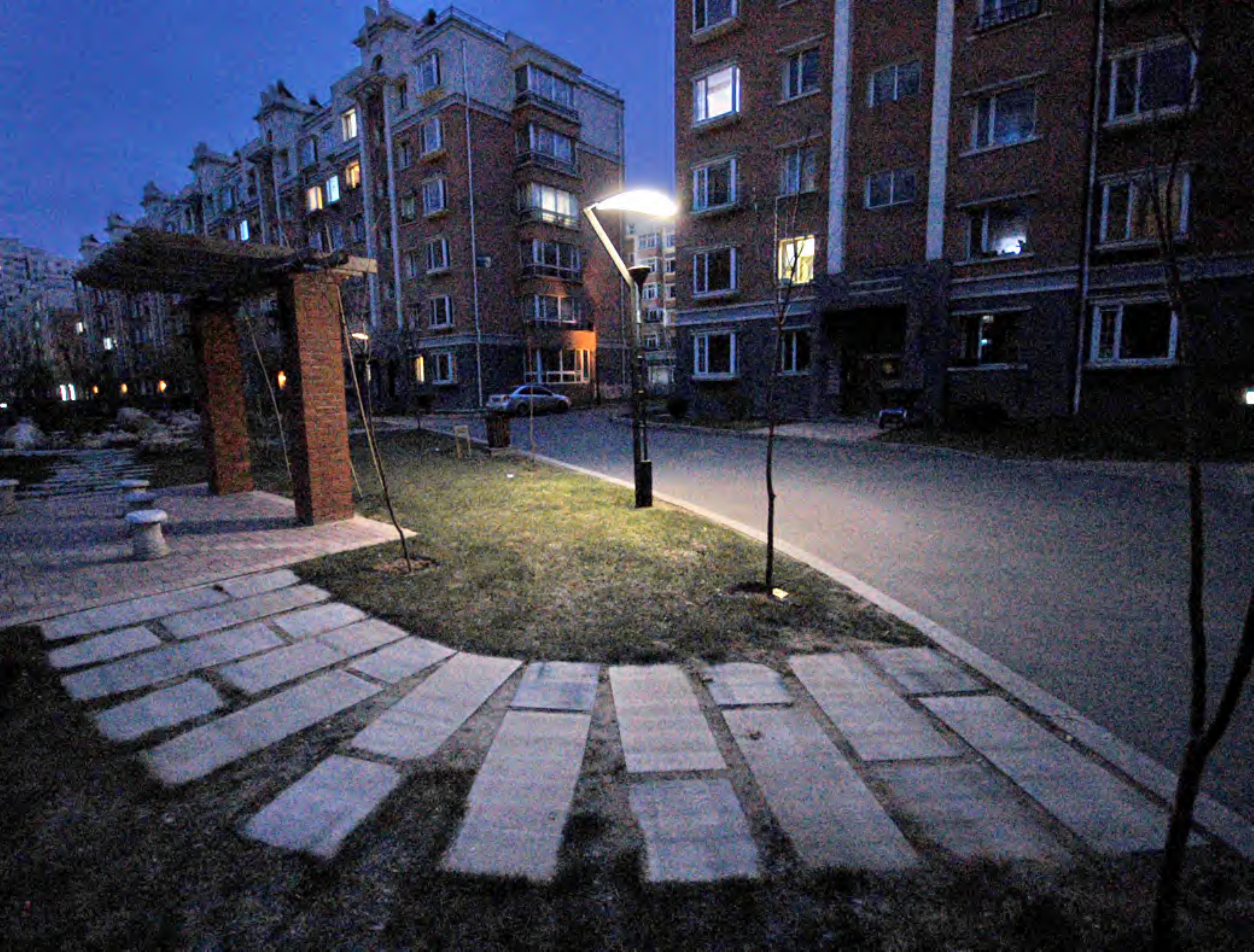}\vspace{-0.2em}
    \centerline{LECARM}\vspace{-0.5em}\medskip
    
    \includegraphics[width=1.\linewidth]{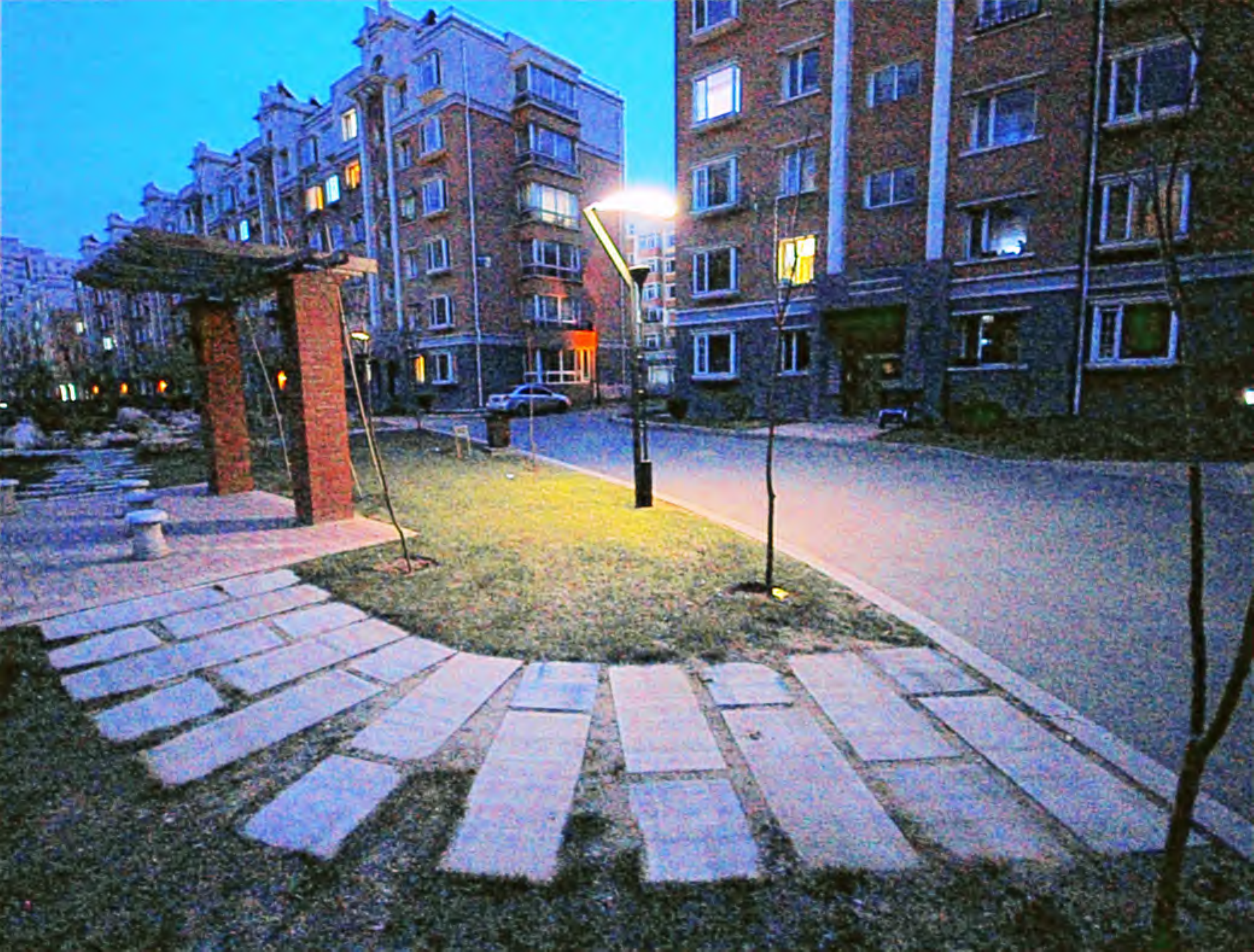}\vspace{-0.2em}
    \centerline{SSIENet}\medskip
  \end{subfigure}
  \hfill
  \begin{subfigure}{0.162\linewidth}
    \includegraphics[width=1.\linewidth]{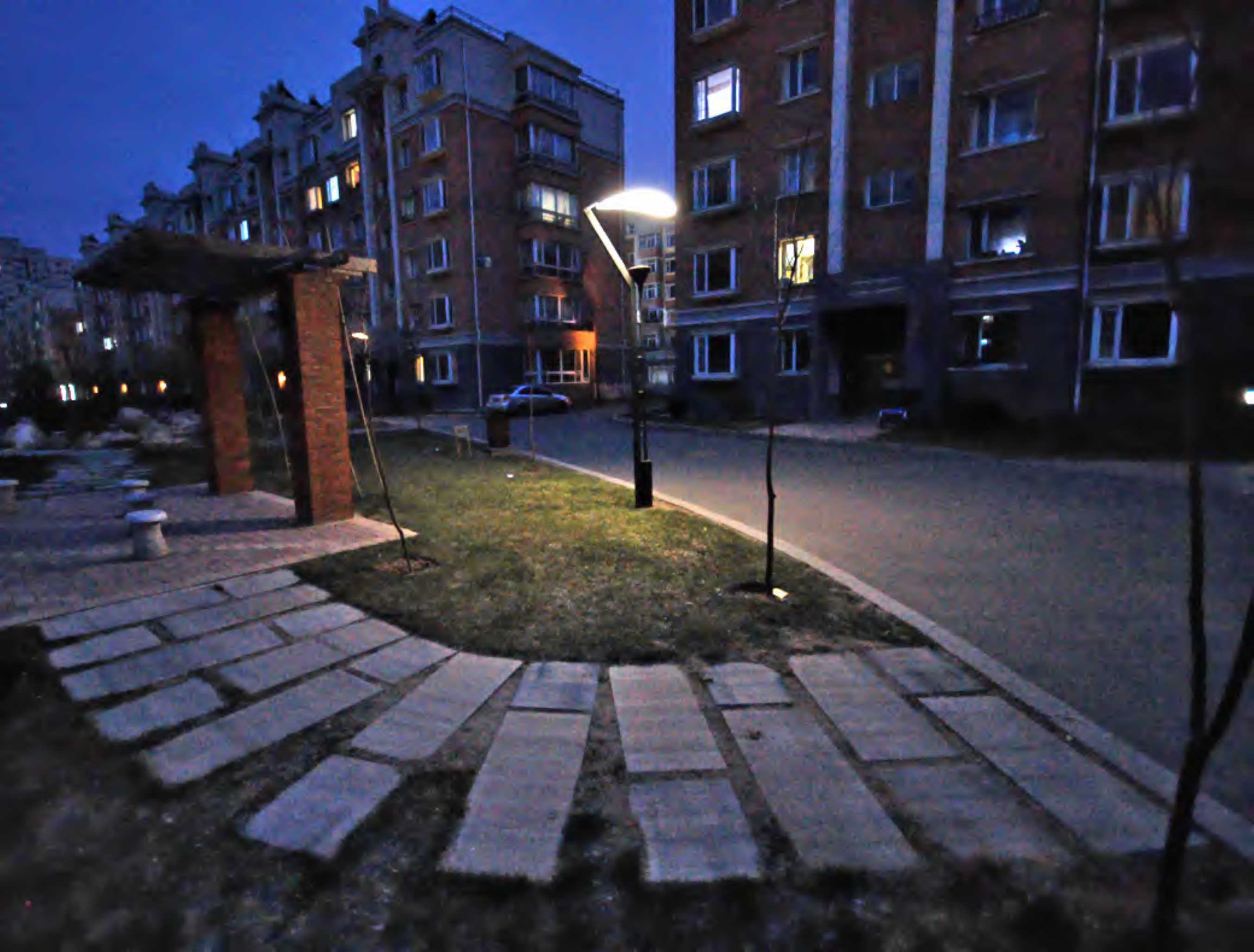}\vspace{-0.2em}
    \centerline{SDD}\vspace{-0.5em}\medskip
    
    \includegraphics[width=1.\linewidth]{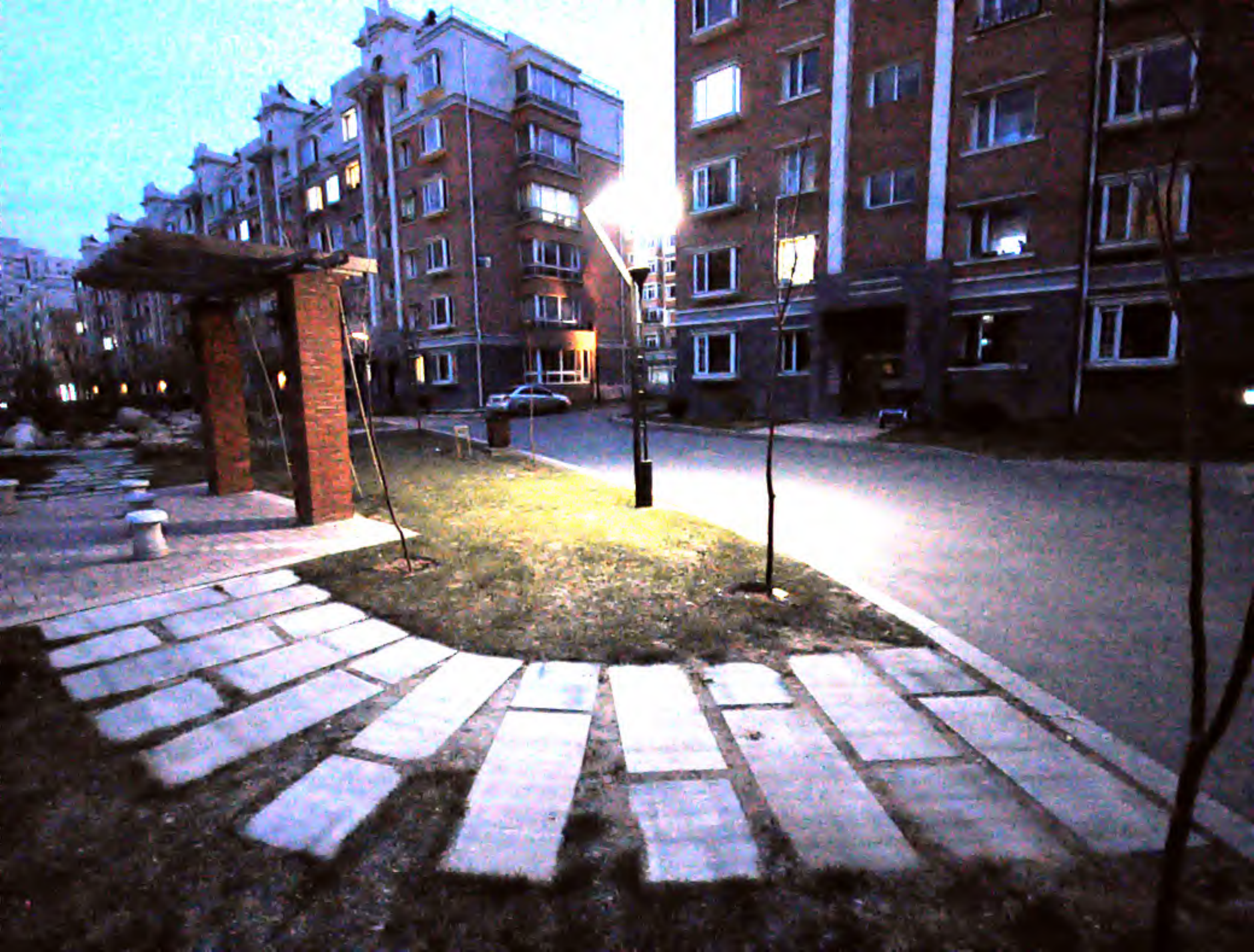}\vspace{-0.2em}
    \centerline{RUAS}\medskip
  \end{subfigure}
  \hfill
  \begin{subfigure}{0.162\linewidth}
    \includegraphics[width=1.\linewidth]{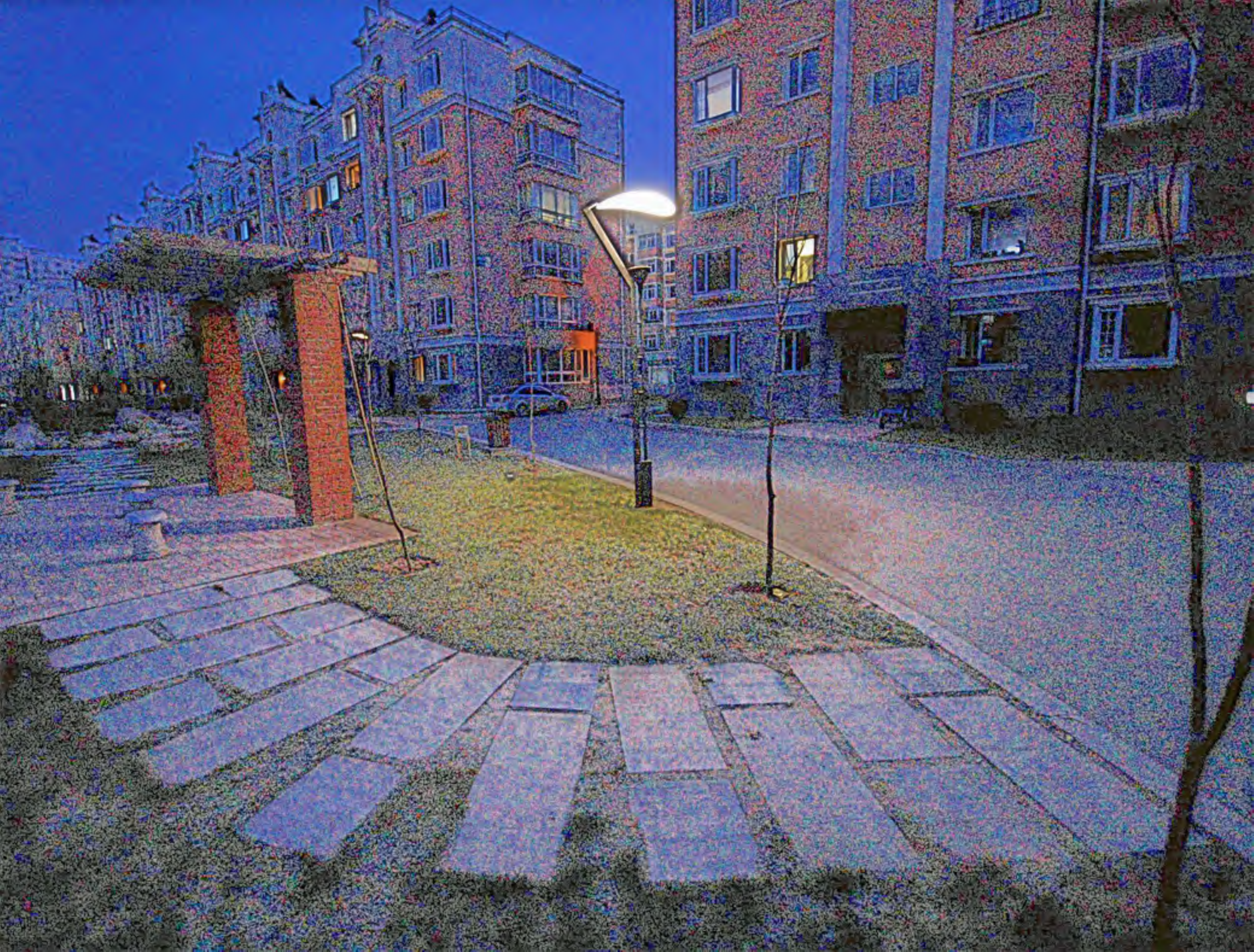}\vspace{-0.2em}
    \centerline{RetinexNet}\vspace{-0.5em}\medskip
    
    \includegraphics[width=1.\linewidth]{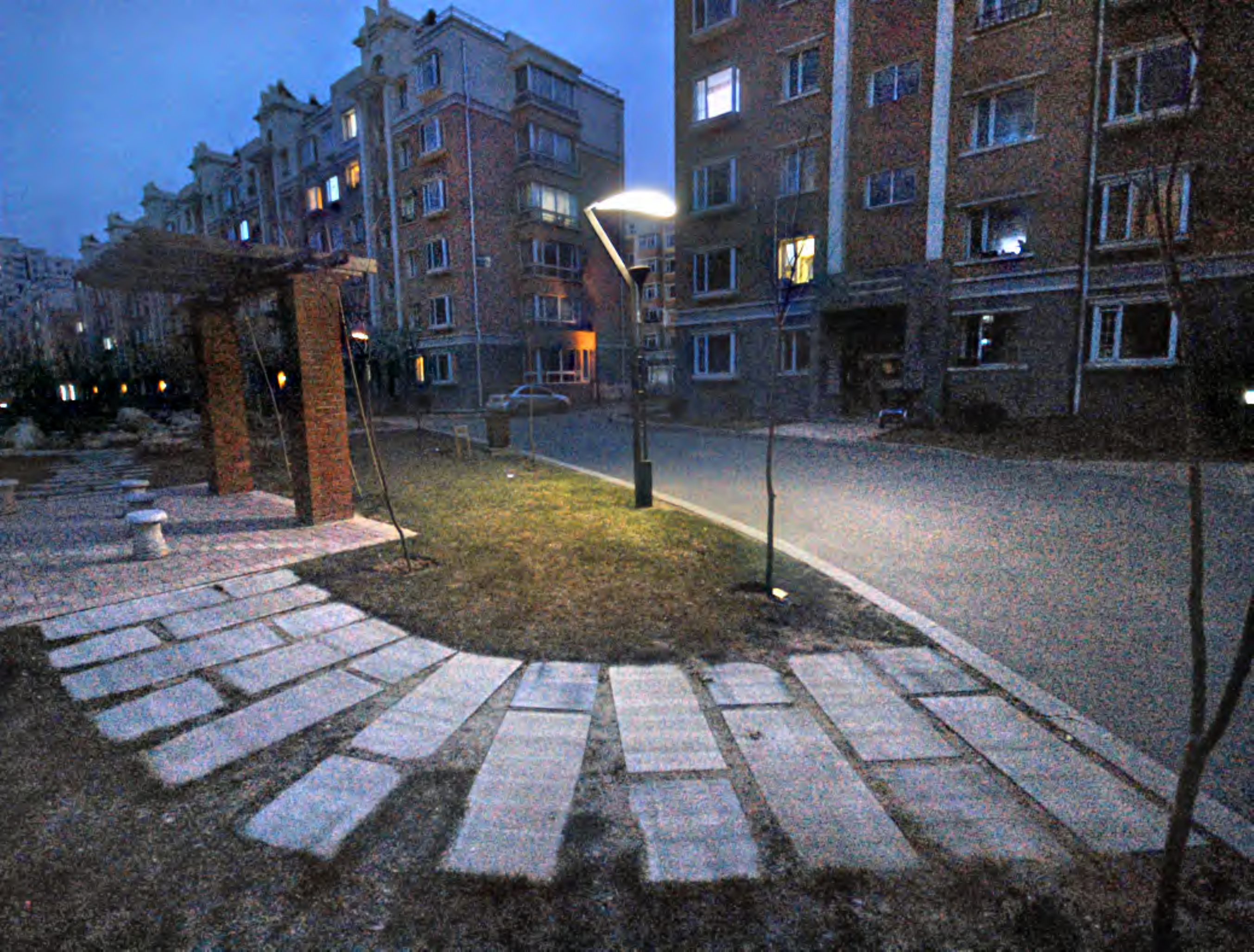}\vspace{-0.2em}
    \centerline{EnGAN}\medskip
  \end{subfigure}
  \hfill
  \begin{subfigure}{0.162\linewidth}
    \includegraphics[width=1.\linewidth]{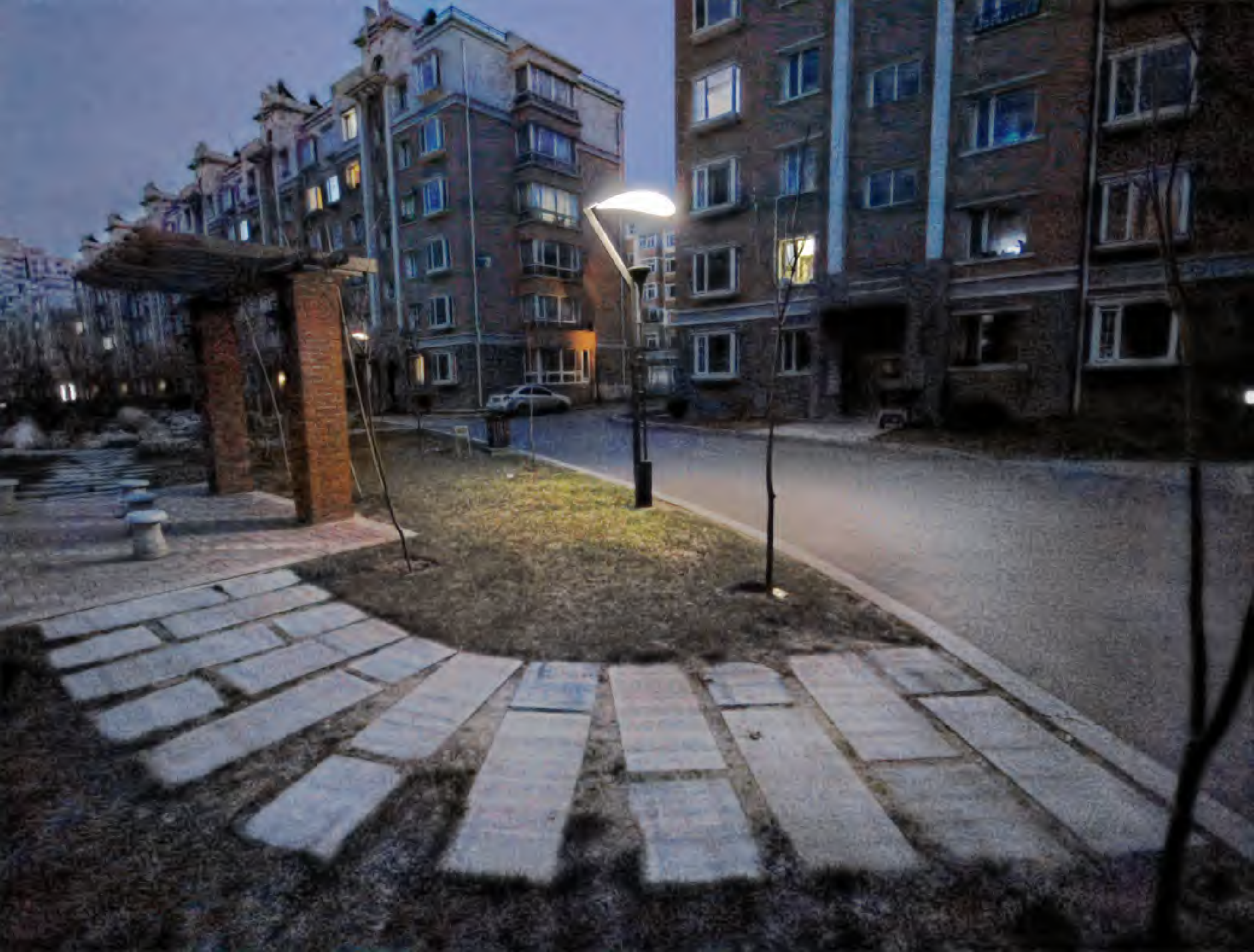}\vspace{-0.2em}
    \centerline{KinD}\vspace{-0.5em}\medskip
    
    \includegraphics[width=1.\linewidth]{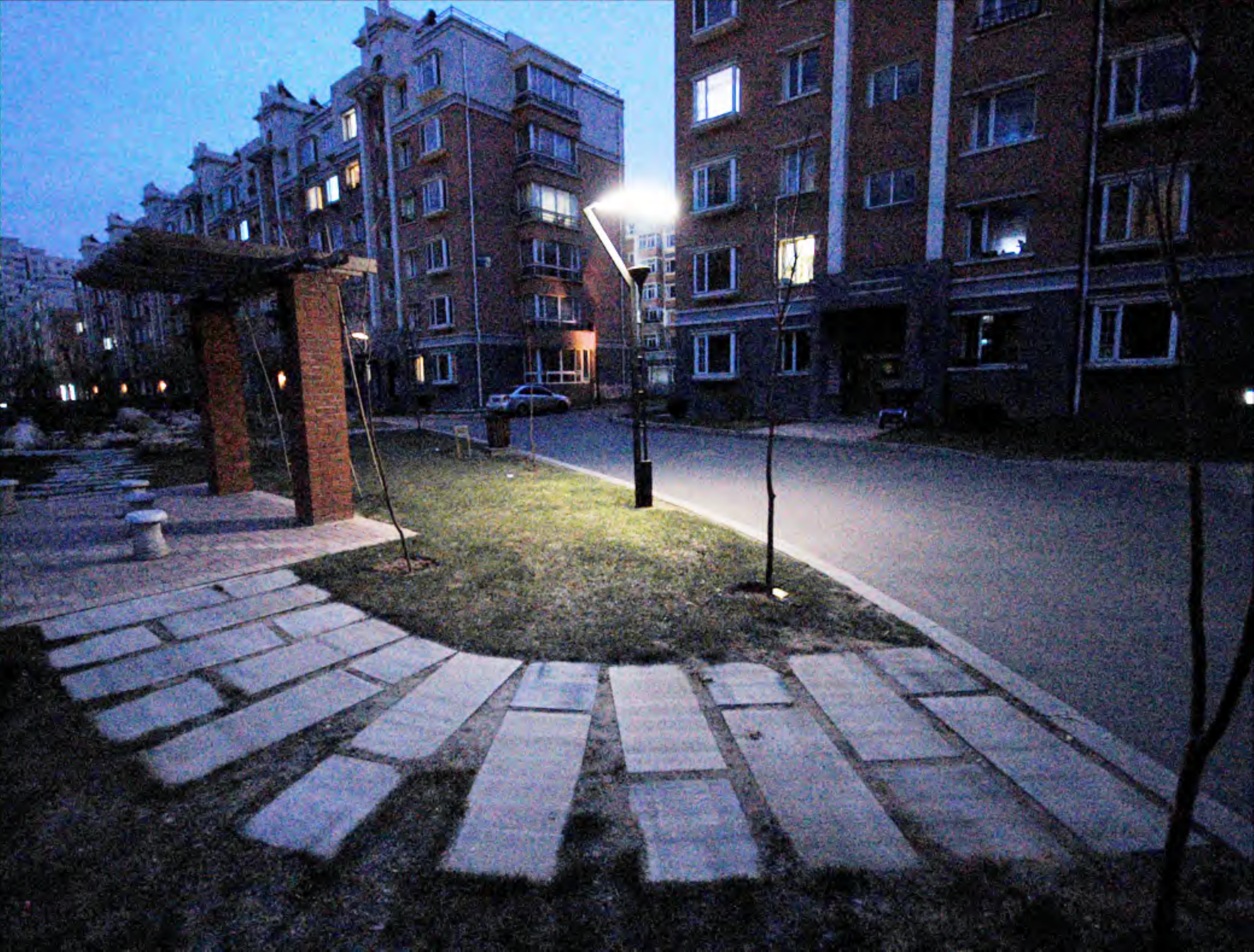}\vspace{-0.2em}
    \centerline{SCI}\medskip
  \end{subfigure}
  \begin{subfigure}{0.162\linewidth}
    \includegraphics[width=1.\linewidth]{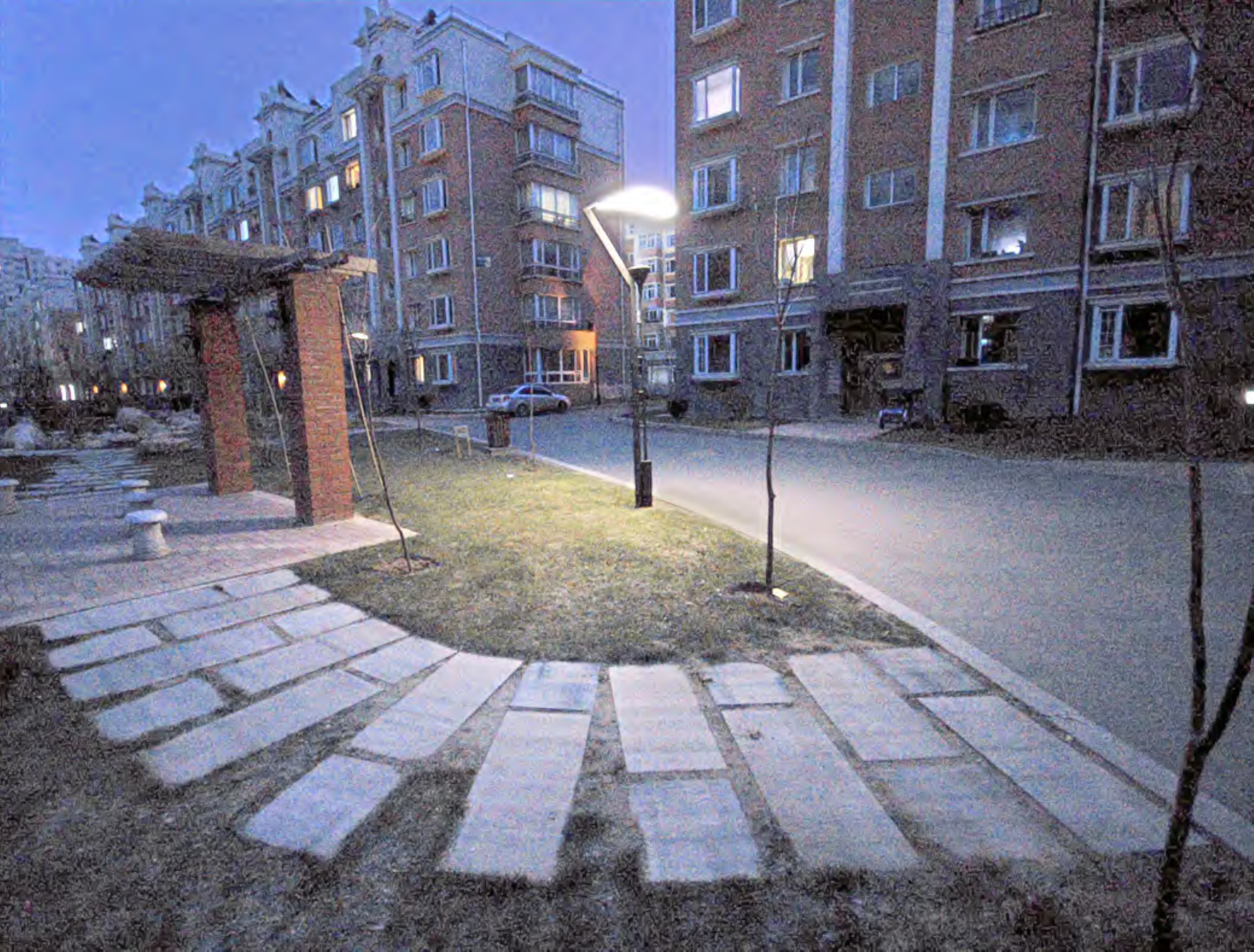}\vspace{-0.2em}
    \centerline{URetinexNet}\vspace{-0.5em}\medskip
    
    \includegraphics[width=1.\linewidth]{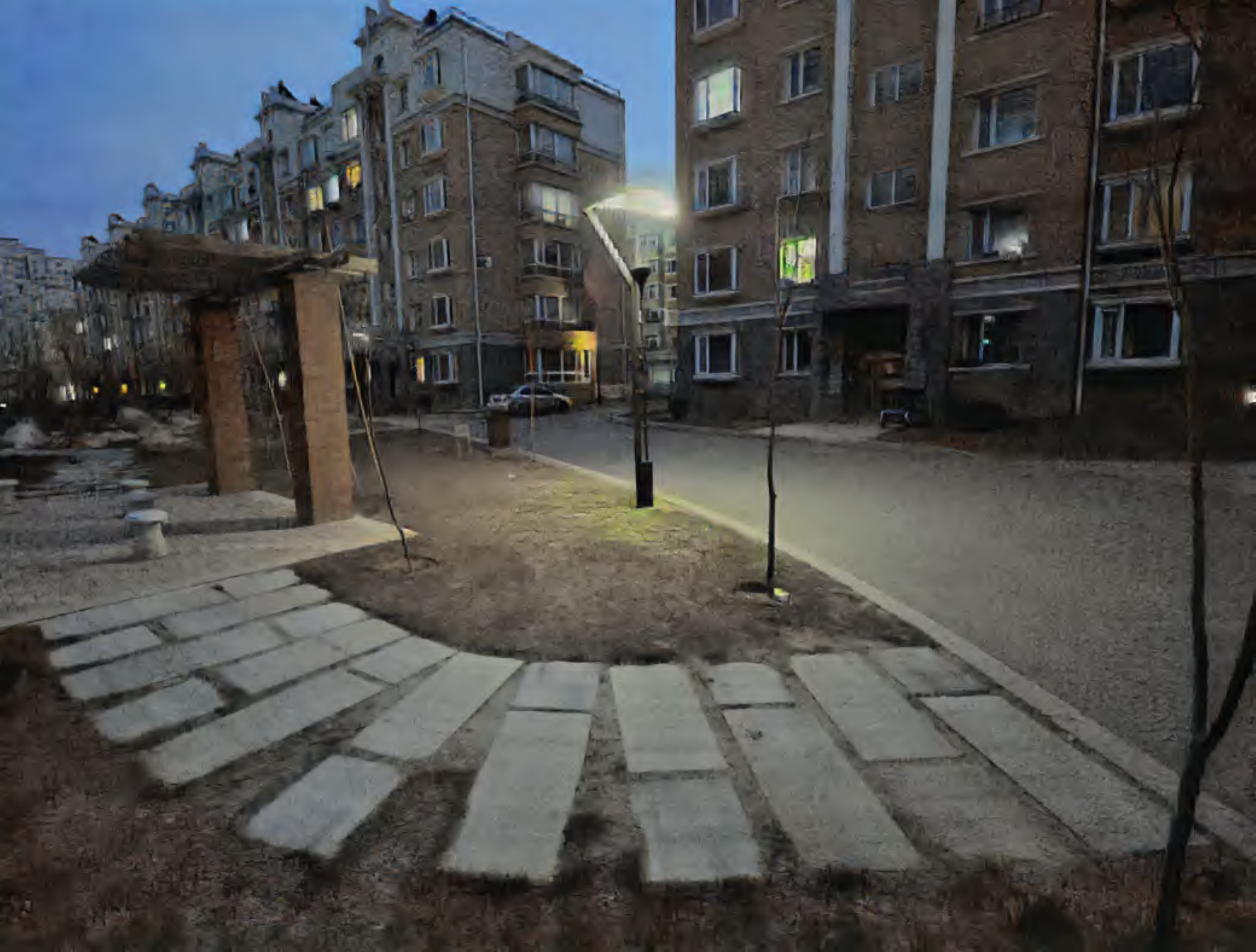}\vspace{-0.2em}
    \centerline{Ours}\medskip
  \end{subfigure}
  \vspace{-1.em}

  \caption{Subjective comparison on the LIME dataset among state-of-the-art low-light image enhancement algorithms. Our model still performs best on this low-light image-only dataset, which proves its effectiveness.}
  \label{comlime}
  \vspace{-1.em}
  
\end{figure*}

\cref{comlsrw1} displays the enhanced results on LSRW dataset. One can see that conventional model-based methods cannot recover sufficient brightness, while some other comparison methods suffer from color cast. RetinexNet, KinD, ZeroDCE, and RUAS, \textit{etc.} develop the post-processing denoising operations to remove the inherent noise in dark regions, but they tend to discard details. In general, our NeRCo is capable of color adjustment and detail preservation, demonstrating its superiority over other algorithms. Furthermore, we provide visual comparisons between our proposed NeRCo and other SOTA mthods on other well-known benchmarks. \cref{comlol} shows the comparisons on LOL dataset and \cref{comlime} displays the qualitative results on LIME dataset. Obviously, across all these comparisons, our method recovers the most authentic tones and provides visual-friendly results, which proves its effectiveness.

\end{document}